\documentclass[10pt,twocolumn,letterpaper]{article}

\usepackage[pagenumbers]{cvpr} 

\usepackage{array}

\usepackage{amssymb}
\usepackage{xcolor}
\definecolor{visible}{HTML}{00A4FD}
\definecolor{masked}{HTML}{E8B6E3}
\definecolor{recovered}{HTML}{9BE4B3}
\definecolor{lost}{HTML}{F7EEF6}
\definecolor{neighborhood}{HTML}{B2B9C0}

\definecolor{bosch_purple}{RGB}{228,114,219}
\definecolor{bosch_purple_dark}{HTML}{9E2896}
\definecolor{bosch_turquoise}{RGB}{84,171,165}
\definecolor{bosch_turquoise_dark}{HTML}{18837E}
\definecolor{bosch_green_dark}{HTML}{00884A}
\definecolor{bosch_green}{RGB}{74,176,115}

\usepackage{tikz}
\usepackage{pgfplots}
\pgfplotsset{width=10cm,compat=1.9}

\definecolor{cvprblue}{rgb}{0.21,0.49,0.74}
\usepackage[pagebackref,breaklinks,colorlinks,allcolors=cvprblue]{hyperref}
\usepackage{pifont}
\usepackage{xcolor}
\usepackage{microtype}
\usepackage{flushend}

\usepackage{multirow}
\newcommand{\cmark}{\ding{51}}%

\newcommand\worrynomore[1]{\textcolor{black}{#1}}

\def\paperID{1815} 
\def\confName{CVPR}
\def\confYear{2025}

\title{Multi-Scale Neighborhood Occupancy Masked Autoencoder for\\Self-Supervised Learning in LiDAR Point Clouds}

\def\authorBlock{
    Mohamed Abdelsamad$^{1,2}$\quad
    Michael Ulrich$^1$\quad
    Claudius Gläser$^1$\quad
    Abhinav Valada$^2$\\

    ${}^1$Bosch Center for AI \quad\quad
    ${}^2$University of Freiburg
}

\author{\authorBlock}

\begin{document}
\maketitle

\begin{abstract}

Masked autoencoders (MAE) have shown tremendous potential for self-supervised learning (SSL) in vision and beyond. 
However, point clouds from LiDARs used in automated driving are particularly challenging for MAEs since large areas of the 3D volume are empty.
Consequently, existing work suffers from leaking occupancy information into the decoder and has significant computational complexity, thereby limiting the SSL pre-training to only 2D bird's eye view encoders in practice.
In this work, we propose the novel neighborhood occupancy MAE (NOMAE) that overcomes the aforementioned challenges by employing masked occupancy reconstruction only in the neighborhood of non-masked voxels. We incorporate voxel masking and occupancy reconstruction at multiple scales with our proposed hierarchical mask generation technique to capture features of objects of different sizes in the point cloud.
NOMAEs are extremely flexible and can be directly employed for SSL in existing 3D architectures. We perform extensive evaluations on the nuScenes and Waymo Open datasets for the downstream perception tasks of semantic segmentation and 3D object detection, comparing with both discriminative and generative SSL methods. The results demonstrate that NOMAE sets the new state-of-the-art on multiple benchmarks for multiple point cloud perception tasks.
\end{abstract}    
\section{Introduction}
\label{sec:intro}


\begin{figure}
  \centering
    \includegraphics[width=0.85\linewidth]{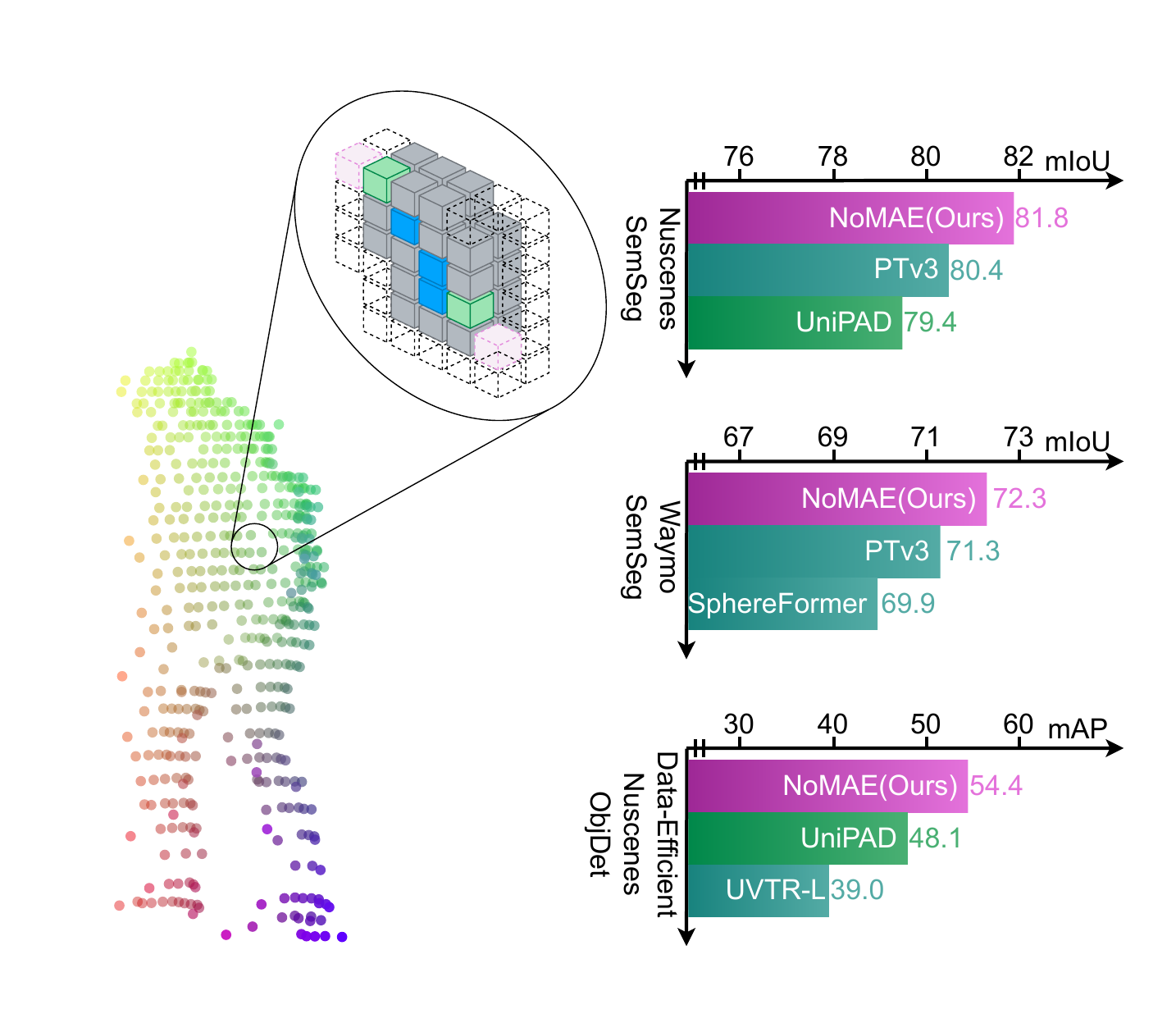}
    \vspace{-1em}
  \caption{NOMAE enables masking and reconstructing occupancy as a self-supervised pretext task for large-scale point clouds. 
It limits the reconstruction of masked voxels\,\fcolorbox{lost}{lost}{\rule{0pt}{4pt}\rule{0pt}{0pt}}\fcolorbox{recovered}{recovered}{\rule{0pt}{4pt}\rule{0pt}{0pt}} to the neighborhood\,\fcolorbox{neighborhood}{neighborhood}{\rule{0pt}{4pt}\rule{0pt}{0pt}}\fcolorbox{recovered}{recovered}{\rule{0pt}{4pt}\rule{0pt}{0pt}} of visible\,\fcolorbox{visible}{visible}{\rule{0pt}{4pt}\rule{4pt}{0pt}} voxels 
  and reconstructs the masked occupancy at multiple scales. 
  NOMAE\,\fcolorbox{bosch_purple}{bosch_purple}{\rule{0pt}{4pt}\rule{4pt}{0pt}} achieves state-of-the-art performance on nuScenes semantic segmentation, Waymo semantic segmentation, and nuScenes object detection tasks, outperforming existing self-supervised methods\,\fcolorbox{bosch_green}{bosch_green}{\rule{0pt}{4pt}\rule{4pt}{0pt}} as well as transformer methods\,\fcolorbox{bosch_turquoise}{bosch_turquoise}{\rule{0pt}{4pt}\rule{4pt}{0pt}}. 
  }
  \label{fig:teaser}
  \vspace{-1em}
\end{figure}

Sensors that generate point clouds, such as LiDARs or radars, have become a cornerstone in automated driving as they provide high-resolution three-dimensional representations of the environment~\cite{schramm2024bevcar}. The rich spatial information represented in point clouds enables vehicles to accurately detect and classify objects, navigate complex environments, and enhance safety through real-time situational awareness. However, annotated point cloud datasets are significantly smaller than their image-based counterparts, which makes learning large-scale perception models extremely challenging. Self-supervised learning (SSL)~\citep{bao2021beit,chen2020simclr,he2020moco,chen2021simsiam,he2022mae,tian2023spark,chen2020improved,gao2022convmae,gosala2023skyeye,lang2024self,lang2023self}, through contrastive learning or masked modeling, provides an effective solution to this problem by learning meaningful representations from vast amounts of unlabeled data. SSL also reduces the reliance on arduous annotation processes while improving performance and generalization. Pioneering works~\citep{liu2022maskpoint,pang2022pointmae,yu2022pointbert,zhang2022pointm2ae} have successfully employed SSL to small-scale indoor point clouds, with more recent efforts extending it to large-scale outdoor point clouds~\citep{hess2022voxelmae,xv2023mvjar,tian2023geomae}. 

Outdoor point clouds, however, pose a unique challenge for masked modeling as most of the measured 3D volume is empty space. Current approaches resort to reconstructing precise point locations within occupied voxels~\citep{hess2022voxelmae,xv2023mvjar,tian2023geomae}, but this often leaks information to the decoder, signaling that the queried voxel is occupied.  Recent methods~\citep{yang2023gd-mae,min2022OccupancyMAESP} attempt to overcome this problem by reconstructing the entire scene, but the computational complexity and class imbalance caused by the large number of empty voxels limit these approaches to either 2D bird's-eye-view representations or coarse-grained 3D reconstructions.

In this work, we propose \textbf{N}eighborhood \textbf{O}ccupancy \textbf{MAE} (NOMAE), the first multi-scale sparse self-supervised learning framework for LiDAR point clouds that directly addresses the problem of 3D point cloud sparsity in masked modeling. The novelty in NOMAE lies in the concept that the occupancy of fine-grained 3D voxels is only evaluated (loss) in the neighborhood of visible (not masked) occupied voxels. This is motivated by the fact that LiDAR points are typically clustered in the proximity of other LiDAR points in outdoor driving scenarios. Thereby, we avoid leaking information about masked voxels to the decoder and eliminate the need for dense feature spaces or reconstructing large unoccupied areas. This makes our approach lightweight and usable with more modern sensors that have finer resolutions as well as state-of-the-art 3D transformer architectures.
Our framework employs self-supervision at multiple scales, made feasible by the lightweight nature of our reconstruction task. 
To facilitate this, we introduce a hierarchical mask generation module that is suitable for multi-scale SSL. We perform extensive experiments with NOMAE on the competitive nuScenes~\citep{caesar2020nuscenes} and Waymo Open~\citep{sun2020wod} datasets that demonstrate state-of-art pretraining performance for multiple downstream tasks (illustrated in Fig.~\ref{fig:teaser}). 
Our main contributions are as follows:
\begin{itemize}
    \item The novel \textit{localized reconstruction} self-supervised learning framework for point clouds, \textbf{N}eighborhood \textbf{O}ccupancy \textbf{MAE}.
    \item A multi-scale SSL strategy, where different feature levels are supervised at different scales.
    \item A novel mask generating scheme suitable for multi-scale SSL.
    \item Extensive benchmarking on two standard autonomous driving datasets, achieving state-of-the-art results across two perception tasks.
    \item Comprehensive ablation studies to highlight the impact of our proposed contributions.
\end{itemize}

\begin{figure*}
  \centering
    \includegraphics[width=1.0\textwidth]{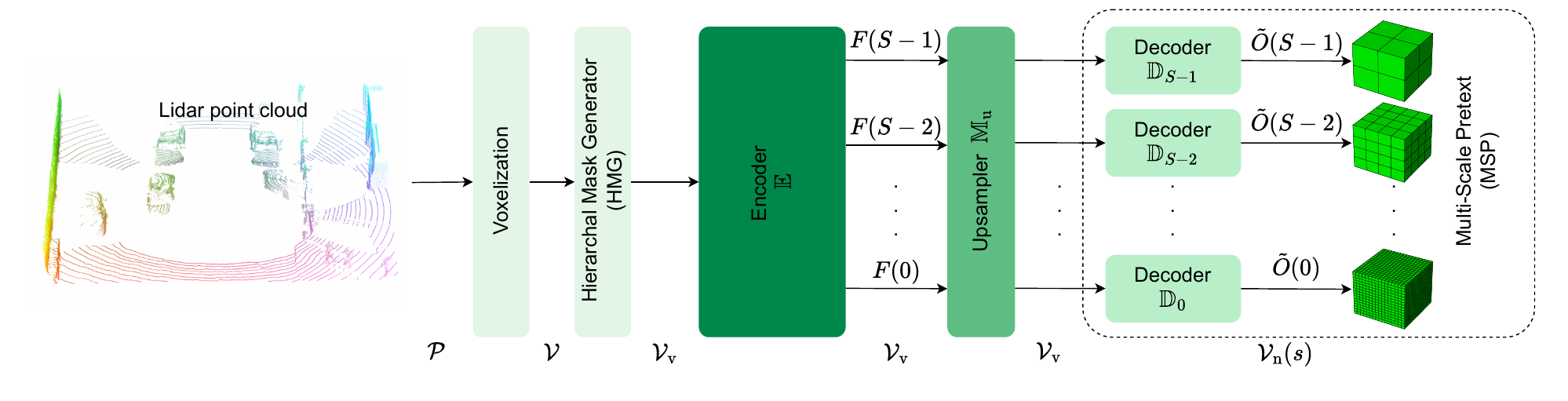}
  \vspace{-2em}
  \caption{Overview of the proposed NOMAE approach. The input point cloud is first voxelized and masked by the hierarchal mask generator. The encoder $\mathbb{E}$ processes the visible voxels $\mathcal{V}_{\text{v}}$ to yield a hierarchical representation. The upsampler $\mathbb{M}_\text{u}$ then fuses the multi-scale representations to capture high-level features at each scale. For each feature scale, a separate neighboring decoder predicts occupancy in $\mathcal{V}_{\text{n}}$, corresponding to the immediate neighborhood of the visible voxels. The combination of independent learning tasks across multiple feature scales and the localized predictions by the neighboring decoders enables learning representations that are well-suited for 3D point clouds.}
  \label{fig:framework}
  \vspace{-0.5em}
\end{figure*}

\section{Related Work}
 


In this section, we review the existing works related on point cloud SSL and Automotive LiDAR SSL.

{\parskip=2pt
\noindent\textbf{3D Self-Supervised Learning}:} The success of generative self-supervised learning in natural language processing and computer vision has inspired several works~\citep{pang2022pointmae, Wang2020UnsupervisedPC, liu2022maskpoint, AbouZeid2023Point2VecFS} to explore masked auto-encoders for 3D point clouds. 
These approaches are typically tailored towards small-scale single object recognition tasks, where a standard ViT~\citep{Dosovitskiy2020AnII} architecture suffices to encode the point cloud. 
For example, PointMAE~\citep{pang2022pointmae} explicitly reconstructs point cloud patches using the Chamfer distance. 
MaskPoint~\citep{liu2022maskpoint} discriminates between the reconstructed points and noise points. 
In OcCo~\citep{Wang2020UnsupervisedPC}, point cloud completion is performed on occluded regions. 
Most prominently, Point2Vec~\citep{AbouZeid2023Point2VecFS} reconstructs the encoded features of a teacher model for the masked patches. 
Alternatively, contrastive methods can be employed with point clouds to distinguish multiple partial views~\citep{xie2020pointcontrast} or point-level correspondences~\citep{Hou2020ExploringD3}. 
These pioneering works achieve promising results on object-scale and room-scale 3D point clouds but are not usable for large-scale automotive LiDAR point clouds due to their inefficient scaling with the size of the point cloud. 
In contrast, our work focuses on large-scale automotive LiDAR point clouds and employs efficient hierarchical architectures. 
Additionally, our work proposes self-supervision for multiple feature levels, contrary to the single-scale supervision employed in these works. 

{\parskip=2pt
\noindent\textbf{Automotive LiDAR Self-Supervised Learning}: 
The focus of existing works on automotive large-scale point clouds is computational efficiency. 
One common technique is to reduce 3D point clouds to a 2D bird's-eye-view (BEV) grid, using pillar architectures~\citep{xv2023mvjar,yang2023gd-mae,hess2022voxelmae,tian2023geomae, boulch2023also}. 
\citep{hess2022voxelmae} reconstructs 3D points inside masked 2D pillars and \citep{xv2023mvjar} additionally predicts the order of the 2D BEV pillars. 
GeoMAE~\citep{tian2023geomae} predicts centroids and 3D sub-occupancy in the pretraining. 
GD-MAE~\citep{yang2023gd-mae} utilizes a generative decoder to reconstruct the whole scene to alleviate the leakage of positional information to the decoder. ALSO~\citep{boulch2023also} reconstructs the surface occupancy of the point cloud. UniPAD~\citep{Yang2023UniPADAU} is a pioneering work that generates coarse 3D features of the whole scene and uses a neural rendering approach for supervision.
Occupancy-MAE~\citep{min2022OccupancyMAESP} proposes to utilize the occupancy as a compressed representation of the point cloud and reconstruct the occupancy in the 3D space using a masked point cloud. 
Despite the significant performance achieved by them, reconstructing occupancy or features over the entire 3D volume is an expensive task, which allows supervision only on a coarse-scale.
Furthermore, the large computational cost prohibits employing modern 3D scene understanding architectures with high voxel resolutions.
As a result, the state of the art for self-supervised learning lags behind the performance of a fully supervised training of transformer architectures from scratch. 
In contrast, this paper addresses fully sparse SSL as a remedy, scaling well with higher 3D voxel resolutions.} 
\section{Technical Approach}

Fig.~\ref{fig:framework} presents an overview of our proposed framework for self-supervised representation learning. The network consists of an encoder (to be trained), a token upsampling module, and multiple decoders for the hierarchical masked voxel reconstruction. We employ PTv3~\citep{Wu2023PointTV} as the encoder. In contrast to earlier works, we maintain a sparse feature space and generate fine-grained features only for the visible voxels using an upsampling module. 
We employ a sparse decoder for masked voxel reconstruction. This decoder is designed to be simple and lightweight, as described in Sec.~\ref{sec:decoder}, allowing us to deploy a separate decoder instance at each feature scale in the multi-scale pretext (MSP), which is further detailed in Sec.~\ref{sec:MSP}. 
The input point clouds are first voxelized and then masked before being fed into the encoder.  Our masking strategy ensures that there is adequate masking coverage while also maintaining a sufficient number of occupied voxels in the reconstructed neighborhoods at multiple hierarchical scales, as explained in Sec.~\ref{sec:HMG}.

\subsection{Encoder and Token Upsampling}
\label{sec:encoder}
This input point cloud $\mathcal{P}$ is first voxelized to obtain the set of all occupied voxels $\mathcal{V}$, which is then split into a set of visible voxels $\mathcal{V}_\text{v}$ and masked voxels $\mathcal{V}_\text{m}$, as detailed in Sec.~\ref{sec:HMG}. 
A sparse transformer encoder $\mathbb{E}$ based on PTv3~\citep{Wu2023PointTV} is used to encode the features $V$ at the positions of $\mathcal{V}_\text{v}$ to generate the set of tokens
\begin{align}
    F(s) &= \mathbb{E}(V)(s).
\end{align}
PTv3 employs partition-based pooling on the tokens to generate more abstract representations for coarser resolutions, similar to pooling in CNNs. 
$F(s)$ is the features (tokens) at the $s$-th scale level of PTv3 and $s\in\{0,...,S-1\}$. 

NOMAE uses an upsampling module $\mathbb{M}_\text{u}$ to propagate the abstract encoding of coarser resolution tokens to the tokens of finer resolution while keeping the representation sparse. This is similar to a feature pyramid network for CNNs. $\mathbb{M}_\text{u}$ consists of a single PTv3 transformer block at every scale, which is very lightweight. 

\subsection{Neighboring Decoder}
\label{sec:decoder}
\begin{figure}
  \centering
  \begin{subfigure}{\linewidth}
    \centering
    \includegraphics[trim={0 22cm 6cm 0},clip,width=\textwidth]{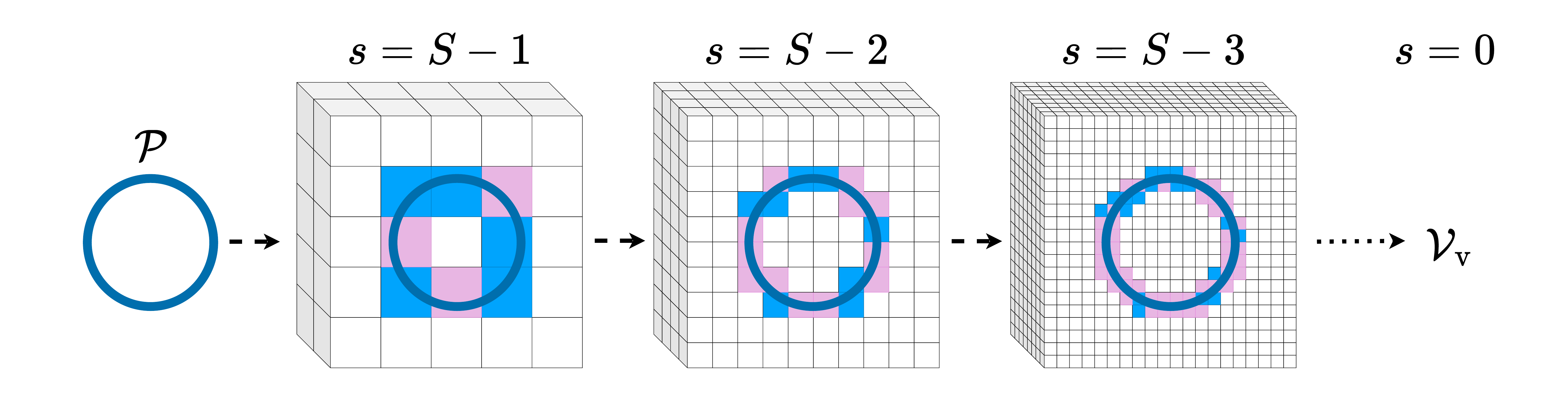}
    \caption{Hierarchical mask generation renders $\mathcal{P}$ to the coarsest scale $s=S-1$ and then applies a random mask to divide occupied voxels into visible\,\fcolorbox{visible}{visible}{\rule{0pt}{2pt}\rule{2pt}{0pt}} and masked\,\fcolorbox{masked}{masked}{\rule{0pt}{2pt}\rule{2pt}{0pt}} voxels. 
    For the subsequent scales, only the visible voxels of the previous scale undergo random masking. 
    Masked voxels of a coarser scale always correspond to masked\,\fcolorbox{masked}{masked}{\rule{0pt}{2pt}\rule{2pt}{0pt}} or empty\,\fcolorbox{black}{white}{\rule{0pt}{2pt}\rule{2pt}{0pt}} voxels at a finer scale.}
    \label{fig:HMG}
  \end{subfigure}
  \begin{subfigure}{\linewidth}
    \centering
    \includegraphics[trim={0 0 6cm 23cm},clip,width=\textwidth]{figure/HMG.pdf}
    \hfill
    \caption{Multiscale pretext reconstructs the cells around visible voxels\,\fcolorbox{visible}{visible}{\rule{0pt}{2pt}\rule{2pt}{0pt}} at multiple scales. 
    The neighborhood $\mathcal{V}_n$ contains voxels that are reconstructed but empty\,\fcolorbox{neighborhood}{neighborhood}{\rule{0pt}{2pt}\rule{2pt}{0pt}} as well as masked and recovered voxels\,\fcolorbox{recovered}{recovered}{\rule{0pt}{2pt}\rule{2pt}{0pt}}. 
    Some masked voxels which are too distant from visible voxels are not recovered\,\fcolorbox{lost}{lost}{\rule{0pt}{2pt}\rule{2pt}{0pt}} and do not contribute to the loss. 
    The predicted occupancy $\Tilde{O}$ at coarser scales covers a wider region, while finer scales reconstruct more detail. 
    }
    \end{subfigure}
\caption{Illustration of multiscale pretext (MSP) and hierarchical mask generation (HMG).}
\label{fig:HMG_MSP_example}
\vspace{-.5em}
\end{figure}

Prior work on self-supervised learning for large-scale point clouds reconstruct exact locations of points~\citep{hess2022voxelmae}, geometrical properties~\citep{tian2023geomae} or voxel ordering~\citep{xv2023mvjar}, for each masked voxel $\mathcal{V}_\text{m}$. 
This requires passing $\mathcal{V}_\text{m}$ to the decoder, causing information leakage. 
\citep{min2022OccupancyMAESP,yang2023gd-mae} avoid this well-known information leakage by reconstructing the scene as a whole, which is computationally expensive. In contrast, NOMAE reconstructs the occupancy $O(v_\text{n},s)$ of all voxels $v_\text{n} \in \mathcal{V}_\text{n}$ within a certain neighborhood $n$ of visible voxels $\mathcal{V}_\text{v}$ at scale $s$. 
To achieve this, we employ a decoder $\mathbb{D}_\text{s}$ consisting of sparse convolution layers. 
\begin{align}
    \tilde{O}(v_\text{n},s) &= \mathbb{D}_\text{s}(\mathbb{M}_\text{u}(F)(s)),
\end{align}
where $\tilde{O}$ is the networks prediction of $O$.
Visible voxels of $\mathcal{V}_\text{v}$ are excluded from $\mathcal{V}_\text{n}$. 
An example is depicted in Fig.~\ref{fig:HMG_MSP_example}. 

We note that $\mathcal{V}_\text{n}$ will not cover masked voxels in $\mathcal{V}_\text{m}$ that are not nearby visible voxels. This is an approximation that we make in our approach and we observed that LiDAR points in outdoor point clouds typically lie in the proximity of other points on the surfaces of objects. Hence, sufficient number of masked voxels in $\mathcal{V}_\text{m}$ are included in $\mathcal{V}_\text{n}$. Faraway isolated points do not contribute to the loss, which improves performance, as evaluated in the ablation study presented in Sec.~\ref{sec:ablations_neigh_size}. Our interpretation is that such isolated points belong to strongly occluded or masked objects and are infeasible to reconstruct, hence affecting the pretraining. 
This approximation allows us to avoid reconstructing large volumes of unoccupied space without leaking information about $\mathcal{V}_\text{m}$ to the decoder $\mathbb{D}_\text{s}$ at the same time. 

\subsection{Multi-Scale Pretext}
\label{sec:MSP}
GeoMAE~\citep{tian2023geomae} is the only prior work that exploits multiple hierarchical scales in the reconstruction during self-supervised training. 
Most other works~\citep{min2022OccupancyMAESP,yang2023gd-mae,hess2022voxelmae,tian2023geomae,boulch2023also} use a single scale for their pretraining tasks. This is unexpected since it is common practice in automated driving perception models to attach task heads to feature representations at different scales. The intuition is that finer resolutions are more suitable for small objects, such as pedestrians, while coarser resolutions are more suitable for larger objects, such as trucks. However, the multi-scale reconstruction in GeoMAE~\citep{tian2023geomae} is derived from a single feature map. 
In our approach, we use $s$ instances $\mathbb{D}_\text{s}$ of the decoder architecture $\mathbb{D}$ with separate weights. Coupled with the neighborhood size being scale-dependent, this implicitly encourages the coarse-grained features to contain information from a larger area, while the fine-grained features contain more localized details.\looseness=-1

\subsection{Hierarchal Mask Generator}
\label{sec:HMG}
Generally, the scale of the masking can be different from the scale of reconstruction. For example, a random mask can be generated on a coarse scale and then upsampled to match the resolution of the reconstruction. However, experiments in \citep{hess2022voxelmae,yang2023gd-mae,tian2023geomae} showed that random masking on the same scale as the reconstruction scale performs best, which is intuitive. Consequently, we generate random masks for multiple scales in the MSP. These masks should be consistent to avoid information leakage, i.e., a voxel that is masked on a coarser scale should not be visible on a finer scale~\citep{zhang2022pointm2ae}. A straightforward manner to generate consistent masks would be to create a random mask for the finest scale and then derive the masks of coarser scales by defining a coarse-scale voxel as masked if all its corresponding fine-scale sub-voxels are masked. However, this would lead to rapidly decreasing masking ratios when moving to coarser scales because a single visible sub-voxel is sufficient to mask a coarser-scale voxel visible. Another alternative is masking at the coarsest scale and upsampling the mask~\citep{zhang2022pointm2ae} to finer scales which leads to a more consistent masking ratio across scales. However, we found that this approach rapidly decreased the size of reconstructed neighborhoods $n$ moving to finer scales.

Hence, we propose the masking scheme depicted in Fig.~\ref{fig:HMG}. We mask the coarsest scale first, using a random sampling of all occupied voxels $\mathcal{V}$, and the probability that a voxel is masked equals masking ratio $r$. Next, we take all voxels at scale $s-1$ within visible voxels of the previous, coarser, scale $s$ and repeat the sampling of additional masked voxels at scale $s-1$ with probability $r$.Consequently, the total masking ratio at scale $r_\text{t}(s)$ is approximately
\begin{align}
    \label{equ:total_masking_ratio}
    r_\text{t}(s) \approx 1-(1-r)^{S-s+1}.
\end{align}
By doing so, we ensure that coarser scales have a sufficient number of masked voxels without reducing the size of reconstructed neighborhoods at finer scales.
Only the visible voxels $\mathcal{V}_\text{v}$ of the finest scale are fed to the encoder backbone $\mathbb{E}$. An ablation study presented in Sec.~\ref{sec:ablation_MSP_HMG} quantifies the positive effect of HMG on the pretraining and downstream task performance. 

\subsection{Pretrainging Loss}
We use the Binary Cross Entropy loss (BCE) as the occupancy loss per scale. The final loss $L$ is the average of all single scale losses $L(s)$:
\begin{equation}
    L = \frac{1}{S} \sum_{s=0}^{S-1} \frac{1}{|\mathcal{V}_\text{n}(s)|} \sum_{v \in \mathcal{V}_\text{n}(s)} \text{BCE}(\tilde{O}(v,s),O(v,s)),
\end{equation}
with the ground truth occupancy $O(v,s) \in \{0,1\}$.

\section{Experiments}
\label{sec:experiments}

\worrynomore{In this section, we discuss the datasets, metrics, and the evaluation protocol that we use for benchmarking. We compare our proposed approach with state-of-the-art methods in the benchmarks and present extensive ablation studies to demonstrate the novelty of our contributions.}

 \begin{table*}
    \centering
    \footnotesize
    \caption{Comparison of LiDAR semantic segmentation performance on the nuScenes and Waymo Open datasets. For the first time, an SSL pretraining method outperforms strong supervised learning models. Methods marked with $^{*}$ are our implementation.}
    \vspace{-0.5em}
\begin{tabular}{lcccccccc}\toprule
\multirow{2}{*}{\textbf{Method}} & \multirow{2}{12mm}{\textbf{SSL\\pretraining}} &\multicolumn{4}{c}{\textbf{nuScenes}}  &\multicolumn{3}{c}{\textbf{Waymo}} \\ \cmidrule(lr){3-6} \cmidrule(lr){7-9}
 & &\textbf{val mIoU} & \textbf{val mAcc} & \textbf{test mIoU} & \textbf{test fwIoU}  & \textbf{val mIoU} & \textbf{val mAcc}& \textbf{test mIoU} \\\midrule
MinkUNet~\cite{ChoyGS2019mink} & - &73.3 &-&-&- &65.9 &76.6&69.8\\
SPVNAS~\cite{tang2020spvnas} & - &- &- &77.4 & 89.7 &- &-&68.0 \\
Cylinder3D~\cite{zhu2021cylinder3d} & - &76.1&- &77.2 & 89.9  &- &-  &-\\
AF2S3Net~\cite{Cheng2021AF2S3NetAF} & - &62.2&- &78.0 & 88.5 &- &-&- \\
2DPASS~\cite{Yan20222DPASS2P} & - &- &-&80.8 & - &- &- &-\\
SphereFormer~\cite{Lai2023SphericalTF} & - &78.4 &- &81.9 & -  &69.9 &-&-\\
PTv2~\cite{wu2022pointtransv2} & - &80.2 &- &82.6 & - &70.6 &80.2&- \\
PTv3~\citep{Wu2023PointTV} & - &80.4 &87.3 &\textbf{82.7} & 91.1 &71.3  &80.5&- \\
\midrule
UniPAD\,~\citep{Yang2023UniPADAU}&\cmark&79.4& -& 81.1&-&-&-&-\\
GEO-MAE\,\citep{tian2023geomae}&\cmark &78.6&-&-&-&-&-&-\\
GEO-MAE\,\citep{tian2023geomae} + PTv3\,\citep{Wu2023PointTV}$^{*}$ &\cmark &78.9& 84.7 &-&-&-&-&-\\
Occupancy-MAE\,\citep{min2022OccupancyMAESP}&\cmark &72.9&-&-&-&-&-&-\\
Occupancy-MAE\,\citep{min2022OccupancyMAESP} + PTv3\,\citep{Wu2023PointTV}$^{*}$&\cmark &80.0&86.1&-&-&-&-&-\\

\midrule
NOMAE (ours)  &\cmark&\textbf{81.8} & \textbf{87.7} & 82.6 & \textbf{91.5} &\textbf{72.3} &\textbf{82.5}&\textbf{70.3} \\
\bottomrule
\end{tabular}
    \vspace{-0.5em}
    \label{tab:sota_semseg}
\end{table*}

\subsection{Datasets and Evaluation Metrics}

{\parskip=2pt
The \textbf{nuScenes} dataset~\citep{caesar2020nuscenes} is a challenging dataset due to the sparsity of the LiDAR point cloud. It consists of 700 driving sequences for training, 150 for validation, and 150 for testing, with annotations for a variety of tasks. We evaluate both semantic segmentation and object detection on the nuScenes dataset. We use the mean intersection over union (mIoU) as the main evaluation metric for semantic segmentation and the nuScenes detection score (NDS) and mean average precision (mAP) for 3D object detection.}

{\parskip=2pt
The \textbf{Waymo Open Dataset}~\citep{sun2020wod} is a large-scale autonomous driving dataset. 
It consists of 798 driving sequences for training, 202 validation sequences, and 150 test sequences. 
We use the mean intersection over union (mIoU) and mean accuracy (mAcc) as the main metrics for evaluating the semantic segmentation performance.}

\subsection{Task Heads}

This section discusses the methods to evaluate the effectiveness of the pretraining and the quality of the learned representation.

{\parskip=2pt
\noindent\textbf{Fine-tuning}: 
\label{sec:fine-tune}
In our comparisons with state-of-the-art methods, fine-tuning follows the self-supervised pre-training of the encoder. For this purpose, a task-specific head is added with randomly initialized weights, and both the encoder $\mathbb{E}$ and the task-specific head are trained using the annotated dataset. We use the same head as PTv3~\citep{Wu2023PointTV} for the semantic segmentation tasks and the same head as our baseline UVTR~\citep{li2022uvtr} for the object detection task. 
We use layer-wise learning rate decay (LLRD)~\cite{he2022mae} to avoid forgetting the SSL representations in the encoder.}

{\parskip=2pt
\noindent\textbf{Non-linear Probing}: 
\label{sec:nonLP}
The purpose of the ablation study is to evaluate the learned representation from our SSL approach on the downstream semantic segmentation task. 
Therefore, the encoder is kept frozen after pre-training, i.e., no fine-tuning. 
Following Probe3D~\citep{ElBanani2024ProbingT3} and the insights of earlier works~\citep{he2022mae,chen2021simsiam}, we use a multi-scale non-linear probe (NonLP) instead of the commonly used linear probing protocol. NonLP aggregates the feature tokens of all scales after up-sampling tokens of coarser scales before passing them to a voxel-wise small MLP. NonLP avoids that the representation learning happens in the head. Still, it is probing the stronger but non-linear features, correlating better with transfer performance~\citep{he2022mae,chen2021simsiam}.
Similar to the commonly used linear probe, NonLP is trained for a few epochs using annotated data.}

\subsection{Implementation Details}
\label{sec:implementation_details}
We perform the experiments for semantic segmentation in the Pointcept~\cite{pointcept2023} framework and in the MMDetection3D~\citep{mmdet3d2020} framework for the object detection task. We use PTv3~\cite{Wu2023PointTV} as the encoder $\mathbb{E}$, unless stated otherwise. We use a single NVIDIA A100 GPU for the pretraining. 
We use $S=4$ for the reconstruction and masking scales, corresponding to target voxel sizes of $\{0.05,0.10,0.20,0.40\}$ meters. The input voxel dimension is $0.05$ meters. The masking ratio $r_\text{s}$ of the finest scale is $70\%$ for nuScenes and $85\%$ for Waymo. In the self-supervised pre-training, the decoder $\mathbb{D}$ consists of sparse convolution layers~\citep{spconv2022} of kernel size $5$, followed by a single sparse submanifold convolution layer to generate $\tilde{O}$. We use common augmentation techniques such as rotation, scaling, and jittering from the semantic segmentation literature~\cite{ChoyGS2019mink,tang2020spvnas,Wu2023PointTV} during pretraining. 

\def\objdetsec{tf} 

\if\objdetsec
\begin{table}
	\centering
        \footnotesize
        \caption{Comparisons of object detection performance on the nuScenes validation set without the use of test-time augmentation.}
	\vspace{-0.5em}
        \begin{tabular}{lcc}
		\toprule
		\textbf{Methods} & \textbf{NDS$\uparrow$} & \textbf{mAP$\uparrow$} \\
		\midrule
		PVT-SSD~\cite{yang2023pvtssd} & 65.0 & 53.6 \\
		CenterPoint~\cite{Yin2020Centerbased3O} & 66.8 & 59.6 \\
		FSD~\cite{fan2022fsd} & 68.7 & 62.5 \\
		VoxelNeXt~\cite{chen2023voxelnext} & 68.7 & 63.5 \\
		LargeKernel3D~\cite{chen2023largekernel3d}  & 69.1 & 63.3 \\
		TransFusion-L~\cite{bai2022transfusion} &  70.1 & \textbf{65.1} \\
        CMT-L~\cite{yan2023cmt} & 68.6 & 62.1 \\
		UVTR-L-V0.1~\cite{li2022uvtr} & 66.4 & 59.3 \\
		UVTR-L-V0.075~\cite{li2022uvtr} & 67.7 & 60.9 \\
		UniPAD-L~\cite{Yang2023UniPADAU} & \textbf{70.7} & 65.0 \\
        \midrule
		NOMAE (ours) & 69.7 & 63.7 \\
		\bottomrule
	\end{tabular}
	\label{tab:objectdet_sota}
\end{table}
\else
\begin{table}
\centering
\footnotesize
\caption{Comparison of object detection performance on the nuScenes dataset with state-of-the-art point-based pre-training methods. Following the evaluation protocol of~\citep{yang2023gd-mae,Yang2023UniPADAU}, the methods are finetuned using 20\% labeled frames, without CBGS and copy-paste augmentation.}
\begin{tabular}{p{3.6cm} >{\centering\arraybackslash}p{1.6cm} >{\centering\arraybackslash}p{1.6cm}}
    \toprule
    \textbf{Methods} & \textbf{NDS} & \textbf{mAP} \\
    \midrule
    UVTR-L (Baseline)  & 46.7 & 39.0 \\
    \quad+ALSO\,\citep{boulch2023also} &  48.2 & 41.2 \\
    \quad+GD-MAE\,\citep{yang2023gd-mae} &  48.8 & 42.6 \\
    \quad+Learning from 2D\,\citep{Liu2021LearningF2}& 49.2 & 48.8 \\
    \quad+UniPAD\,\citep{Yang2023UniPADAU}  & 55.8 & 48.1 \\
    \midrule
    \quad+NOMAE (ours) & \textbf{60.9} & \textbf{54.4} \\
\bottomrule
\end{tabular}
\label{tab:objectdet_ssl}
\vspace{-0.5em}
\end{table}
\fi

\subsection{Benchmarking Results}
\label{sec:benchmarking_results}

In this section, we present the benchmarking results for both 3D semantic segmentation and 3D object detection.

{\parskip=2pt
\noindent\textbf{3D Semantic Segmentation}: 
Tab.~\ref{tab:sota_semseg} summarizes the best-performing methods on the nuScenes and Waymo Open Dataset leaderboards for the LiDAR semantic segmentation task. The results include the most performant self-supervised pretraining methods. We fine-tuned our model as described in Sec.~\ref{sec:fine-tune}. We observe that our proposed self-supervised pretraining achieves a mIoU score of $81.8$ on the nuScenes validation set, adding $1.4$ mIoU points over the baseline and setting the state-of-the-art. The results on the nuScenes semantic segmentation test set are on par with the strong PTv3 baseline for the mIoU score and outperforms it in the frequency-weighted IoU (fwIoU) score by 0.4$\%$. 
The improvement on the test set is lesser than the validation dataset because NOMAE has relatively poor performance for the minority class \textit{bicycle}, and we did not make special adaptations for the test set submission.}

On the Waymo Open dataset val set, our method achieves a mIoU score of $72.3$ and mAcc of $75.2$. NOMAE improves by $1.0$ and $2.0$ points in the mIoU and mAcc, respectively, over the baseline PTv3~\citep{Wu2023PointTV}. The current version of the semantic segmentation challenge is relatively new for the Waymo Open dataset, with only a few submissions. With 70.3$\%$ mIoU score on the test set, NOMAE sets a new state-of-the-art on the Waymo Open Dataset single frame semantic segmentation challenge.

\begin{table}
    \centering
    \footnotesize
    \caption{Comparison of different SSL methods using the same backbone with NonLP. The iteration time (iter.time) is computed using a batch size of 1. Sup.res is the finest resolution of pretraining supervision.}
    \vspace{-0.5em}
%
%
\addtolength{\tabcolsep}{-0.5em}
\begin{tabular}{lcccc}
\toprule
\textbf{Model} & \textbf{mIoU} & \textbf{mACC} & \textbf{iter.time} & \textbf{sup.res} \\
\midrule
  GEO-MAE\,\citep{tian2023geomae} + PTv3\,\citep{Wu2023PointTV} &53.8&67.7&40.1ms&0.20m\\ 
  Occupancy-MAE\,\citep{min2022OccupancyMAESP} + PTv3\,\citep{Wu2023PointTV} &59.0&73.4&74.0ms&0.10m\\ 
  NOMAE (Ours) &\textbf{74.8}&\textbf{85.0}&\textbf{39.8ms}&\textbf{0.05m}\\ 
\bottomrule
\end{tabular}

    \vspace{-1.0em}
    \label{tab:other_semseg}
\end{table}

{\parskip=2pt
\noindent\textbf{3D Object Detection}:} 
\if\objdetsec
We present results for object detection on the nuScenes validation set using the fine-tuning approach described in Sec.~\ref{sec:fine-tune}. Other training settings are the same as the baseline UVTR~\citep{li2022uvtr}. In our experiment, we use a larger voxel size of \SI{0.1}{\meter} compared to the more commonly used \SI{0.075}{\meter}. We also do not use test-time augmentation or model ensembling. 

Tab.~\ref{tab:objectdet_sota} shows that our approach achieves 69.7 and 63.7 NDS and mAP, improving by 3.3 NDS points and 2.8 mAP points over the UVTR-L-V0.1 ~\citep{Wu2023PointTV} baseline. Putting NOMAE close to state-of-the-art methods despite the larger voxel size. This showcases the ability of NOMAE to learn a representation suitable for object detection.
\else
We present results for object detection on the nuScenes validation set using the fine-tuning approach described in Sec.~\ref{sec:fine-tune}. We follow the experiment setup of GD-MAE
~\citep{yang2023gd-mae} and UniPAD~\citep{Yang2023UniPADAU}, utilizing only 20\% of the annotated frames during fine-tuning, without the use of CBGS~\citep{zhu2019cbgs} or Copy-and-Paste\citep{yan2018second} augmentation. 
We adopt the same training settings as the baseline UVTR~\citep{li2022uvtr} and do not use test-time augmentation or model ensembling.  

Tab.~\ref{tab:objectdet_ssl} shows that our approach achieves 60.9 and 54.4 NDS and mAP scores respectively, improving by 14.2 in NDS points and 15.4 in mAP points over the UVTR-L~\citep{Wu2023PointTV} baseline, and by 5.1 NDS points and 5.6 mAP over the closest contrastive SSL method Learning-from-2D~\citep{Yang2023UniPADAU}. The significant improvement demonstrates the effectiveness of our proposed localized multi-scale SSL for 3D object detection with limited annotated data. 
\fi

\subsection{Comparison with SSL Methods}
In this experiment, we compare the performance of our proposed method with the self-supervised pretraining methods of Occupancy-MAE~\citep{min2022OccupancyMAESP} and GeoMAE~\citep{tian2023geomae}, with the same encoder architecture of PTv3~\citep{Wu2023PointTV}. 
Tab.~\ref{tab:sota_semseg} shows that our re-implementation of Occupancy-MAE~\citep{min2022OccupancyMAESP} and GeoMAE~\citep{tian2023geomae} with the state-of-the-art PTv3~\citep{Wu2023PointTV} backbones (marked with $^{*}$) outperforms the results reported in the original papers by $0.3$ and $7.1$ mIoU points respectively for semantic segmentation with fine-tuning on the nuScenes dataset. 

Tab.~\ref{tab:other_semseg} shows the results of non-linear probing (NonLP). 
We observe that the NonLP performance in Tab.~\ref{tab:other_semseg} correlates with the fine-tuning results in Tab.~\ref{tab:sota_semseg}. Furthermore, the relative performance improvement of NOMAE over the baselines is higher for NonLP in comparison to fine-tuning, which indicates richer representation learning. Additionally, the time of a single training step (iteration time) is lowest for NOMAE, despite the multi-scale pretraining and a much finer resolution of the pretraining supervision. We note that the iteration time of NOMAE is $10$ms for the pretraining with a single scale.

\begin{table}
    \centering
    \footnotesize
    \caption{Ablation study on the various components in NOMAE for semantic segmentation on the nuScenes validation set. Lines with * are with fine-tuning, and all the other results are with NonLP. For more details refer to Sec.~\ref{sec:incremental_design}}
    \vspace{-0.5em}
%
%
\addtolength{\tabcolsep}{-0.2em}
\begin{tabular}{l c c c }
\toprule
  \textbf{Model} & \textbf{mIoU} & \textbf{mACC} & \textbf{ACC} \\
\midrule
  Occupancy-MAE + PTv3 &59.0&73.4&91.0\\ 
  + reconstruct only $\mathcal{V}_\text{n}$ &66.7&78.2&92.4\\
  + MSP &&&\\
    \qquad naive masking &70.2&82.3&93.5\\
    \qquad Point-M2AE\,\cite{zhang2022pointm2ae} masking&70.5&82.3&93.6\\ 

  + HMG &72.6&83.9&93.7\\
  + $n=9$ &73.3&84.3&94.1\\
  + batch size $8$ &74.8&85.0&94.0\\
  + fine-tuning = noMAE$^{*}$&\textbf{81.8}&\textbf{87.7}&\textbf{94.9}\\
\bottomrule
\end{tabular}

    \vspace{-1.0em}
    \label{tab:incremental}
    
\end{table}

\subsection{Ablation Study}
\label{sec:ablations}
In this section, we present ablation studies on the nuScenes semantic segmentation validation set to investigate the design choices of the proposed method. We performed the experiments using the pre-trained frozen encoder using NonLP. Please refer to Sec.~\ref{sec:nonLP} for further details. 

{\parskip=2pt
\noindent\textbf{Detailed Study of NOMAE}: 
\label{sec:incremental_design}
This experiment evaluates the improvement due to our proposed contributions, and the results are presented in Tab.~\ref{tab:incremental}. 
We start from our implementation of Occupancy-MAE~\citep{min2022OccupancyMAESP} as in Tab.~\ref{tab:other_semseg}. Reconstructing only the local neighborhood $\mathcal{V}_\text{n}$ of visible voxels $\mathcal{V}_\text{v}$ increases the NonLP mIoU from 59.0$\%$ to 66.7$\%$. This requires replacing the Occupancy-MAE decoder with our proposed upsampling module and neighborhood decoder. Adding the multi-scale reconstruction (MSP) from Sec.~\ref{sec:MSP} further improves the mIoU to 70.1 for naive mask construction and to 70.5 for masking strategy from Point-M2AE~\cite{zhang2022pointm2ae}, as opposed to single-scale reconstruction in Occupancy-MAE. Our proposed hierarchical mask generation from Sec.~\ref{sec:HMG} yields an improvement of 72.46 mIoU, as further investigated in Sec.~\ref{sec:ablation_MSP_HMG}. Moreover, increasing the reconstruction neighborhood size from $5$ to $9$ improves the mIoU to $74.8$, as further investigated in Sec.~\ref{sec:ablations_neigh_size}. Reducing the batch size to 8 (further investigated in Sec.~\ref{sec:further_ablaitons} of the supplementary material) yields the final NonLP performance of 74.8$\%$ and fine-tuning performance of 81.8$\%$ mIoU score.}

\addtolength{\tabcolsep}{-0.2em}
\begin{table}
    \centering
    \footnotesize
    \caption{Reconstruction and masking on different single scales, with multi-scale pretext (MSP) and hierarchical mask generation (HMG). 
    Results are reported on the nuScenes val set with the encoder frozen after SSL pretraining, with nonlinear probing. IoU(\textbf{p}) and IoU(\textbf{t}) are the IoU for the classes pedestrian and truck, respectively.}
    \vspace{-0.5em}
    \begin{tabular}{l c c c c c}
    \toprule
     \textbf{Model} & \textbf{mIoU} & \textbf{mACC} & \textbf{ACC} & \textbf{IoU(\textbf{p})} & \textbf{IoU(\textbf{t})} \\
        \midrule
        Single scale $2^s=1$  &63.8&74.4&91.5&72.6&68.8\\
        Single scale $2^s=2$  &66.7&78.2&92.4&75.6&74.3\\
        Single scale $2^s=4$  &68.0&80.0&92.8&76.5&75.5\\
        Single scale $2^s=8$  &68.3&80.5&93.0&75.8&76.2\\
        Single scale $2^s=16$ &67.3&80.3&92.7&73.4&76.6\\
        MSP $2^s\in\{1,2,4,8\}$ &&&&&\\
        \quad naive masking &70.2&82.3&93.5&83.2&79.6\\
        \quad Point-M2AE\,\cite{zhang2022pointm2ae} &70.5&82.3&93.6&82.5&81.2\\
        \quad HMG (ours) &\textbf{72.6}&\textbf{83.9}&\textbf{93.7}&\textbf{85.2}&\textbf{83.6}\\
        \bottomrule
    \end{tabular}
    \vspace{-1em}
    \label{tab:mask_block_size}
\end{table}

{\parskip=2pt
\noindent\textbf{Mask Block Size, MSP and HMG}: 
\label{sec:ablation_MSP_HMG}
This experiment investigates different reconstruction and mask scales $s$ for our SSL task. Tab.~\ref{tab:mask_block_size} compares the performance of reconstructing different {\sl single} scales with the proposed multi-scale pretext (MSP) from Sec.~\ref{sec:MSP}. The experiment of MSP without hierarchical mask generation (HMG) uses either a random mask at the finest scale, which is pooled to generate masks of coarser scales, as described in the naive solution in Sec.~\ref{sec:HMG}, or the method of Point-M2AE~\cite{zhang2022pointm2ae}.}

It can be observed that no single-scale occupancy task is suitable for all object types. For example, trucks benefit from a coarser occupancy reconstruction, while pedestrians prefer a finer resolution in the pretraining task, except for the very fine scales of $2^s\in\{1,2\}$. MSP combines coarse and fine tasks, thereby maximizing the overall mIoU. We observe that HMG achieves an additional improvement of 2.4 and 2.1$\%$ mIoU over random mask at the finest scale and the mask generation proposed in Point-M2AE~\citep{zhang2022pointm2ae} respectively. This underlines the importance of proper training examples at all scales.

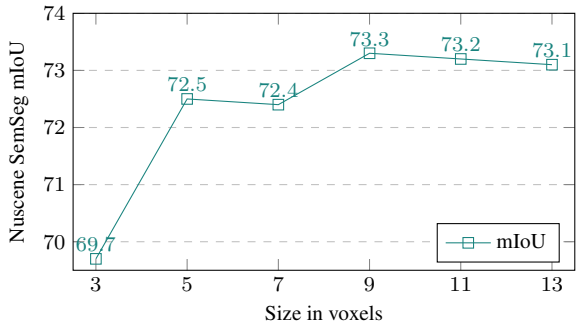
\begin{figure}
    \centering
    \footnotesize
    \begin{tikzpicture}
        \begin{axis}[
            xlabel={Size in voxels},
            ylabel={Nuscene SemSeg mIoU},
            xmin=2.5, xmax=13.5,
            ymin=69.5, ymax=74,
            xtick={3,5,7,9,11,13},
            ytick={68,69,70,71,72,73,74},
            legend pos=south east,
            ymajorgrids=true,
            width=0.99\columnwidth,
            height=0.6\columnwidth,
            grid style=dashed,
             nodes near coords={%
            \footnotesize
            $
            \pgfmathprintnumber
            {\pgfkeysvalueof{/data point/y}}$%
        },
        ]

            \addplot[color=bosch_turquoise_dark,mark=square,]
                coordinates {
                (3,69.7)(5,72.5)(7,72.4)(9,73.3)(11,73.2)(13,73.1)
                };
            \legend{mIoU}
        \end{axis}
    \end{tikzpicture}
    \caption{Size (number of voxels) of the reconstructed neighborhood $n$ around visible voxels $\mathcal{V}_\text{v}$, to create $\mathcal{V}_\text{n}$ in the proposed pretext task. We observe that the downstream NonLP semantic segmentation peaks at $n=9$. Note that $n\rightarrow\infty$ corresponds to the method of~\cite{min2022OccupancyMAESP}.}
    \label{fig:neigh_size}
\end{figure}

{\parskip=2pt
\noindent\textbf{Neighborhood Size}:}
\label{sec:ablations_neigh_size}
Fig.~\ref{fig:neigh_size} investigates the effect of the neighborhood size (number of voxels) to generate $\mathcal{V}_\text{n}$ from $\mathcal{V}_\text{v}$. For example, $n=5$ indicates that a total of 5 voxels are covered in all 3 (x,y,z) dimensions for every scale. Experiments use MSP and HMG. Since every scale has a different voxel size, the reconstructed volume depends on the scale. We observe a maximum of the NonLP performance for $n=9$. 
Our interpretation is that a smaller neighborhood does not cover sufficient LiDAR measurements for reconstruction, while a larger neighborhood hinders local representations in the pretraining. Further, we observe that limiting the reconstruction size performs consistently better than reconstructing the whole space, which would correspond to the method of Occupancy-MAE~\citep{min2022OccupancyMAESP}.
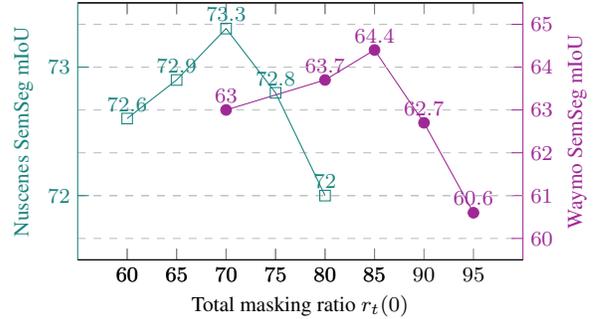
\begin{figure}
    \centering
    \footnotesize
    \vspace{-0.5em}
    \begin{tikzpicture}
        \begin{axis}[
                width=0.9\columnwidth,
                height=0.6\columnwidth,
                axis y line*=right,
                ylabel={Waymo SemSeg mIoU},
                y axis line style = bosch_purple_dark,
                y tick style = bosch_purple_dark,
                yticklabel style = bosch_purple_dark,
                ylabel style = bosch_purple_dark,
                xmin=55, xmax=100,
                ymin=59.5, ymax=65.5,
                xtick={60,65,70,75,80,85},
                ytick={59,60,61,62,63,64,65},
                legend pos=north east,
                ymajorgrids=true,
                grid style=dashed,
                 nodes near coords={%
                \footnotesize
                $
                \pgfmathprintnumber
                {\pgfkeysvalueof{/data point/y}}$%
            },
            ]

            \addplot[color=bosch_purple_dark,mark=*]
                coordinates {
                (70,63.0)(80,63.7)(85,64.4)(90,62.7)(95,60.6)
                };
        \end{axis}
        \begin{axis}[
            width=0.9\columnwidth,
            height=0.6\columnwidth,
            axis y line*=left,
            xlabel={Total masking ratio $r_t(0)$},
            ylabel={Nuscenes SemSeg mIoU},
            xmin=55, xmax=100,
            ymin=71.5, ymax=73.5,
            xtick={60,65,70,75,80,85,90,95},
            ytick={68,69,70,71,72,73,74},
            y axis line style = bosch_turquoise_dark,
            y tick style = bosch_turquoise_dark,
            yticklabel style = bosch_turquoise_dark,
            ylabel style = bosch_turquoise_dark,
            legend pos= north east,
            ymajorgrids=true,
            grid style=dashed,
             nodes near coords={%
            \footnotesize
            $
            \pgfmathprintnumber
            {\pgfkeysvalueof{/data point/y}}$%
        },
        ]

            \addplot[color=bosch_turquoise_dark,mark=square,]
                coordinates {
                (60,72.6)(65,72.9)(70,73.3)(75,72.8)(80,72.0)
                };
        \end{axis}
    \end{tikzpicture}
    \caption{NonLP performance over the masking ratio $r_\text{t}$ on the nuScenes and Waymo datasets. We observe that the optimal $r_\text{t}(0)$ is $70\%$ for the nuScenes and $85\%$ for the Waymo Open Dataset. Our interpretation is that Waymo requires a higher masking ratio due to the higher density of the LiDAR point cloud.}
    \vspace{-1em}
\label{fig:masking_ratio}
\end{figure}
\begin{figure*}
\centering
\footnotesize
\begin{tabular}{cccc}
&Baseline Output& NOMAE Output& Improvement/Error Map\\
(a) &\subfloat{\includegraphics[trim={2cm 5cm 2cm 5cm},clip,width=0.22\textwidth]{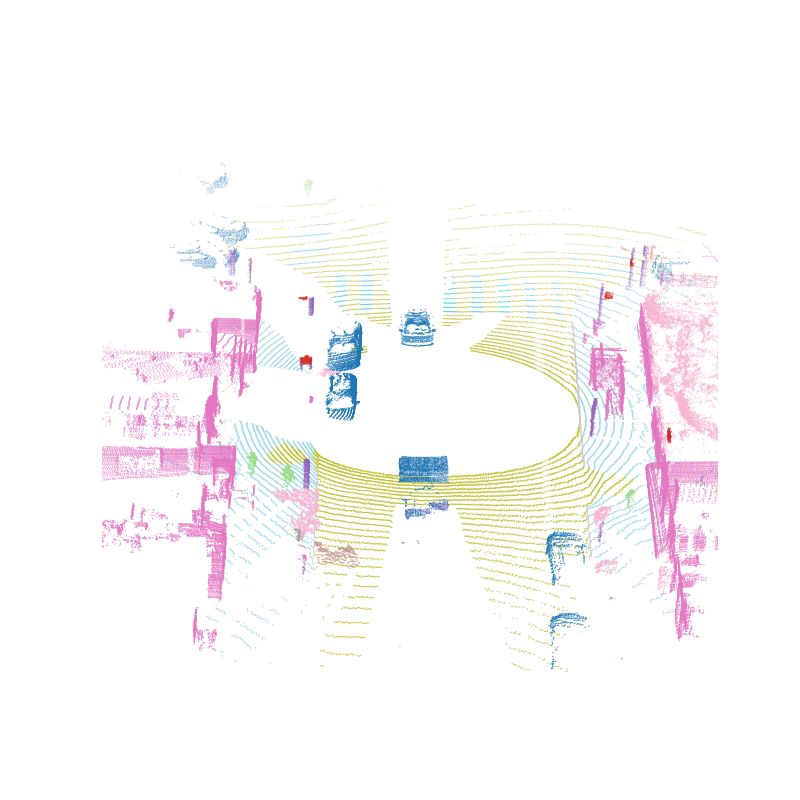}}
&\subfloat{\includegraphics[trim={2cm 5cm 2cm 5cm},clip,width=0.22\textwidth]{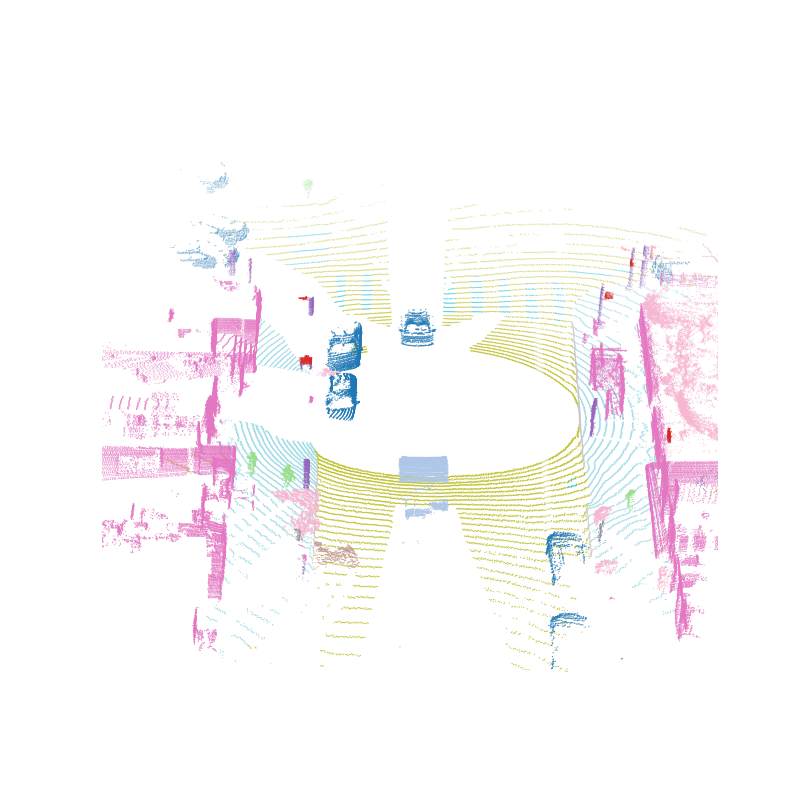}}
&\subfloat{\includegraphics[trim={2cm 5cm 2cm 5cm},clip,width=0.22\textwidth]{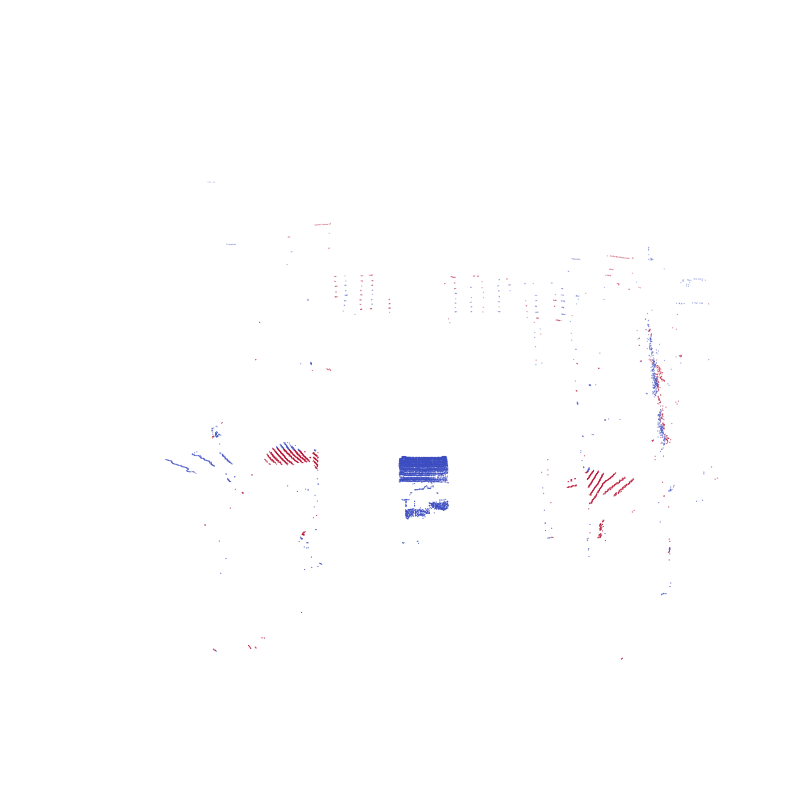}}
\\
\midrule
(b) &\subfloat{\includegraphics[trim={2cm 5cm 2cm 5cm},clip,width=0.22\textwidth]{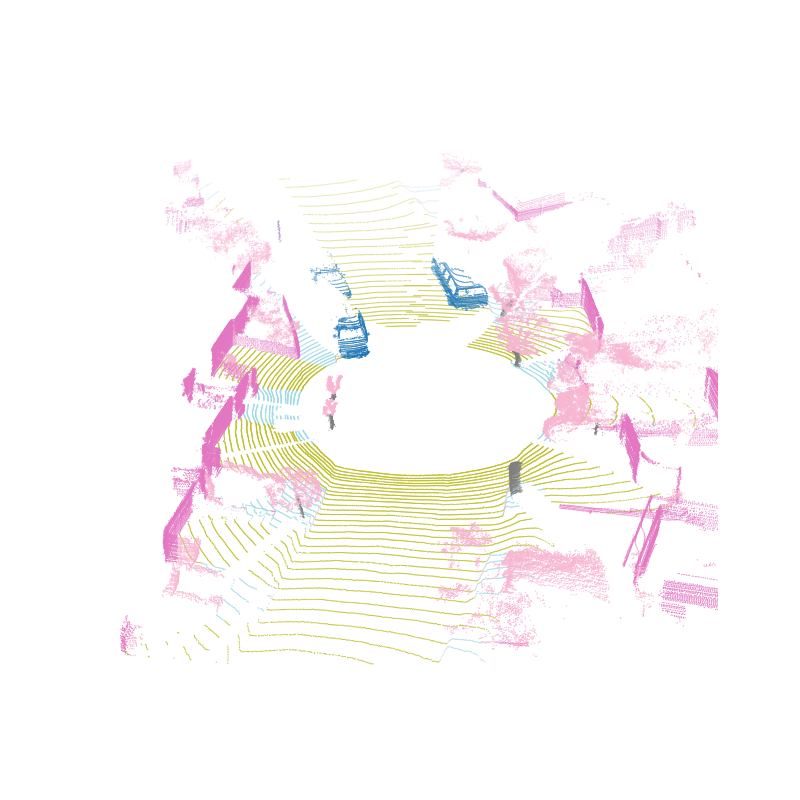}}
&\subfloat{\includegraphics[trim={2cm 5cm 2cm 5cm},clip,width=0.22\textwidth]{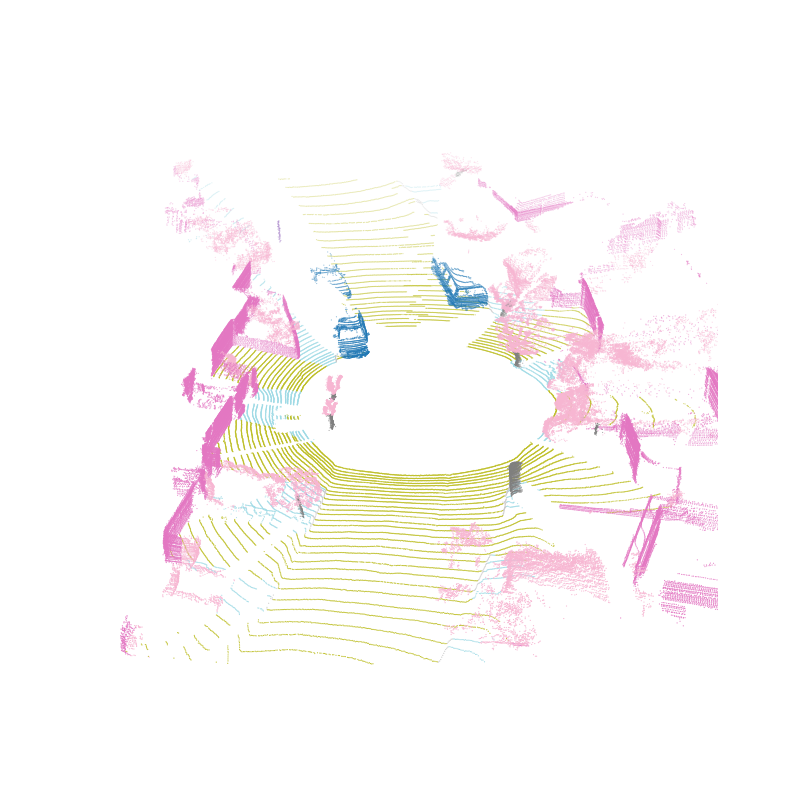}}
&\subfloat{\includegraphics[trim={2cm 5cm 2cm 5cm},clip,width=0.22\textwidth]{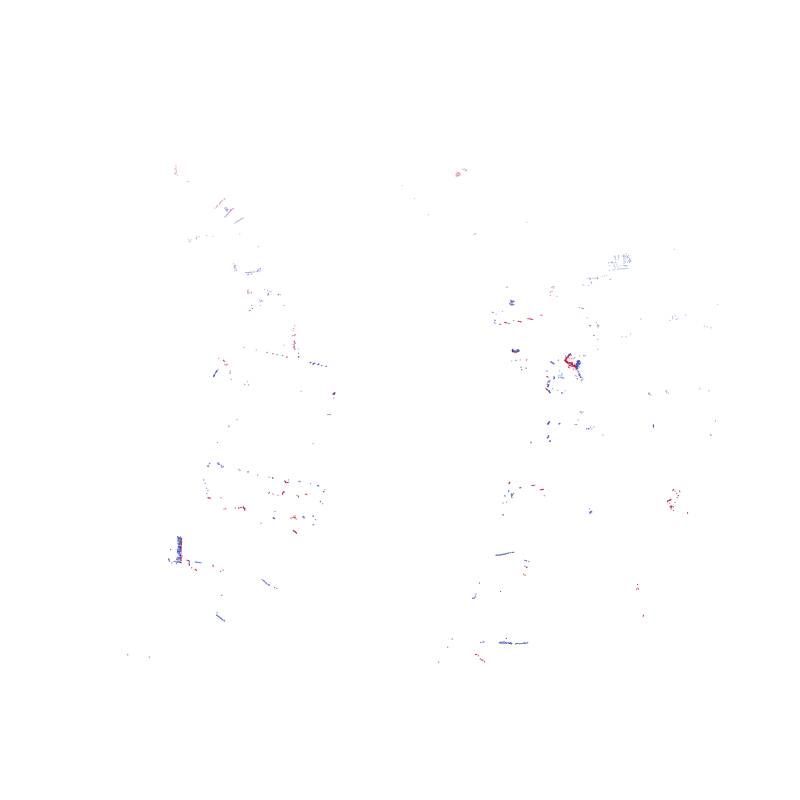}}
\\
\midrule
&Baseline Output& NOMAE Output& Ground Truth\\
(c) &\subfloat{\includegraphics[trim={2cm 35cm 2cm 35cm},clip,width=0.25\textwidth]{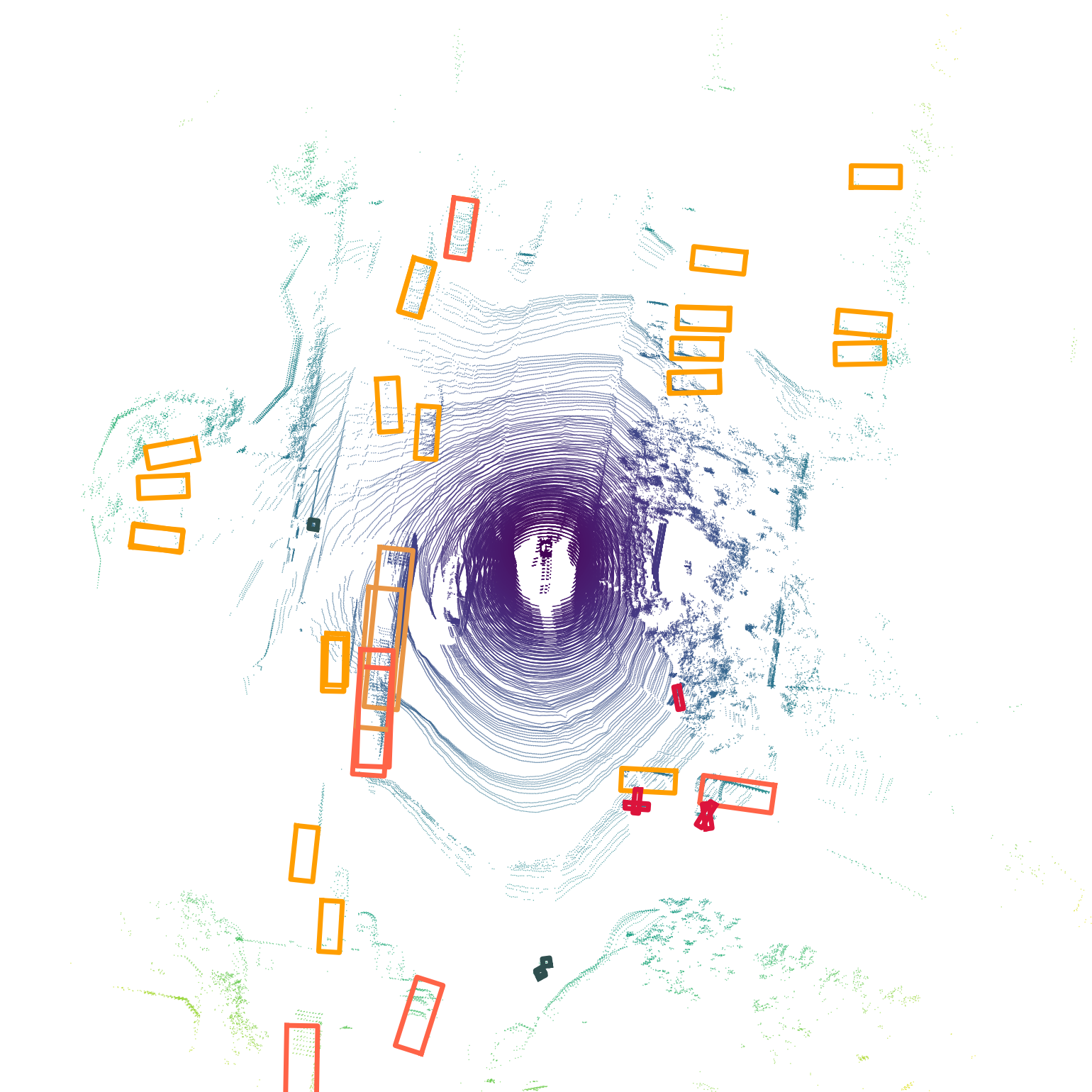}}
&\subfloat{\includegraphics[trim={2cm 35cm 2cm 35cm},clip,width=0.25\textwidth]{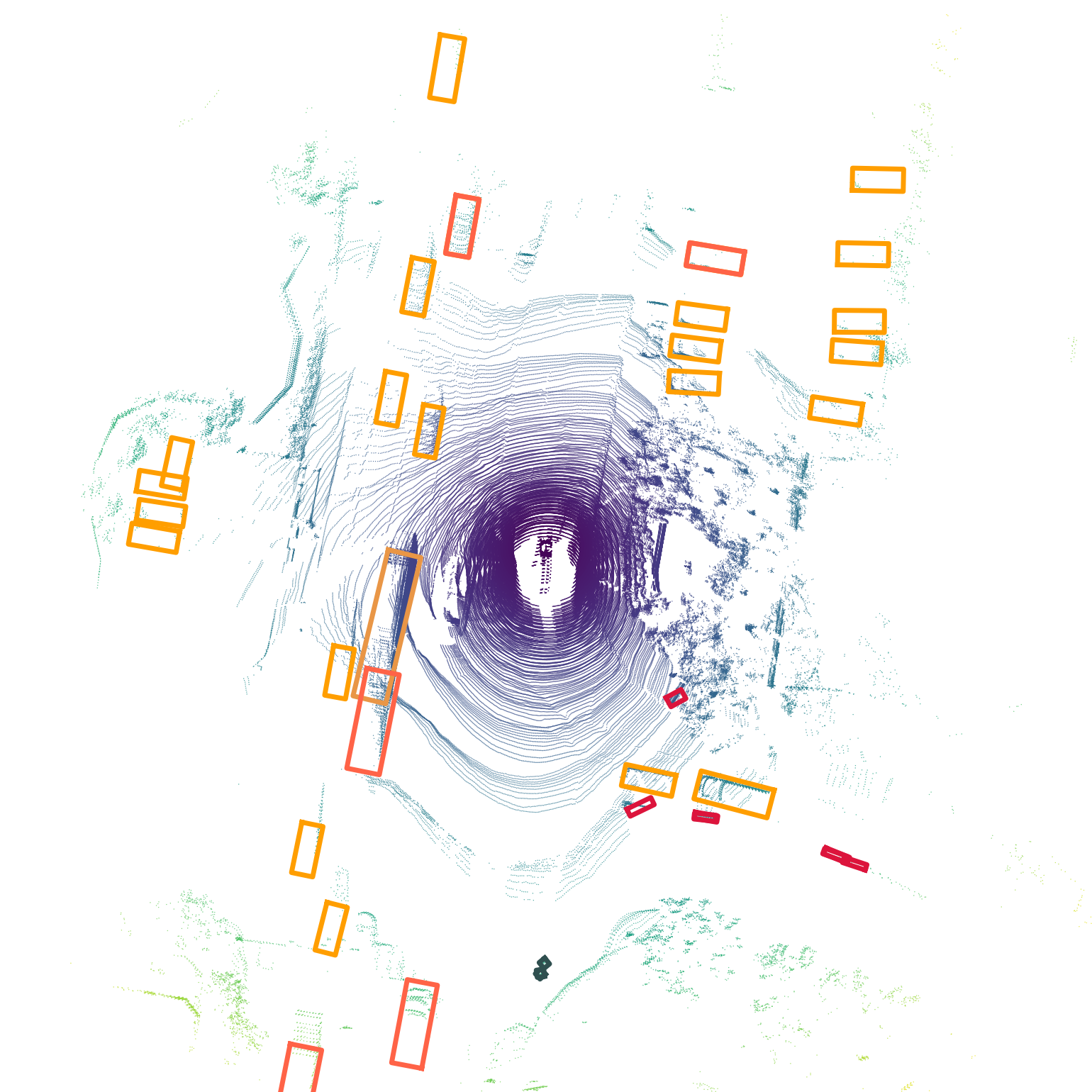}}
&\subfloat{\includegraphics[trim={2cm 35cm 2cm 35cm},clip,width=0.25\textwidth]{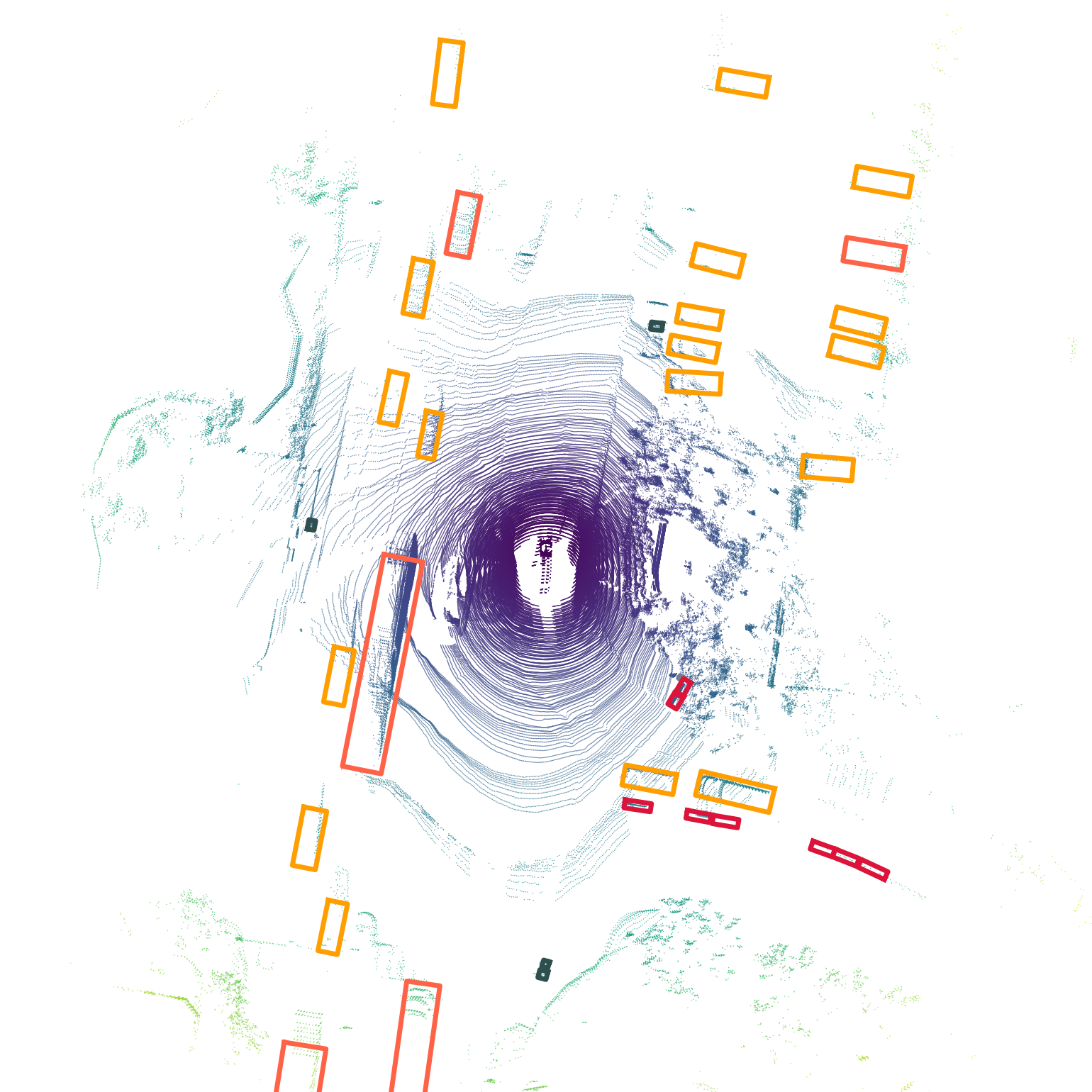}}
\end{tabular}
\caption{Qualitative comparison of semantic segmentation performance with Ptv3~\citep{Wu2023PointTV} and object detection performance with UVTR~\citep{li2022uvtr}.}
\label{fig:qualitative}
\vspace{-0.5em}
\end{figure*}

{\parskip=2pt
\noindent\textbf{Masking Ratio}: 
\label{sec:masking_ratio}
Fig.~\ref{fig:masking_ratio} investigates the effect of different masking ratios on the quality of the representations for semantic segmentation on the Waymo and nuScenes datasets. This experiment uses a constant $r(s)=r$ to achieve the total masking ratio $r_\text{t}$ according to Eq.~\ref{equ:total_masking_ratio}. 
We observe that the optimal total masking ratio is $70\%$ for the nuScenes and $85\%$ for the Waymo Open dataset. This is intuitive since the point cloud density in Waymo (ca. $180$k points per frame) is higher than the relatively sparse nuScenes (ca. $34$k points per frame) LiDAR point clouds.}

{\parskip=2pt
\noindent\textbf{Data Efficient nuScenes}: 
\label{sec:data_efficient_nuscenes}
Tab.~\ref{tab:model_efficiency} analyzes the performance under limited annotated data for semantic segmentation on the nuScenes dataset. The sub-sampling of the data is performed sequence-wise, meaning that all frames of a sequence are either included or excluded. For example, $0.1\%$ indicates that only one of the $1000$ scenes of the nuScenes dataset is used, namely \textsl{scene-0392}. 
In every experiment, all sequences of the training set are used for the self-supervised pertaining. We observe that NOMAE (fine-tune) consistently outperforms training from scratch, which demonstrates the method's ability to benefit from unannotated data. Furthermore, we observe that in experiments with very little data, NonLP performs similarly to fine-tuning. This suggests that NOMAE can benefit from the unannotated data to learn a sufficiently strong representation in the encoder, such that the fine-tuning of the encoder has little benefit.}

\begin{table}
    \centering
    \footnotesize
        \caption{Results with varying amounts of annotated data, evaluated for semantic segmentation mIoU on the nuScenes val set.}        \addtolength{\tabcolsep}{-0.1em}
        \vspace{-0.5em}
        \begin{tabular}{lcccccc}\toprule
           \textbf{Method} &\multicolumn{5}{c}{\textbf{Annotated Scenes}} \\\cmidrule(lr){2-6}
             & \textbf{0.1\%} & \textbf{1\%} & \textbf{10\%} & \textbf{50\%} & \textbf{100\%} \\\midrule
            PTv3\,\citep{Wu2023PointTV} &28.2 &41.6 &68.7 &78.7&80.4 \\
            NOMAE (NonLP) &35.7 &47.5 &67.7 &73.8&74.2  \\
            NOMAE (fine-tune) &\textbf{35.8} &\textbf{48.1} &\textbf{69.9} &\textbf{80.1}&\textbf{81.8} \\
            \bottomrule
        \end{tabular}
        \vspace{-1em}
        \label{tab:model_efficiency}
\end{table}



\subsection{Qualitative Evaluations}
We present qualitative comparisons in Fig.~\ref{fig:qualitative}. In (a) and (b) for semantic segmentation, we observe that NOMAE improves the accuracy by reducing class mix-up and by improving the boundaries between objects. In (a), we observe that the baseline mis-segments the truck while NOMAE accurately recognizes it. In (b), we see that NOMAE recognizes the smaller object missed by the baseline (the bottom left pole). We can also see that it fails to recognize some drivable areas in (a). In Example (c), we visualize the detections from NOMAE. We see that compared to the baseline, NOMAE is able to more accurately estimate the orientation of the objects and has higher true positive detections. We also see both models hallucinating in further away regions (on the left), and NOMAE fails to detect the pedestrian on the left. More qualitative results are presented in Sec.~\ref{sec:further_qualitative} of the supplementary material.

\section{Conclusion}
\label{sec:conclusion}

In this work, we proposed NOMAE, a novel multi-scale self-supervised learning framework for large-scale point clouds. Observing the large-scale nature of LiDAR point clouds, NOMAE reconstructs only local neighborhoods, keeping the computation tractable at higher voxel resolutions, avoiding information leakage, and learning a localized representation suitable for diverse downstream perception tasks. Enabled by its efficiency, NOMAE utilizes multiple scales in the pre-training, enabling the model to learn both coarse and fine representations. A novel hierarchical mask generation scheme balances the pre-training of coarse and fine features, which is important for objects of different sizes, such as pedestrians and trucks
We presented experimental results that underline the benefit of our proposed contributions, achieving state-of-the-art performance on multiple benchmarks.

\noindent\textbf{Limitations}: NOMAE is sensitive to the density of the point cloud and future work will investigate the proposed method for sparse 3D sensors such as radar. 
Additionally, NOMAE does not utilize the temporal nature of LiDAR data which can open the door for further performance improvement. 
Furthermore, the application of high-resolution sparse 3D representations in the encoders should be further investigated for the object detection task, where 2D bird's eye view and low-resolution 3D approaches are still dominant.

{
    \small 
    \bibliographystyle{ieeenat_fullname}
    \bibliography{main}
}

\clearpage
\maketitlesupplementary

\section{Implementation Details}
\label{sec:imp_details}
In this section, we report further details of our implementation for both pretraining and finetuning. 
We use the Pointcept~\citep{pointcept2023} framework for pretraining and finetuning semantic segmentation. 
We performed the fine-tuning for object detection in the MMDetection3D framework~\citep{mmdet3d2020}.
\subsection{Training Settings}
\begin{table}
\centering
\footnotesize
\caption{Training settings for semantic segmentation NOMAE pretraining and fine-tuning}
\begin{tabular}{lclc}\toprule
\multicolumn{2}{c}{Pretraining} &\multicolumn{2}{c}{SemSeg~\cite{wu2023ppt}} \\
\cmidrule(lr){1-2} \cmidrule(lr){3-4}
Config &Value &Config &Value \\\midrule
optimizer &AdamW &optimizer &AdamW \\
scheduler &Cosine &scheduler &Cosine \\
criteria &BCE~(1) &criteria &CE~(1) \\
&& &Lovasz~(1) \\
learning rate &2e-3 &learning rate &2e-3 \\
block lr scaler &0.1 &block lr scaler &0.1 \\
weight decay &5e-2 &weight decay &5e-2 \\
batch size &8 &batch size &12 \\
Datasets & nuScenes /  & Datasets & nuScenes / \\
         & Waymo       &          & Waymo \\
Masking Ratio & 0.7 / 0.85 & Masking Ratio & -~/~- \\
HMG & $[5,10,20,40]\text{cm}$ &HMG & - \\
LLRD & - &LLRD & 0.65 \\
warmup epochs &2 &warmup epochs &2 \\
epochs &50 &epochs &50 \\
\bottomrule
\end{tabular}
\label{tab:appendix_semseg_settings}
\end{table}

In Tab.~\ref{tab:appendix_semseg_settings}, we report the training details for both pretraining and semantic segmentation finetuning. For the criteria, the weights between brackets represent the weight of the loss in the final objective. During finetuning, we employ LLRD~\cite{he2022mae}, more specifically, we decay the learning rate of every block in the encoder exponentially with a factor of $0.65$ compared to the next block. For the downsampling blocks, we utilize the same learning rate as the next transformer block in the architecture. 

\begin{table}
\centering
\footnotesize
\caption{Training settings for object detection NOMAE pretraining and finetuning}
\begin{tabular}{lclc}\toprule
\multicolumn{2}{c}{Pretraining} &\multicolumn{2}{c}{ObjDet~\cite{li2022uvtr}} \\
\cmidrule(lr){1-2} \cmidrule(lr){3-4}
Config &Value &Config &Value \\\midrule
optimizer &AdamW &optimizer &AdamW \\
scheduler &Cosine &scheduler &Cosine \\
criteria &BCE~(1) &criteria &Focal loss~(2) \\
&& &L1 loss~(0.25) \\
learning rate &2e-3 &learning rate &1e-3 \\
block lr scaler &0.1 &block lr scaler &0.2 \\
weight decay &5e-2 &weight decay &1e-2 \\
batch size &8 &batch size &16 \\
Datasets & nuScenes  & Datasets & nuScenes \\
Size & 100\% & & 20\% \\ 
Masking Ratio & 0.85 & Masking Ratio & - \\
HMG & $[5,10,20,40]cm$ &HMG & - \\
LLRD & - &LLRD & 0.65 \\
warmup epochs &2 &warmup epochs &8 \\
epochs &50 &epochs &20 \\
\bottomrule
\end{tabular}
\label{tab:appendix_objdet_settings}
\end{table}

In Tab.~\ref{tab:appendix_objdet_settings}, we highlight the pretraining and finetuning details for object detection experiments. We follow the standard 10 sweeps aggregation for nuScenes object detection~\citep{li2022uvtr}. To this end, we pretrain NOMAE on the aggregated point clouds and use a masking ratio of 85\% to accommodate for the denser point cloud. For other training settings, we utilize the same implementation details as our object detection baseline UVTR~\citep{li2022uvtr}.
\label{sec:further_ablaitons}
\subsection{Model Settings}
For the encoder $\mathbf{E}$, we use the same implementation details as the Encoder of PTv3~\citep{Wu2023PointTV}. For the upsampling module $\mathbf{M}_u$, we use a PTv3 decoder as a baseline and experiment with different sizes. We found empirically that reducing the size of the upsampling module improves the downstream performance after finetuning. Hence, we utilize a single transformer block per resolution in the upsampling module.

The neighboring decoder uses sparse convolution layers to generate representations for neighboring voxels $\mathcal{V}_n$ and a single sub-manifold convolution layer to convert the representation into occupancy predictions. 
Sec.~\ref{sec:neigh_dec_ablations} gives further details of the hyperparameter choices of the neighborhood decoder. 
\begin{table*}
\centering
\footnotesize
\centering
\caption{Detailed results on NuScenes val and test set compared with the baseline PTv3~\citep{Wu2023PointTV}.}
\tabcolsep 3pt
\begin{tabular}{llcccccccccccccccccc}
\toprule
\textbf{Dataset} &\textbf{Method}                       & \textbf{mIoU} & \textbf{fwIoU} & \rotatebox{90}{\textbf{barrier}} & \rotatebox{90}{\textbf{bicycle}} & \rotatebox{90}{\textbf{bus}} & \rotatebox{90}{\textbf{car}} & \rotatebox{90}{\textbf{construction}} & \rotatebox{90}{\textbf{motocycle}} & \rotatebox{90}{\textbf{pedstrain}} & \rotatebox{90}{\textbf{Traffic cone}} & \rotatebox{90}{\textbf{trailer}} & \rotatebox{90}{\textbf{truck}} & \rotatebox{90}{\textbf{drivable}} & \rotatebox{90}{\textbf{Other flat}} & \rotatebox{90}{\textbf{sidewalk}} & \rotatebox{90}{\textbf{terrian}} & \rotatebox{90}{\textbf{manmade}} & \rotatebox{90}{\textbf{vegetation}} \\
\midrule
\multirow{2}{*}{\textbf{Val}} & {PTv3(Baseline)} & 80.4   & -        & \textbf{81}             & 54             & 96         & 92         & 52                  & 89               & 84               & 72                  & \textbf{74}             & 85           & \textbf{97}                & \textbf{76}                & \textbf{77}              & 76               & 91               & 90                \\
& {NOMAE (Ours)}  & \textbf{81.8} & -  & \textbf{81} & \textbf{56} & \textbf{97}  & \textbf{95} & \textbf{63}  & \textbf{90}  & \textbf{85}  & \textbf{73}              & \textbf{74}                 & \textbf{88}             & \textbf{97}             & 75              & \textbf{77}              & \textbf{77}             & \textbf{92}  & \textbf{90}                             \\
\midrule
\multirow{2}{*}{\textbf{Test}} &{PTv3(Baseline)} & \textbf{82.7} & 91.1         & 83             & \textbf{72}             & \textbf{93}         & 92         & \textbf{71}                  & \textbf{90}              & 83              & 77                  & 86             & 75           & \textbf{98}                & \textbf{69}                & 81              & 77               & 92              & 89                \\
& {NOMAE (Ours)}  & 82.6  & \textbf{91.5} & \textbf{87} & 50 & 92  & \textbf{93} & \textbf{71}  & \textbf{90}  & \textbf{86}  & \textbf{82}               & \textbf{89}                  & \textbf{77}             & \textbf{98}             & \textbf{69}             & \textbf{83}                & \textbf{78}             & \textbf{93}            & \textbf{90}                               \\
\bottomrule
\end{tabular}
\label{tab:appendix_classwise_results}
\end{table*}
\subsection{Data Augmentations}
\begin{table}
\centering
\footnotesize
\caption{Data augmentation details for NOMAE pretraining and finetuning for both semantic segmentation and object detection.}
\begin{tabular}{llccc}\toprule
Augmentations &Parameters &Pretrain &SemSeg &ObjDet \\\midrule
random rotate &\{axis: z,  p: 0.5, &\checkmark &\checkmark& - \\
&angle: [-1, 1]$\pi$\} &&&\\
random rotate &\{axis: z, p: 1.0, &- & -&\checkmark \\
&angle: [-1 / 8, 1 / 8]$\pi$\} & && \\
random scale &scale: [0.9, 1.1] &\checkmark &\checkmark&- \\
random scale &scale: [0.95, 1.05] &- & -&\checkmark \\
random flip &p: 0.5 &\checkmark &\checkmark&\checkmark \\
random jitter &sigma: 0.005, &\checkmark &\checkmark&- \\
&clip: 0.02 & && \\
point clip&range: [-51.2,-51.2,-5, &- & -&\checkmark \\
&51.2,51.2,3]m &&& \\
\bottomrule
\end{tabular}
\label{tab:appendix_data_augmentation}
\end{table}
We adopt common data augmentation methods used in~\citep{Wu2023PointTV,li2022uvtr} during fine-tuning semantic segmentation and object detection. 
Pretraining uses the same data augmentations as semantic segmentation finetuning. 
Details about the data augmentations used are reported in Tab.~\ref{tab:appendix_data_augmentation}.

\section{Further Ablations}
In this section, we present further ablation studies on the nuScenes semantic segmentation validation set. 
We evaluate the settings by freezing weights in the encoder after pretraining and using NonLP as described in Sec.~\ref{sec:nonLP}.

\begin{table}
    \centering
    \footnotesize
    \caption{Results with varying batch size during NOMAE pretraining step.}
        \addtolength{\tabcolsep}{0.25em}
        \vspace{-0.5em}
        \begin{tabular}{lcccccc}\toprule
           &\multicolumn{6}{c}{\textbf{Batch Size}} \\\cmidrule(lr){2-7}
             & \textbf{4} & \textbf{8} & \textbf{16} & \textbf{32} & \textbf{64} & \textbf{128} \\\midrule
            Single Scale $2^s = 1$ &- &- &64.9&64.3&63.8 &61.9 \\
            NOMAE &73.8 &\textbf{74.8} &73.5&73.3& - & -  \\
            \bottomrule
        \end{tabular}
        \label{tab:appendix_batch_size}
\end{table}

{\parskip=2pt
\noindent\textbf{Pretraining Batch Size}: 
\label{sec:ablations_batch_size}
This experiment investigates the effect of different batch sizes for our SSL task. Tab.~\ref{tab:appendix_batch_size} compares the downstream performance of models pretrained with different batch sizes and the same number of epochs (leading to a different number of parameter update steps). It can be observed that for NOMAE, the performance increases with decreasing the batch size (even for a single scale pretraining), and it reaches a maximum of 8 examples per batch before further decreasing. We attribute the higher performance at lower batch sizes to an increased number of parameter update steps. The results also indicate that NOMAE can benefit from a longer training schedule.}

{\parskip=2pt
\noindent\textbf{Neighboring Decoder Design}: 
\label{sec:neigh_dec_ablations}
\begin{figure}
  \centering
  \begin{subfigure}{\linewidth}
    \centering

    \includegraphics[trim={5cm 15.8cm 6.5cm 8.5cm},clip,width=1.0\textwidth]{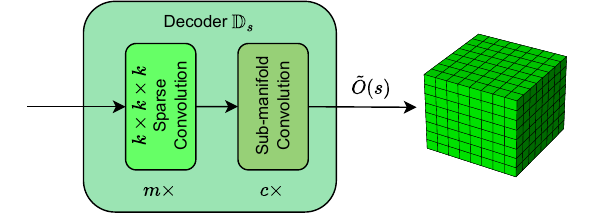}
  \end{subfigure}
  \caption{General architecture of a Neighborhood occupancy decoder $\mathbb{D}_s$ at scale s.}
  \label{fig:appendix_neigh_dec}
\end{figure}
This experiment investigates the effect of design choices for the neighboring decoder. Fig.~\ref{fig:appendix_neigh_dec} shows the general architecture of the neighboring decoder. 
It consists of multiple sparse convolution layers to yield representations for $\mathcal{V}_\text{n}$ from $\mathcal{V}_\text{v}$, followed by sub-manifold convolution layers to predict the occupancy from the representation. 

First, we vary both the network depth $m$ and the size of the neighborhood expansion per layer $k$ (layer receptive field). 
The total size of the neighborhood is given by
\begin{equation}
    n=2m(k-1) + 1.
\end{equation}
Tab.~\ref{tab:appendix_neigh_dec} shows that a shallower decoder helps to learn a better representation compared to a deeper one with the same neighborhood size. 

Furthermore, we investigate the number of sparse submanifold convolution layers $c$ in the decoder. 
$c=0$ indicates that the sparse convolutions must directly predict the neighboring occupancy $\tilde{\mathcal{O}}(s)$. 
Tab.~\ref{tab:appendix_neigh_dec_head} shows that having at least one sparse submanifold convolution layer improves the downstream performance.
Adding more layers performs on par with the single layer despite the extra computation cost.
}
\subsection{Detailed Results}
In this section, we report detailed results for nuScenes semantic segmentation on the test set, in comparison to the validation set.
In particular, we examine the influence of the class \textit{bicycle} on the overall performance and its impact on the test mIoU score. 

In Tab.~\ref{tab:appendix_classwise_results}, we observe that NOMAE outperforms the previous baseline PTv3~\citep{Wu2023PointTV} on almost all classes. 
The performance of the \textit{bicycle} class is lower than for other classes for both PTv3 and NOMAE. 
We attribute this to \textit{bicycle} being a rare class in the nuScenes dataset. 
On the test set, however, state-of-the-art approaches achieve relatively high scores for the \textit{bicycle} class, thereby optimizing the mIoU score, which averages class-wise results irrespectively of the object's frequency in the dataset. We did not take any special measures to address this for minority classes during fine-tuning NOMAE. 
As a consequence, the final mIoU score of NOMAE is on par but slightly lower than the best leaderboard submission, while achieving similar or better performance than PTv3~\citep{Wu2023PointTV} on the other classes. 
As a result, NOMAE sets the new state-of-the-art regarding frequency-weighted IoU (fwIoU) on the test set. 

\begin{table}
    \centering
    \footnotesize
    \caption{Results with varying $k$ and $m$ in the neighborhood decoders during NOMAE pretraining.}
        \addtolength{\tabcolsep}{0.25em}
        \vspace{-0.5em}
        \begin{tabular}{lcccccc}\toprule
             & \textbf{m1k2} & \textbf{m2k2} & \textbf{m3k2} & \textbf{m4k2} & \textbf{m1k3} & \textbf{m2k3} \\
             & \textbf{n=3} & \textbf{n=5} & \textbf{n=7} & \textbf{n=9} & \textbf{n=5} & \textbf{n=9} \\\midrule
            mIoU &69.7 &71.2 &72.4&72.8 &72.5 &73.3 \\
            \bottomrule
        \end{tabular}
        \label{tab:appendix_neigh_dec}
\end{table}

\begin{table}
    \centering
    \footnotesize
    \caption{Results with different neighborhood decoder head depth $c$ during NOMAE pretraining.}
    \addtolength{\tabcolsep}{0.25em}
    \vspace{-0.5em}
    \begin{tabular}{lccc}\toprule
             & \textbf{c=0} & \textbf{c=1} & \textbf{c=2} \\\midrule
            mIoU &72.7 &73.3 &73.2 \\
            \bottomrule
        \end{tabular}
        \label{tab:appendix_neigh_dec_head}
\end{table}

\section{Generalization to Different Architectures}
\label{sec:generalization}
In this section, we present more quantitative results on the Nuscenes Semseg task for pretraining and finetuning different architecutres (MinkUnet\,\cite{ChoyGS2019mink}, OACNN\,\cite{Peng2024OACNNsOS} and OctFormer\,\cite{Wang2023OctFormerOT}). We choose another transformer based architecture and two different CNN based ones. For fair comparison with the older architectures we reproduce their results for training from scratch using the modern optimizers and schedulers which already improves their performance by whopping 5.3 points. For the modern ones, the results are consistent with the original works. In Tab.~\ref{tab:appendix_different_architectures} we can see that NOMAE consistently improves the downstream performance by $\sim 1.4-2.2$ mIoU points compared to training from scratch. It's worth mentioning that such improvement comes without tuning any of the pretraining hyperparameters, demonstrating the superb generalization capabilities of NOMAE pretraining. We hypothesize that the generalization capabilities arises from the diverse feature learning promoted by our MSP (see Sec.~\ref{sec:MSP})

\begin{table}[h]
    \centering
    \footnotesize
    \vspace{-0.6em}
    \begin{tabular}{lcccccc}
    \toprule
    \textbf{Model} & \multicolumn{2}{c}{Scratch(paper)} & \multicolumn{2}{c}{Scratch(ours)} & \multicolumn{2}{c}{+ NOMAE}\\
    & \textbf{mIoU} & \textbf{mACC} & \textbf{mIoU} & \textbf{mACC}& \textbf{mIoU} & \textbf{mACC} \\
    \midrule
      MinkUnet\,\cite{ChoyGS2019mink} &73.3&-&78.6&83.9&80.1&86.2\\ 
      OACNNs\,\cite{Peng2024OACNNsOS}   &78.9&-&78.8&85.8&80.9&86.4\\ 
      Octformer\,\cite{Wang2023OctFormerOT} &-&-&79.4&87.0&81.6&\textbf{88.8}\\ 
      PTv3\cite{Wu2023PointTV} &80.4&-&80.4&87.3&\textbf{81.8}&87.7\\ 
    \bottomrule
    \end{tabular}
    \caption{Results of pretraining and finetuning different architectures on Nuscenes Semseg val set compared to training from scratch.}
    \label{tab:appendix_different_architectures}
    \vspace{-1.1em}
\end{table}

\section{Multi-Scale SSL}
In this section we compare between our formulation of multi-scale SSL (assigning different granularity reconstruction targets to different feature levels at different scales, and the formulation of multi-scale SSL in GeoMAE\,\cite{tian2023geomae} (assigning different granularity reconstruction targets to a single feature level). Intuitively, our formulation encourages feature diversity between different feature levels. While GeoMAE's formulation improves the semantics of a single layer level, yet it doesn't promote feature diversity between different layers and can make said layer a bottleneck. In Tab\,\ref{tab:appendix_msp_vs_geomae} we quantify the improvement from each formulation using neighborhood occupancy as the pretext. The results show that the gain from MSP is $\sim3\times$ GeoMAE’s multi-scale (MS) on nuScenes Semseg val set. This demonstrates the importance of assigning different tasks to different feature levels for an effective SSL on large-scale point clouds.

\begin{table}[h]
    \centering
    \footnotesize
    \begin{tabular}{cccc}
\toprule
        &SS $2^s=8$ NOMAE& +GEO-MAE's MS& +our MSP\\
        \midrule
        NonLP mIoU&68.3 & 69.7 & \textbf{72.6}\\
        \bottomrule
    \end{tabular}
    \caption{Results on nuScenes val set for NonLP mIoU using Neighbourhood occupancy as pretext task and different Multi-Scale formulations}
    \label{tab:appendix_msp_vs_geomae}
\end{table}

\section{NOMAE Improves Sample Efficiency}
The results of our baseline PTv3 in Tab.\,\ref{tab:sota_semseg} are taken from the original paper~\cite{Wu2023PointTV}, To demonstrate that the improvement from NOMAE pretraining is not merely due to faster convergence of the pretrained weights, 
we report additional results for training the model from scratch using longer schedules of $50$, $100$, and $200$ epochs. 
The results in Tab.\,\ref{tab:appendix_longer_scratch} show no significant improvement from longer training, confirming that just 50 pretraining epochs of NOMAE followed by 50 epochs of finetuning improves sample efficiency and reaches higher final performance. We would like to highlight to the reader that we did not experiment with longer pretraining schedules and we suspect that further improvement can be brought by it.

\begin{table}[h]
    \centering
    \footnotesize
    \begin{tabular}{cccccc}
    \toprule
        PTv3Sup,50 ep~\cite{Wu2023PointTV}& Sup, 100ep& Sup, 200ep & NOMAE+50ep\\
        \midrule
        80.4 & 80.7 & 80.6& \textbf{81.8}\\
        \bottomrule
    \end{tabular}
    \caption{Results on Nuscenes SemSeg val set for longer training from scratch compared to NOMAE pretraining}
    \label{tab:appendix_longer_scratch}
\end{table}

\begin{figure*}
\centering
\footnotesize
\begin{tabular}{cccc}
&Baseline Output& NOMAE Output& Improvement/Error Map\\
(a) &\subfloat{\includegraphics[trim={2cm 5cm 2cm 5cm},clip,width=0.25\textwidth]{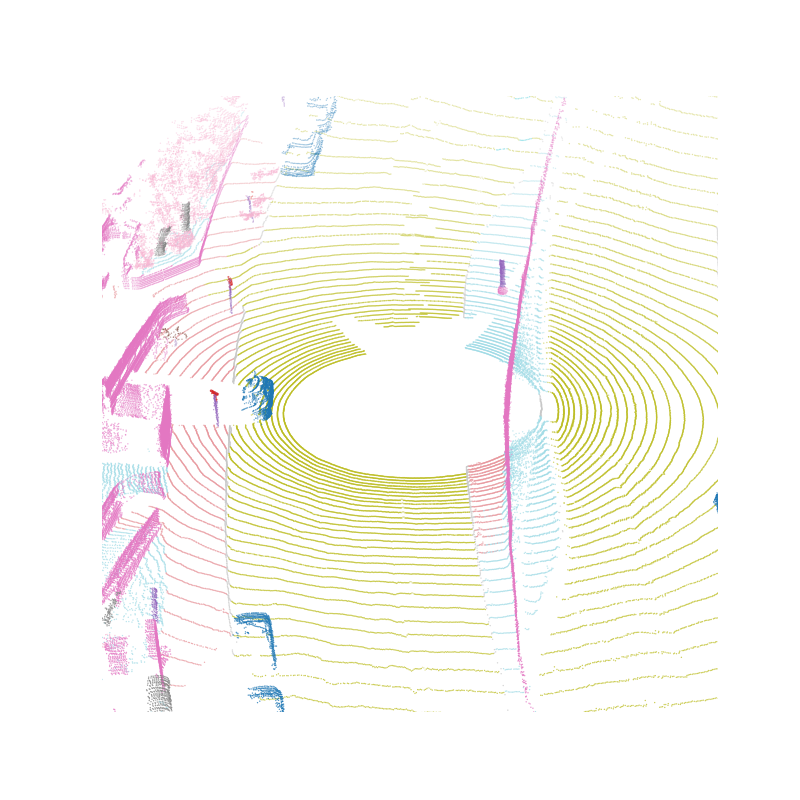}}
&\subfloat{\includegraphics[trim={2cm 5cm 2cm 5cm},clip,width=0.25\textwidth]{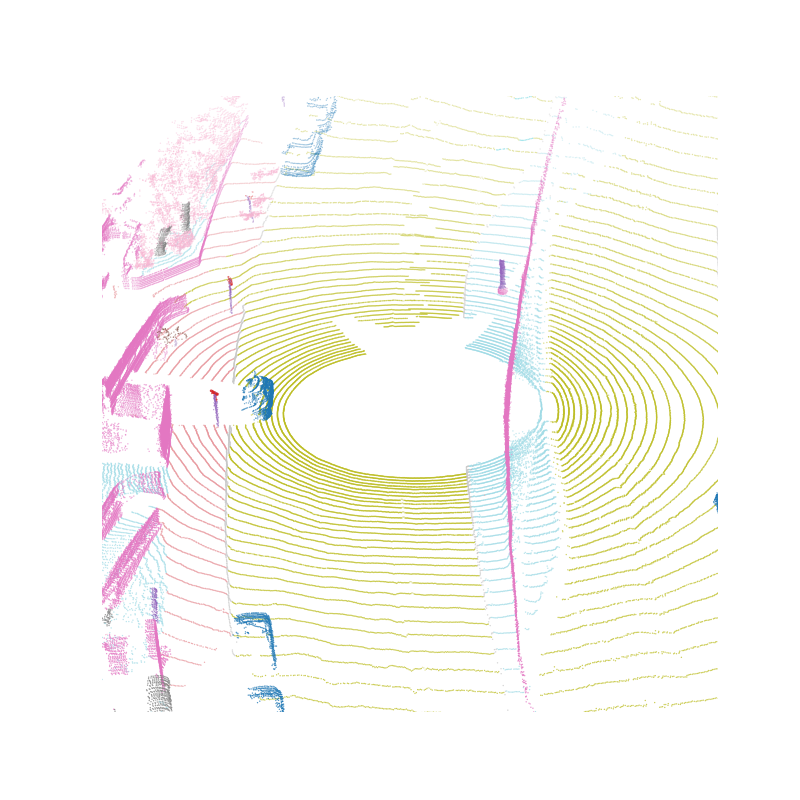}}
&\subfloat{\includegraphics[trim={2cm 5cm 2cm 5cm},clip,width=0.25\textwidth]{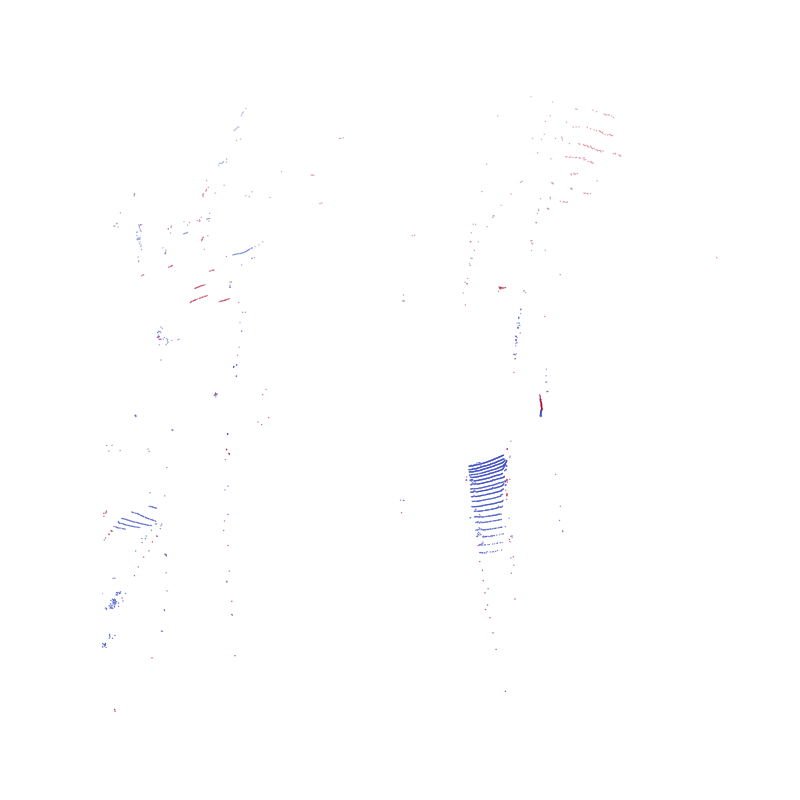}}
\\
\midrule
(b) &\subfloat{\includegraphics[trim={2cm 5cm 2cm 5cm},clip,width=0.25\textwidth]{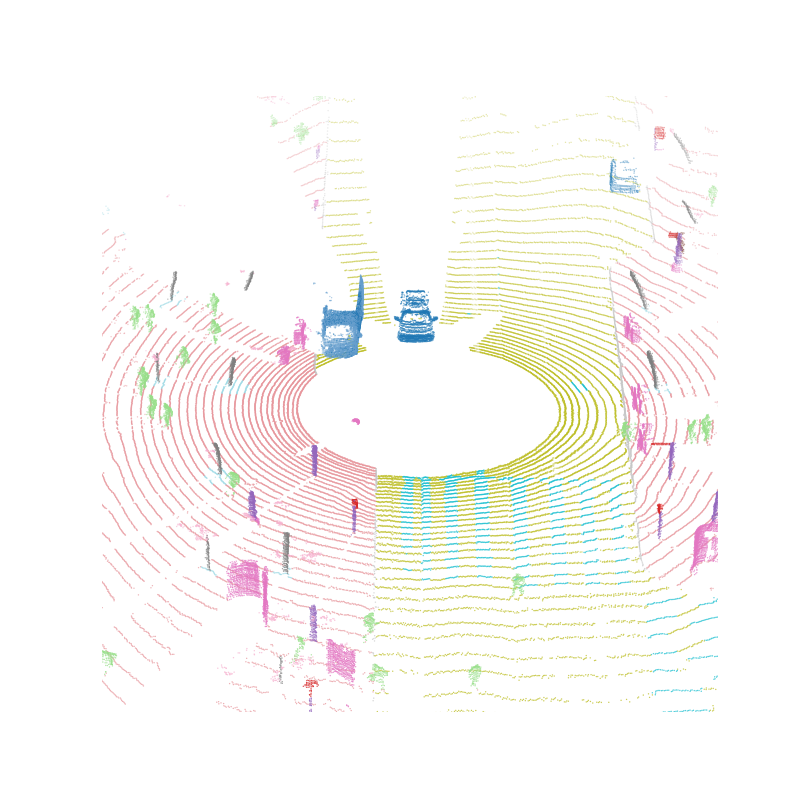}}
&\subfloat{\includegraphics[trim={2cm 5cm 2cm 5cm},clip,width=0.25\textwidth]{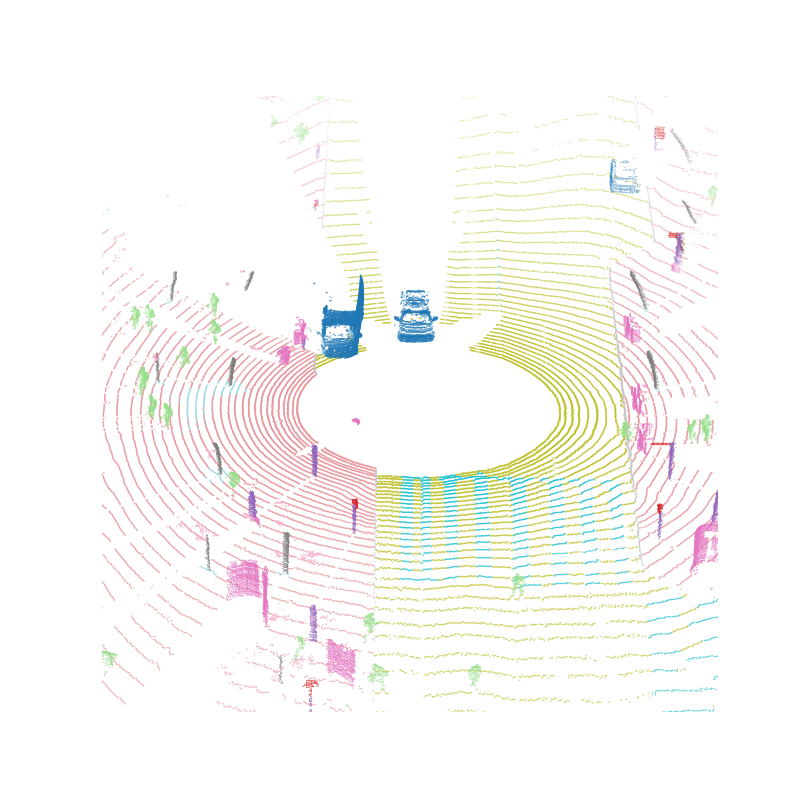}}
&\subfloat{\includegraphics[trim={2cm 5cm 2cm 5cm},clip,width=0.25\textwidth]{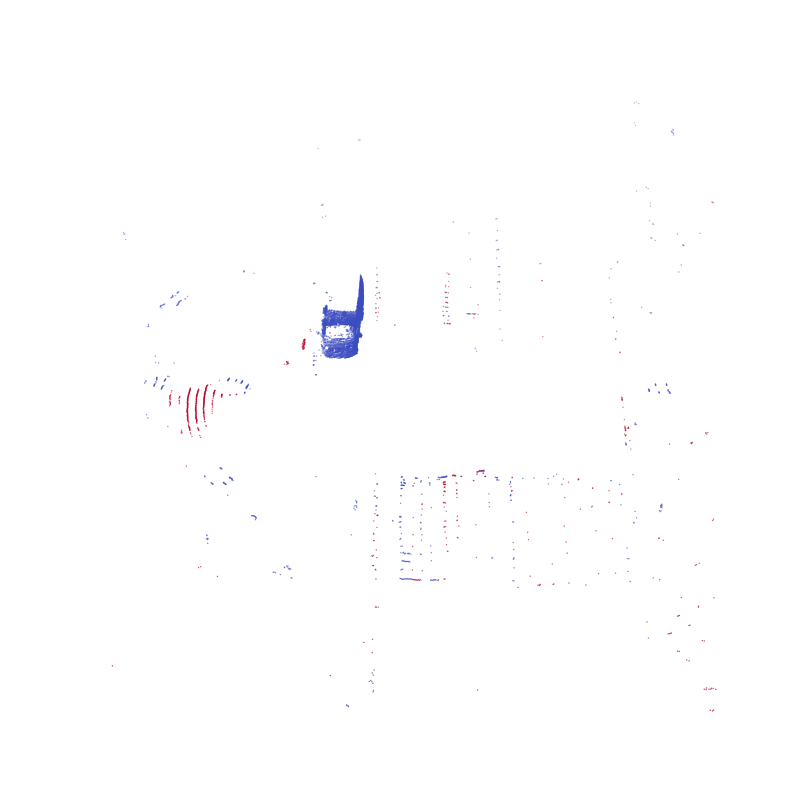}}
\\
\midrule
(c) &\subfloat{\includegraphics[trim={2cm 5cm 2cm 5cm},clip,width=0.25\textwidth]{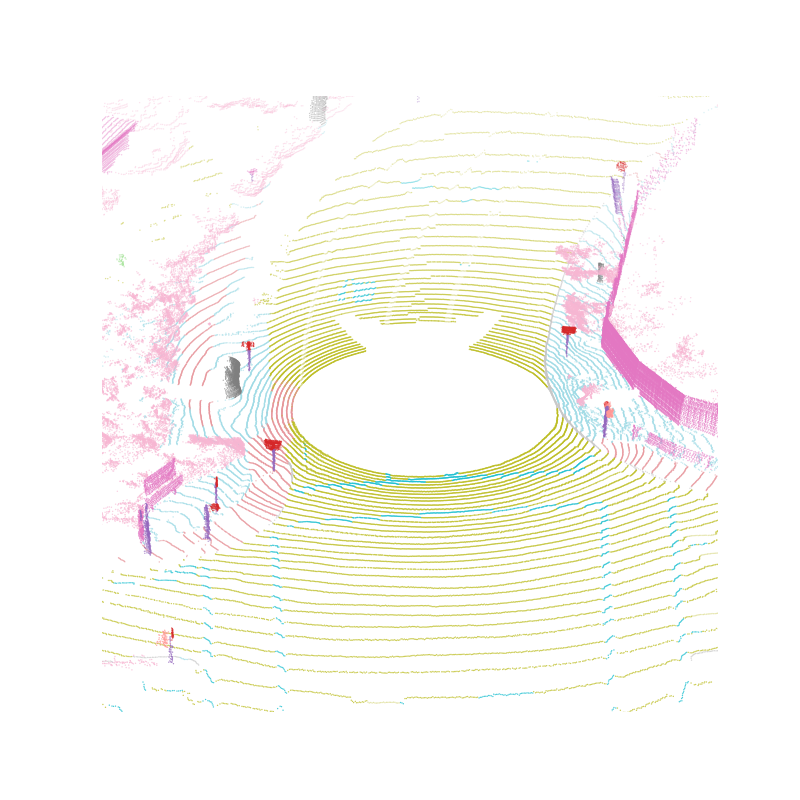}}
&\subfloat{\includegraphics[trim={2cm 5cm 2cm 5cm},clip,width=0.25\textwidth]{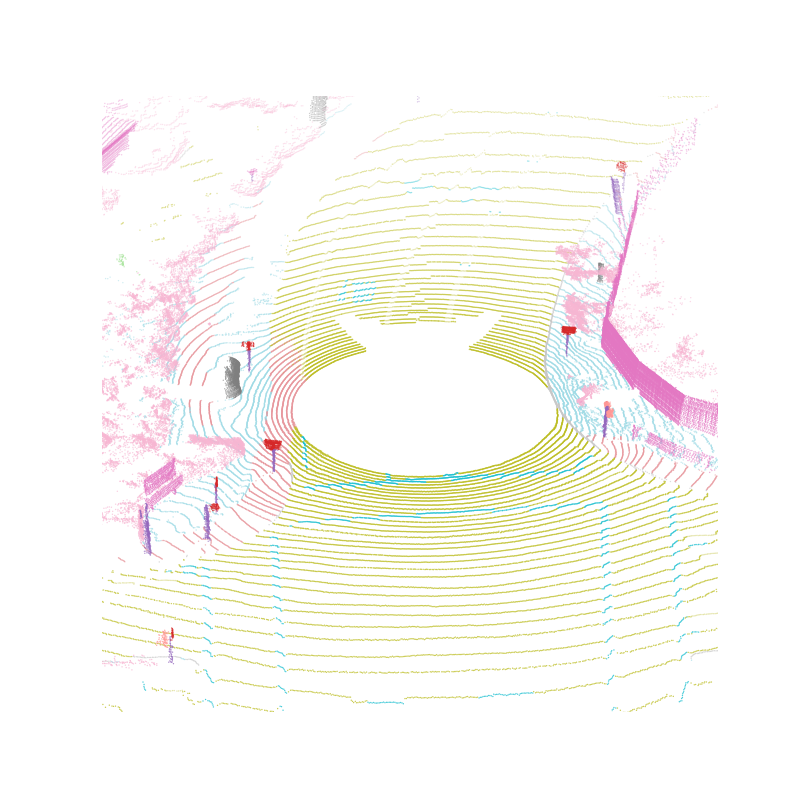}}
&\subfloat{\includegraphics[trim={2cm 5cm 2cm 5cm},clip,width=0.25\textwidth]{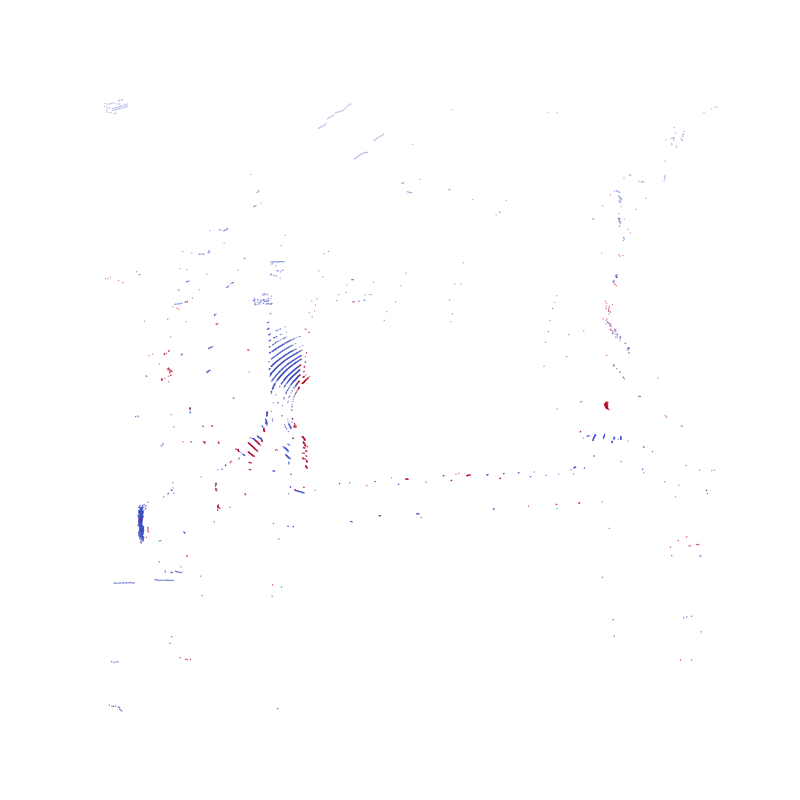}}
\\
\midrule
(d) &\subfloat{\includegraphics[trim={2cm 5cm 2cm 5cm},clip,width=0.25\textwidth]{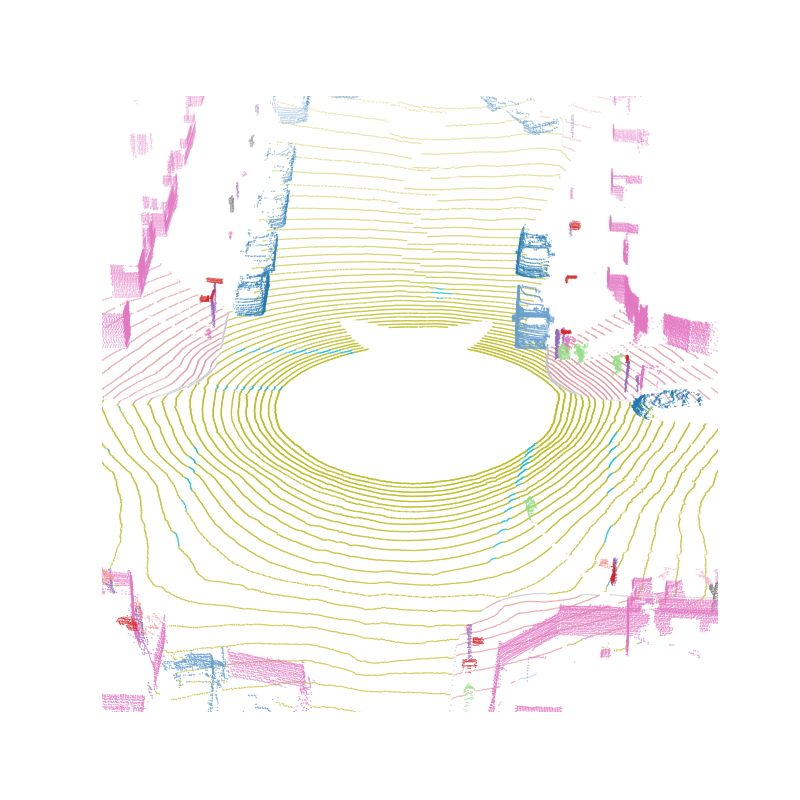}}
&\subfloat{\includegraphics[trim={2cm 5cm 2cm 5cm},clip,width=0.25\textwidth]{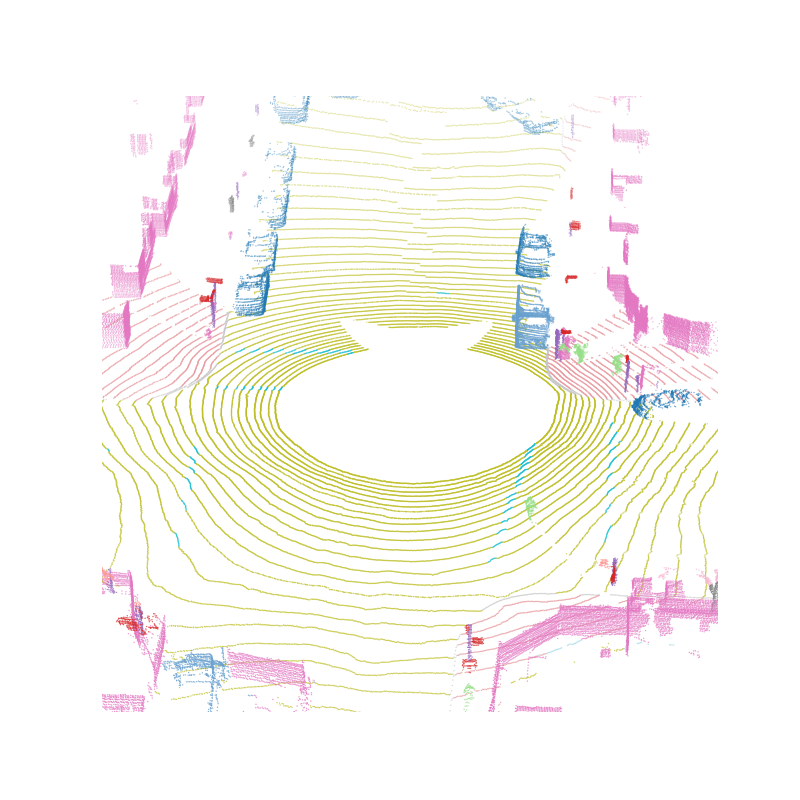}}
&\subfloat{\includegraphics[trim={2cm 5cm 2cm 5cm},clip,width=0.25\textwidth]{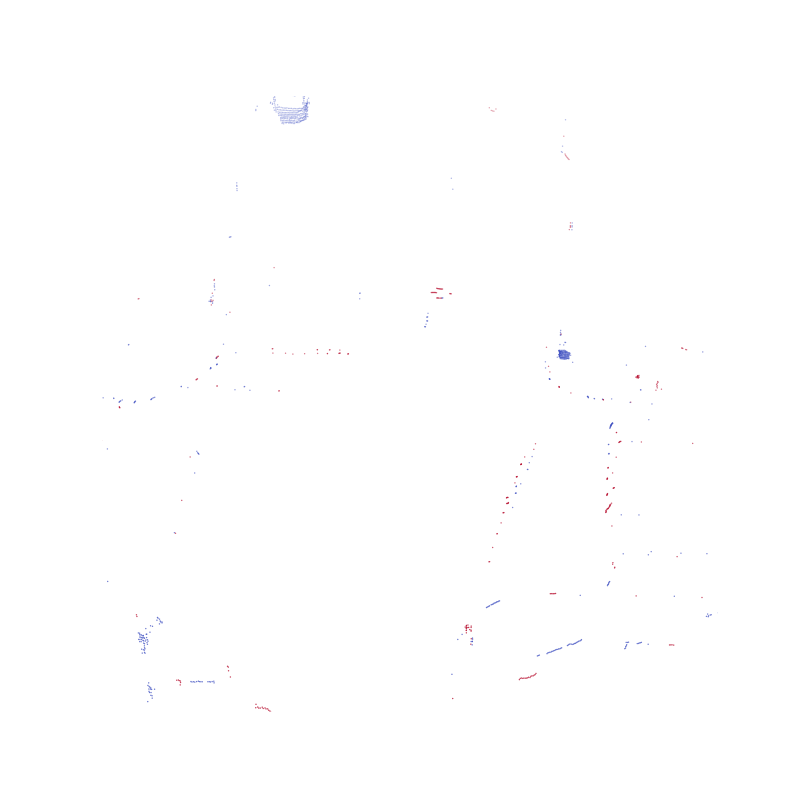}}
\\
\midrule
(e) &\subfloat{\includegraphics[trim={2cm 5cm 2cm 5cm},clip,width=0.25\textwidth]{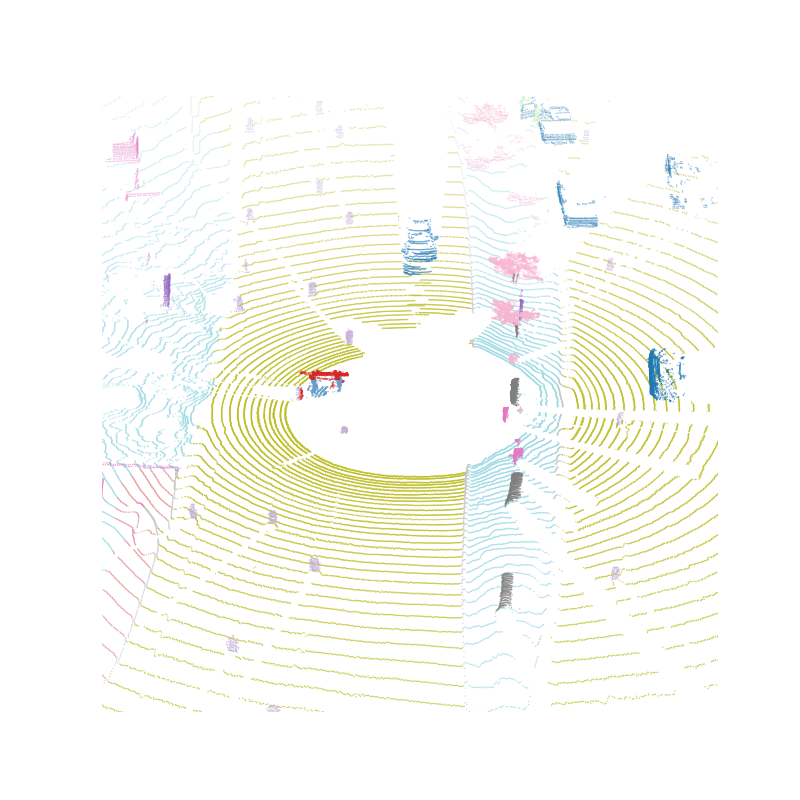}}
&\subfloat{\includegraphics[trim={2cm 5cm 2cm 5cm},clip,width=0.25\textwidth]{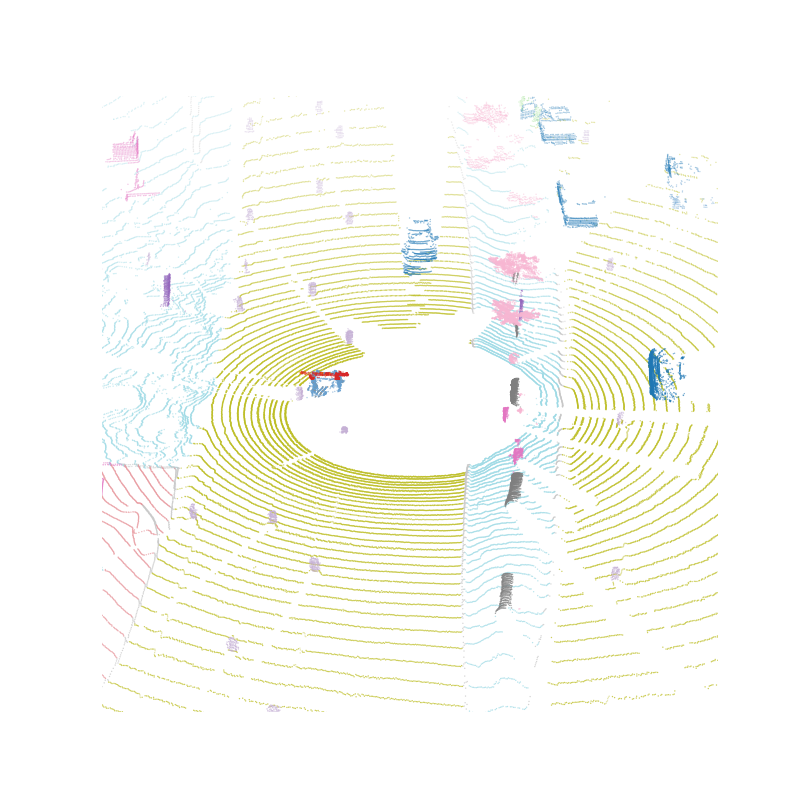}}
&\subfloat{\includegraphics[trim={2cm 5cm 2cm 5cm},clip,width=0.25\textwidth]{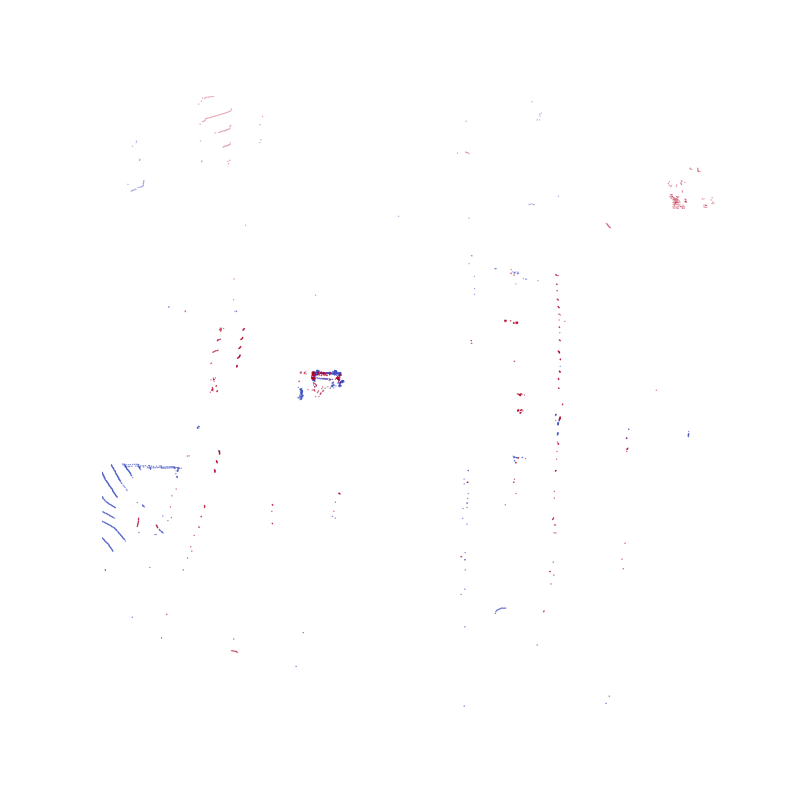}}
\\
\midrule
(f) &\subfloat{\includegraphics[trim={2cm 5cm 2cm 5cm},clip,width=0.25\textwidth]{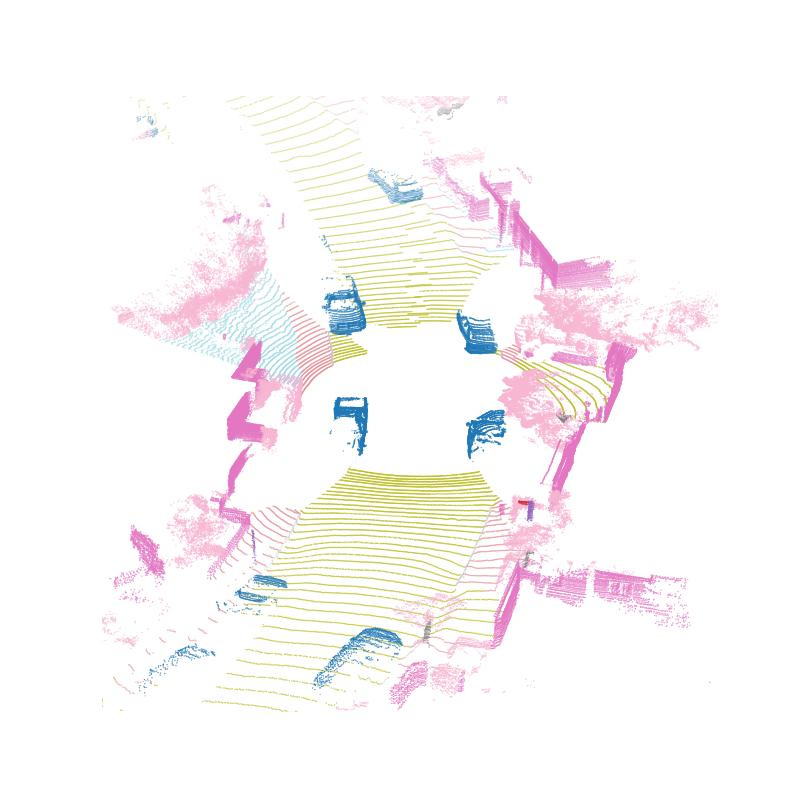}}
&\subfloat{\includegraphics[trim={2cm 5cm 2cm 5cm},clip,width=0.25\textwidth]{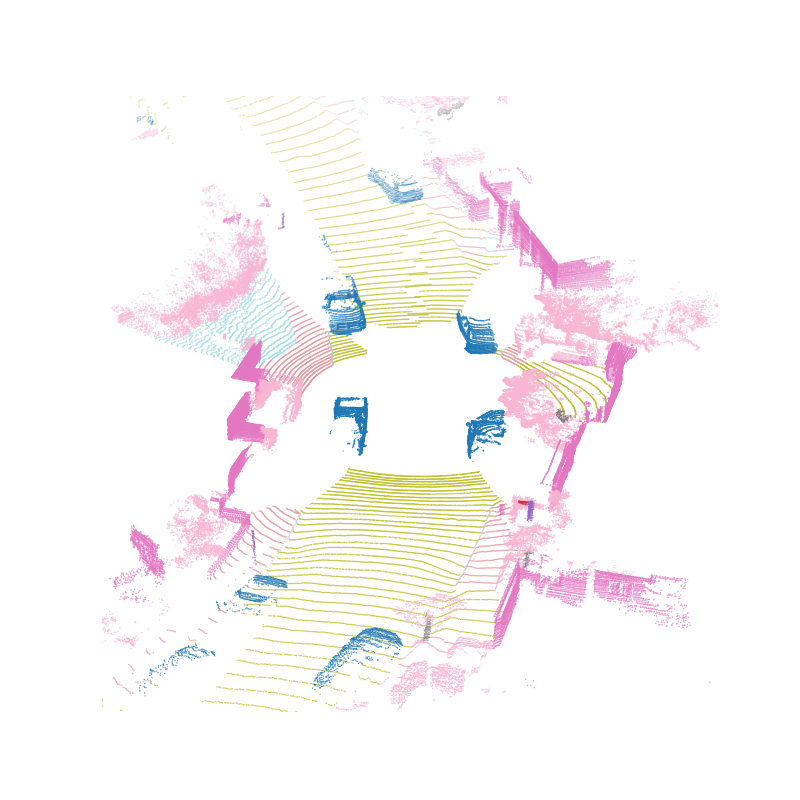}}
&\subfloat{\includegraphics[trim={2cm 5cm 2cm 5cm},clip,width=0.25\textwidth]{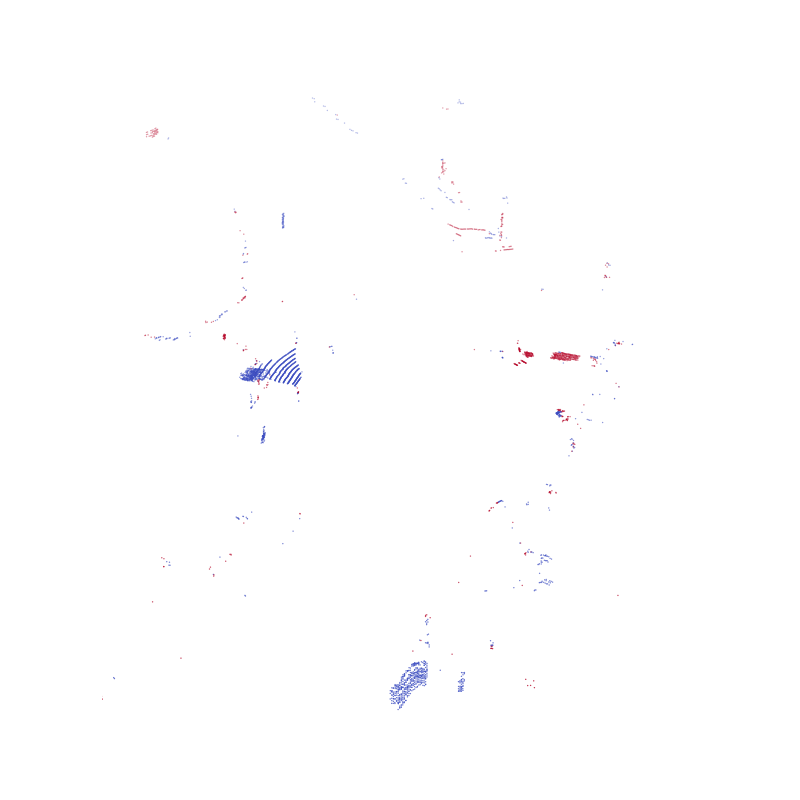}}
\\
\midrule
(g) &\subfloat{\includegraphics[trim={2cm 5cm 2cm 5cm},clip,width=0.25\textwidth]{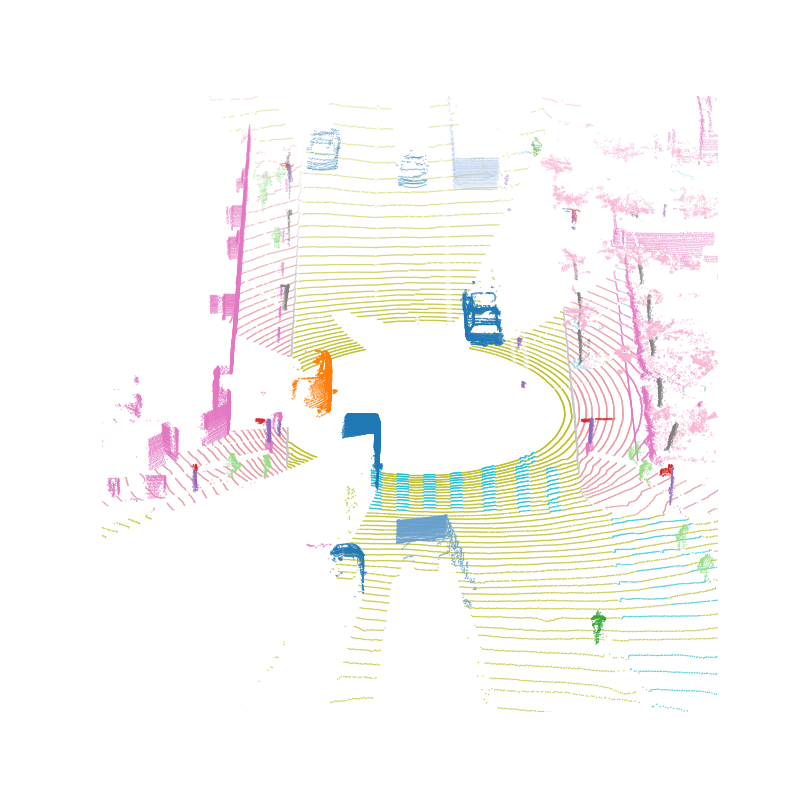}}
&\subfloat{\includegraphics[trim={2cm 5cm 2cm 5cm},clip,width=0.25\textwidth]{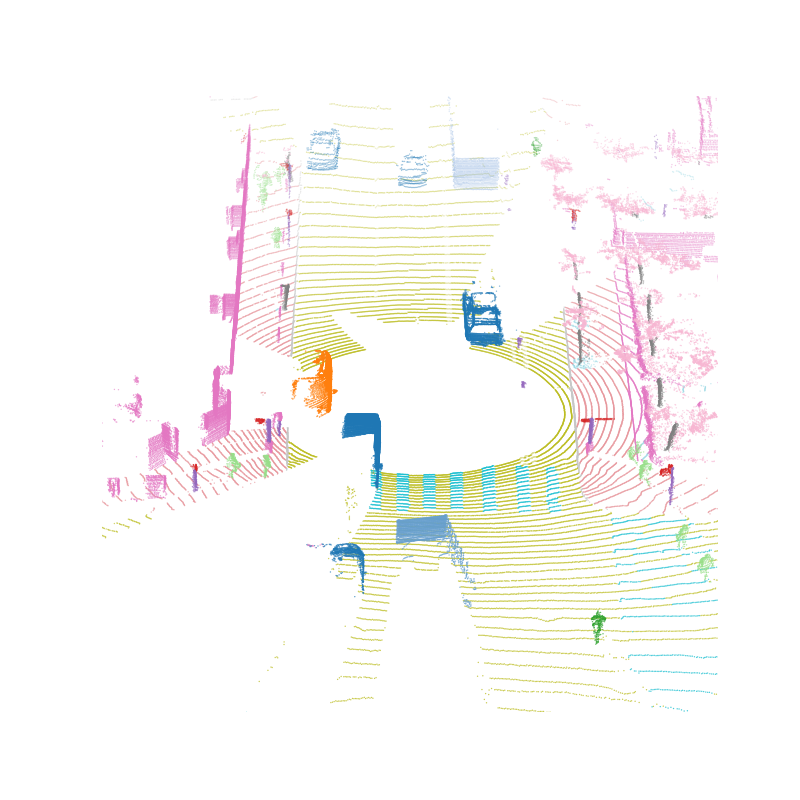}}
&\subfloat{\includegraphics[trim={2cm 5cm 2cm 5cm},clip,width=0.25\textwidth]{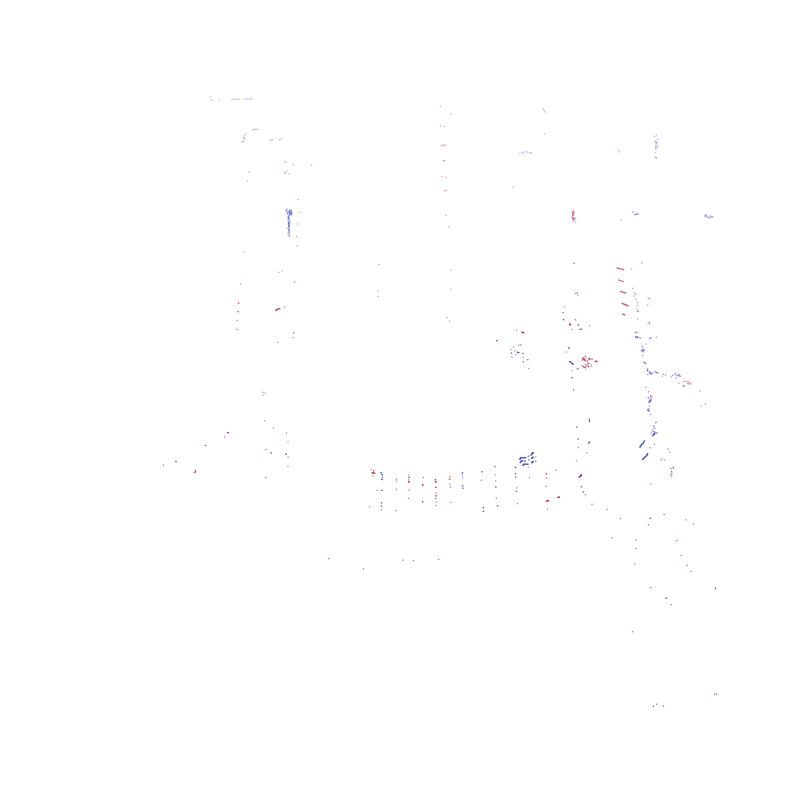}}
\\
\end{tabular}
\caption{Qualitative comparison of semantic segmentation performance with our baseline Ptv3~\citep{Wu2023PointTV} on the Waymo validation set. }
\label{fig:appendix_semseg_qualitative}
\vspace{-0.5em}
\end{figure*}
\begin{figure*}
\centering
\footnotesize
\begin{tabular}{cccc}
&Baseline Output& NOMAE Output& Ground Truth\\
(a) &\subfloat{\includegraphics[trim={2cm 35cm 2cm 35cm},clip,width=0.245\textwidth]{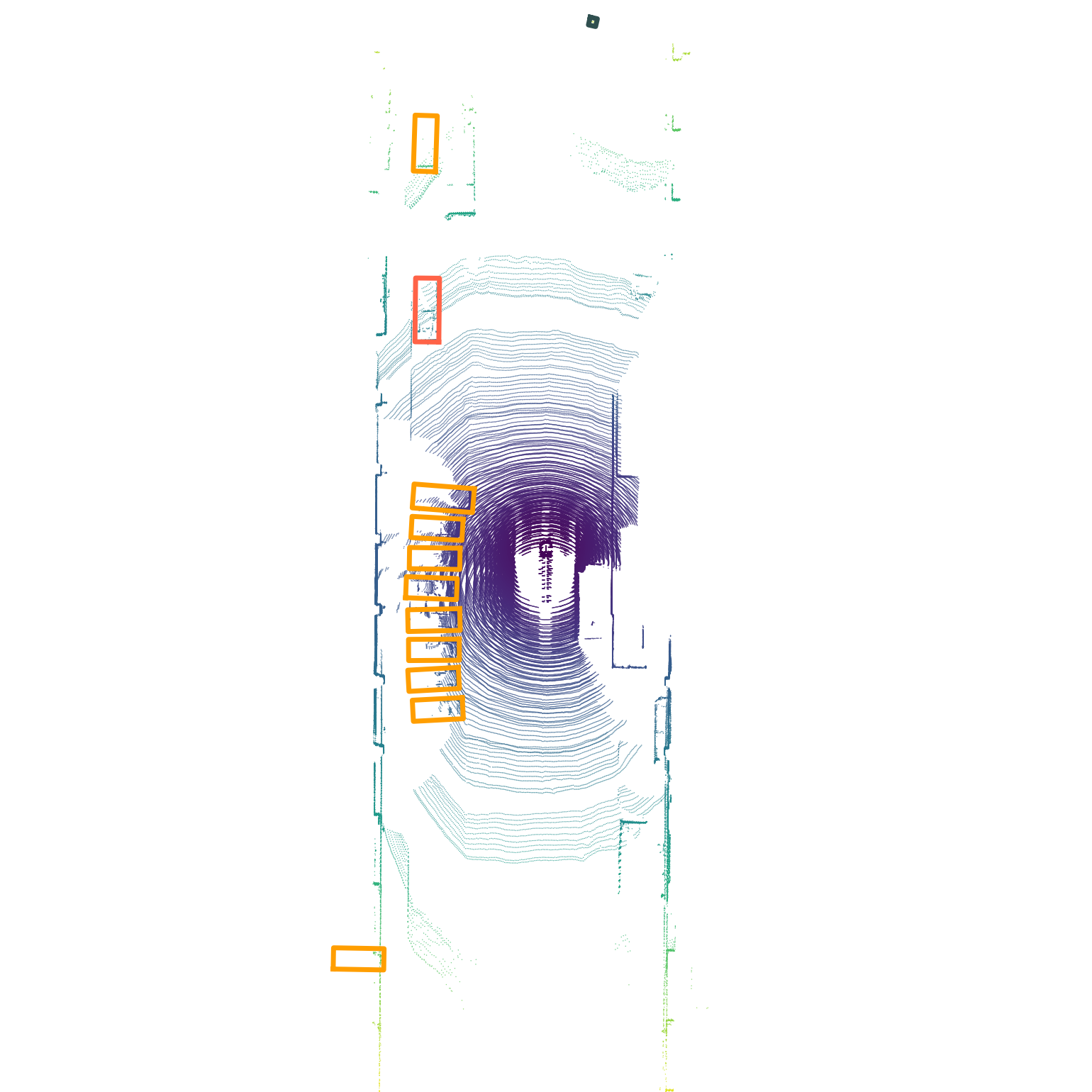}}
&\subfloat{\includegraphics[trim={2cm 35cm 2cm 35cm},clip,width=0.245\textwidth]{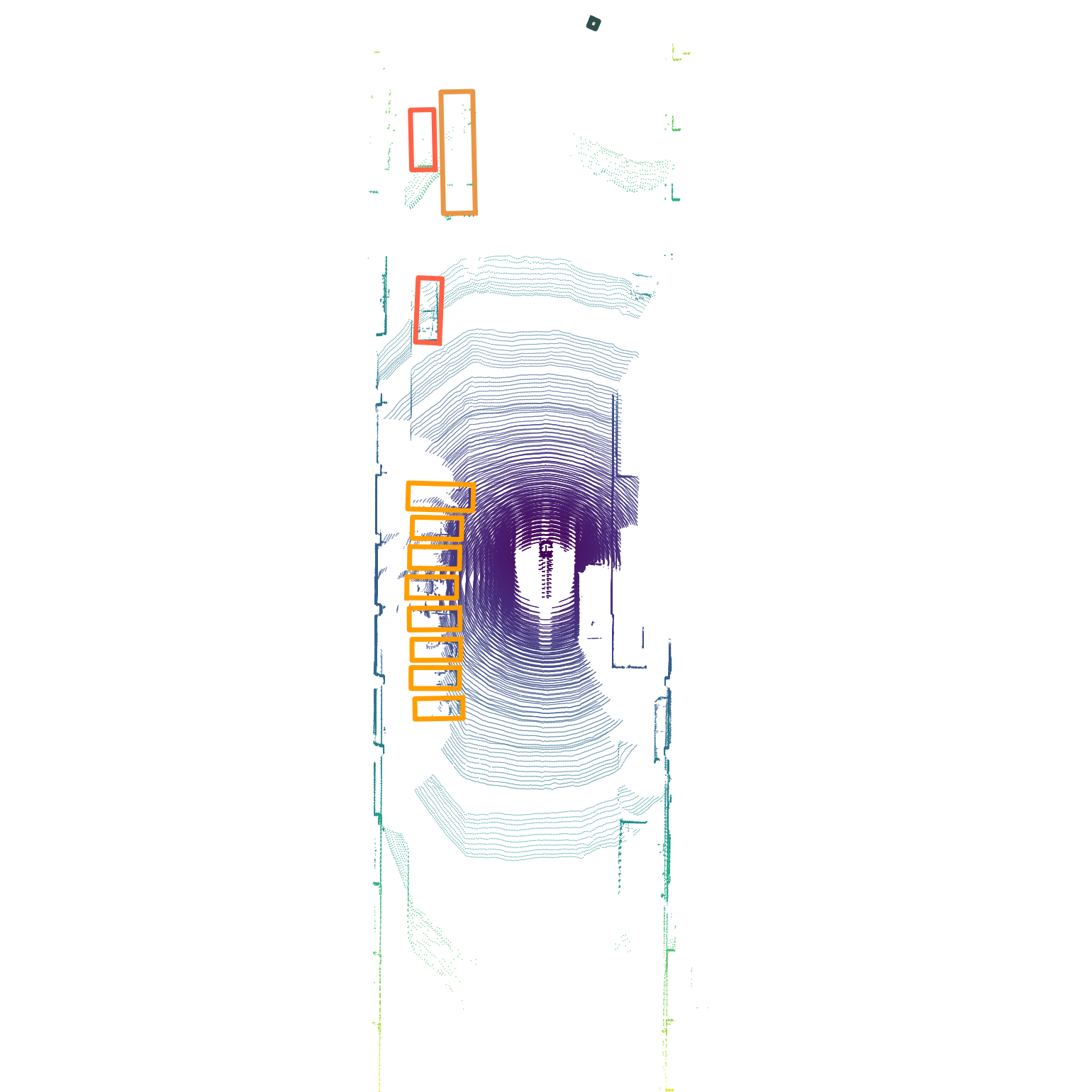}}
&\subfloat{\includegraphics[trim={2cm 35cm 2cm 35cm},clip,width=0.245\textwidth]{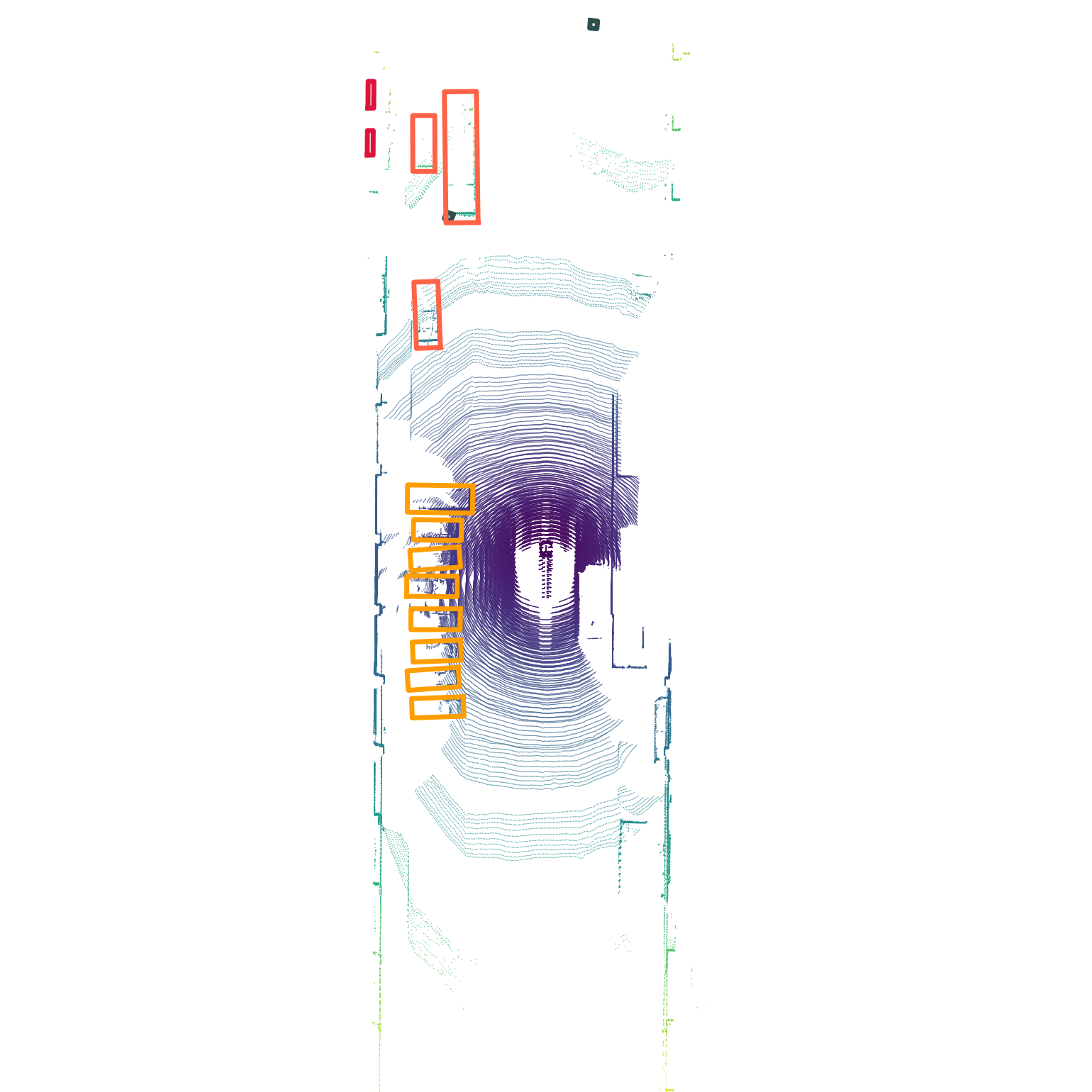}}\\\midrule
(b) &\subfloat{\includegraphics[trim={2cm 35cm 2cm 35cm},clip,width=0.245\textwidth]{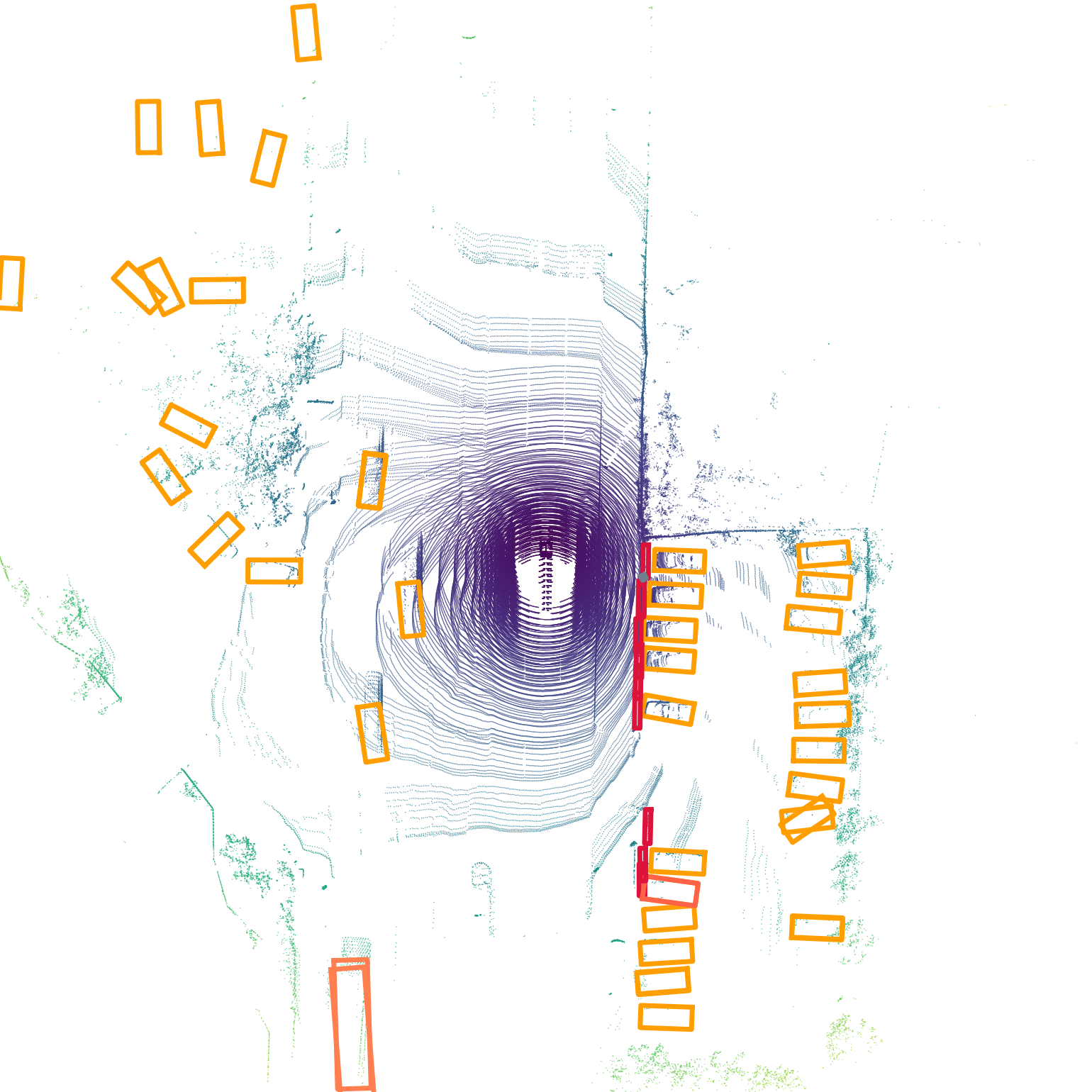}}
&\subfloat{\includegraphics[trim={2cm 35cm 2cm 35cm},clip,width=0.245\textwidth]{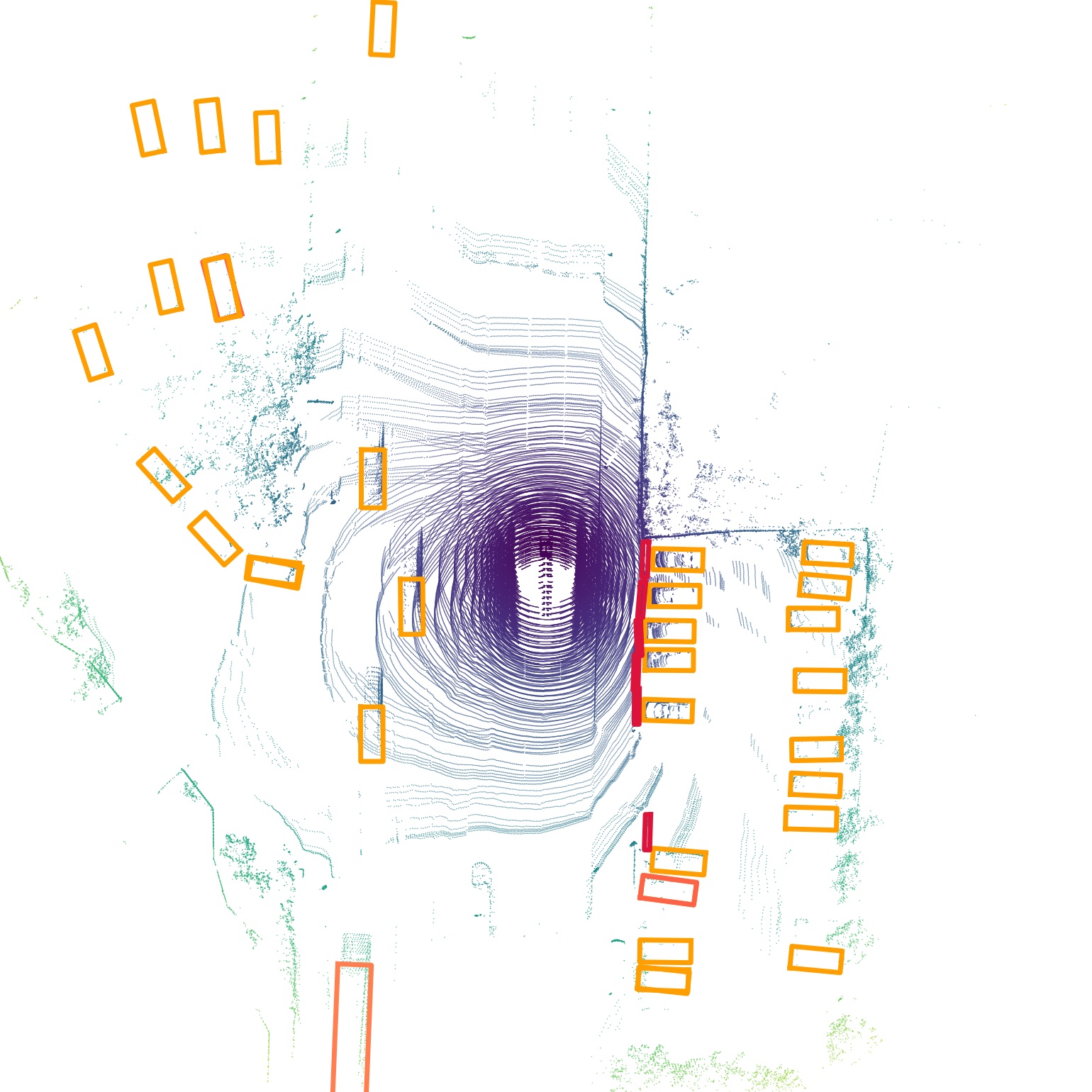}}
&\subfloat{\includegraphics[trim={2cm 35cm 2cm 35cm},clip,width=0.245\textwidth]{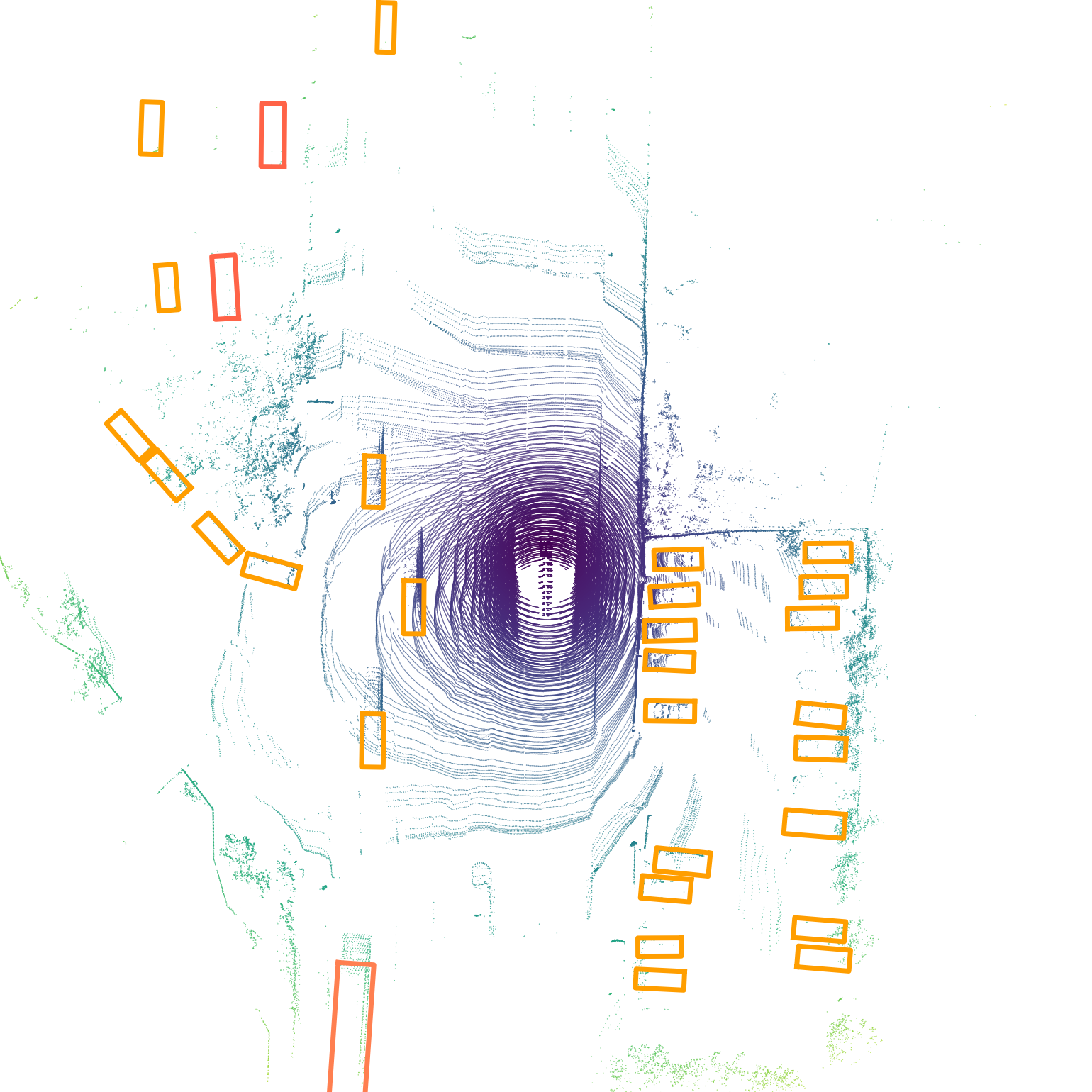}}\\\midrule
(c) &\subfloat{\includegraphics[trim={2cm 35cm 2cm 35cm},clip,width=0.245\textwidth]{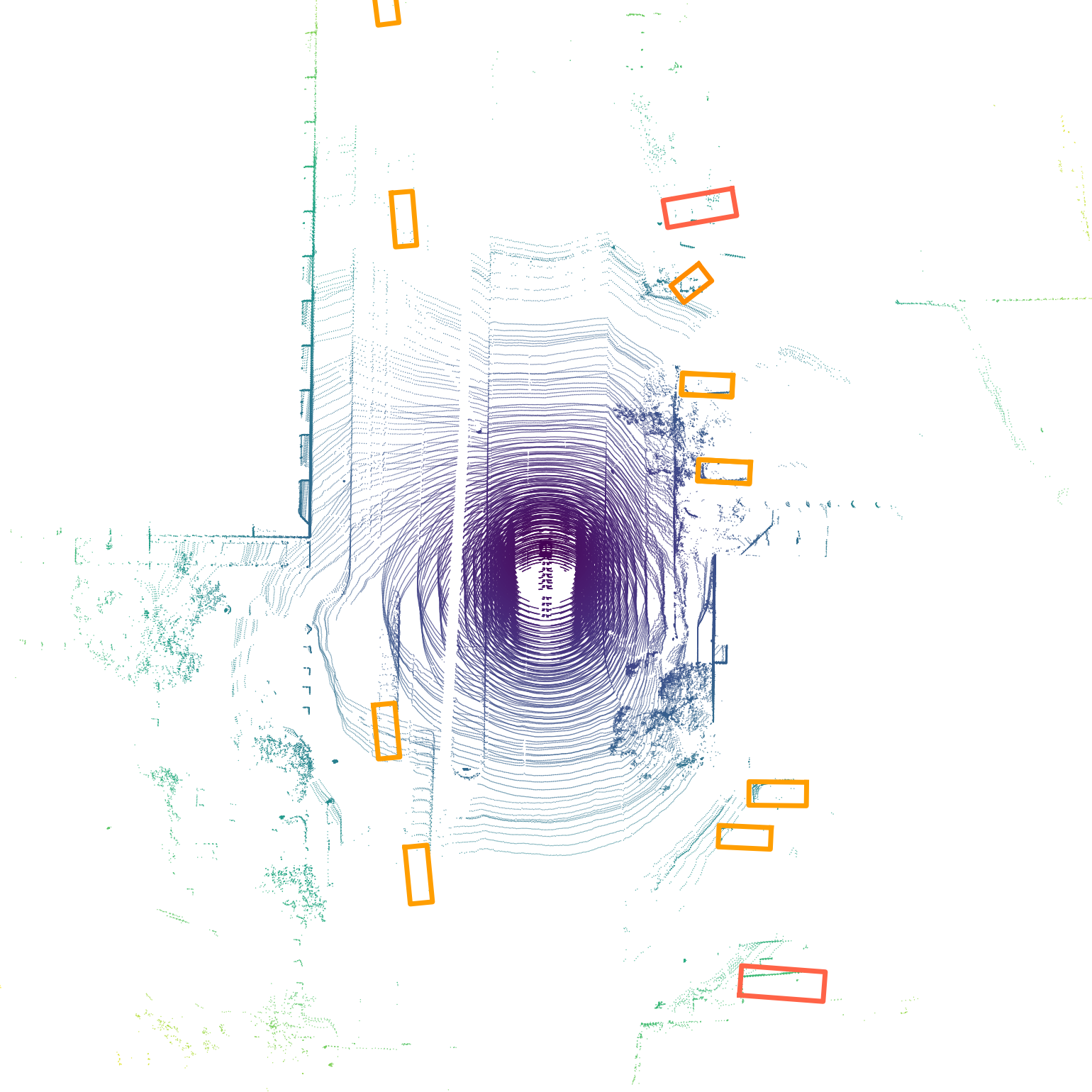}}
&\subfloat{\includegraphics[trim={2cm 35cm 2cm 35cm},clip,width=0.245\textwidth]{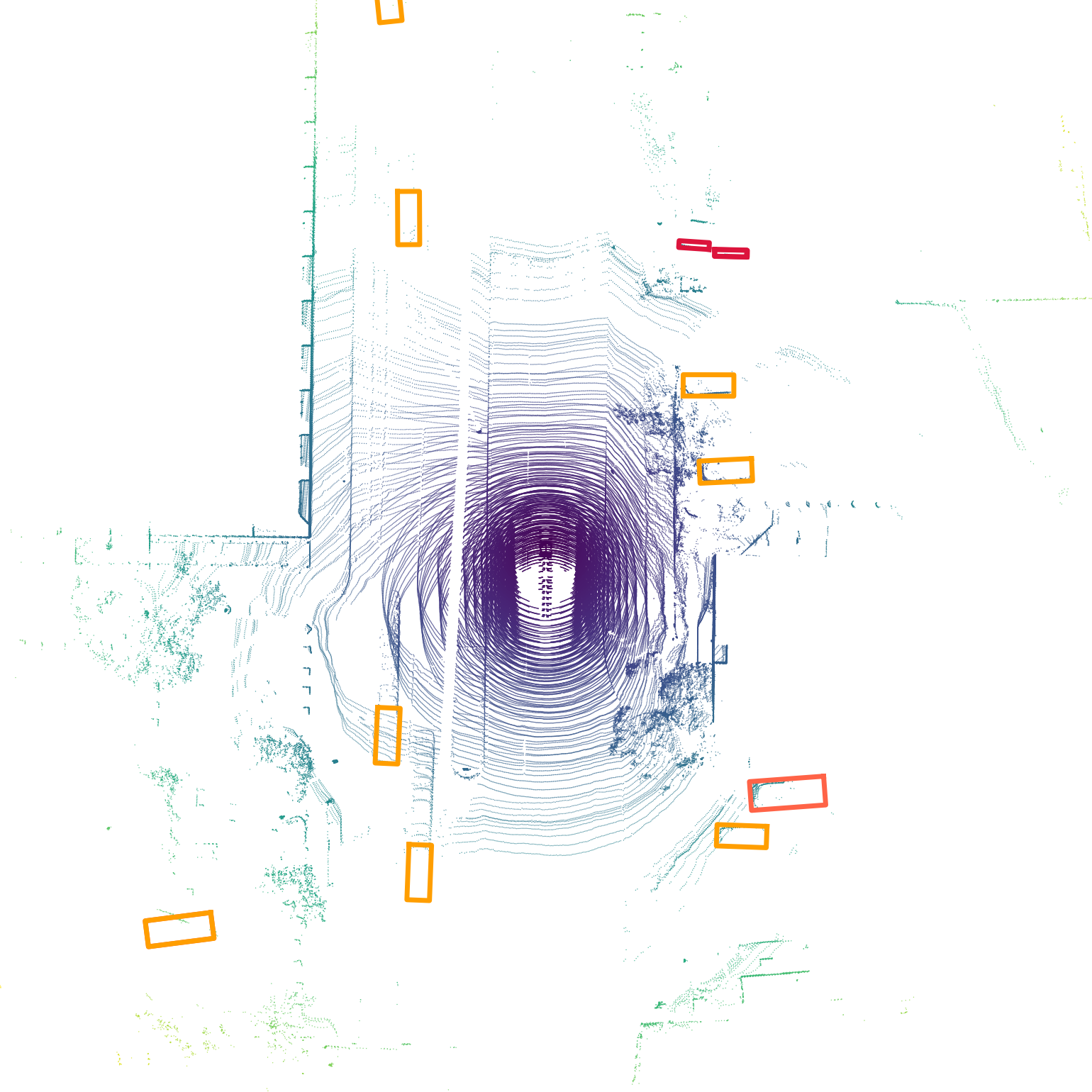}}
&\subfloat{\includegraphics[trim={2cm 35cm 2cm 35cm},clip,width=0.245\textwidth]{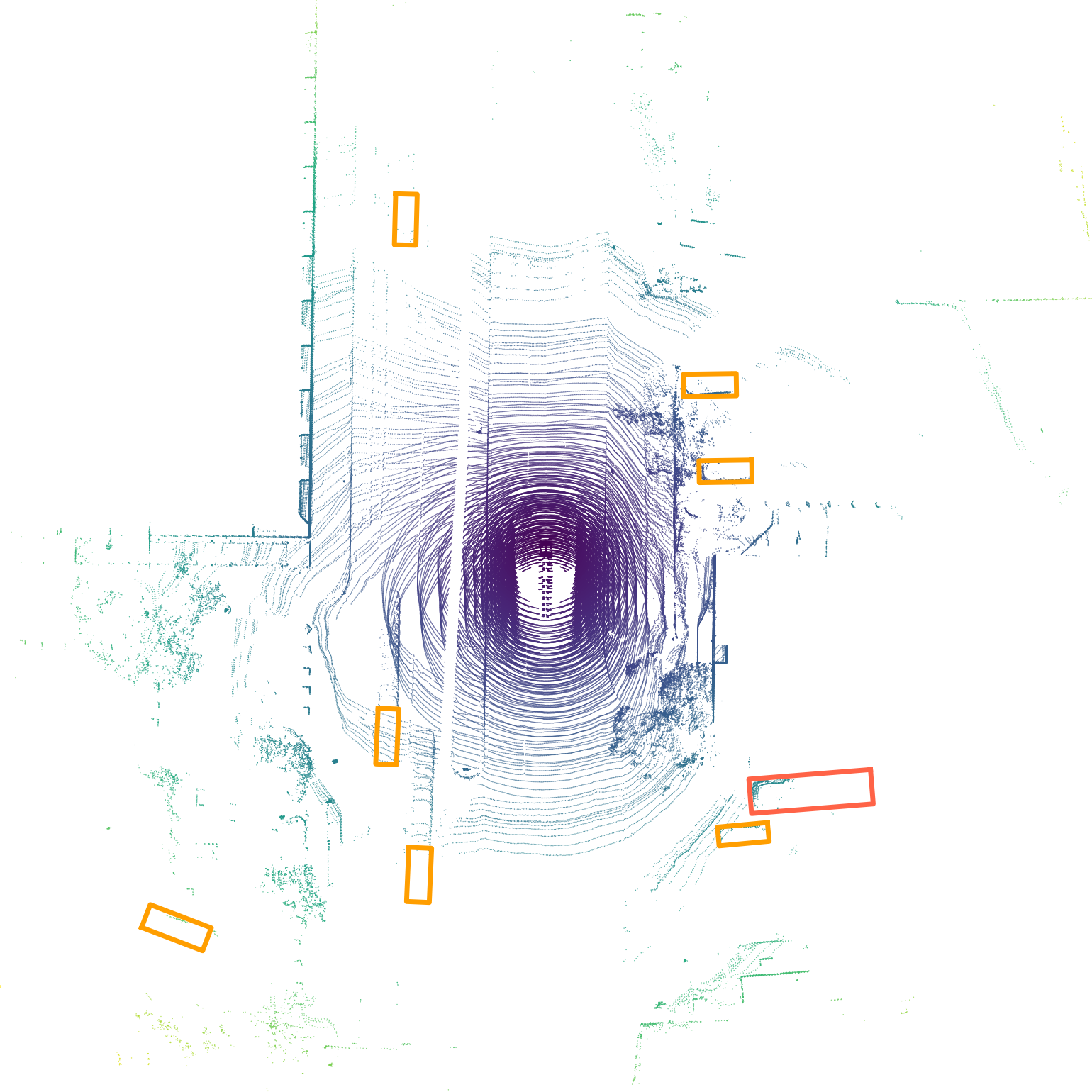}}\\\midrule
(d) &\subfloat{\includegraphics[trim={2cm 35cm 2cm 35cm},clip,width=0.245\textwidth]{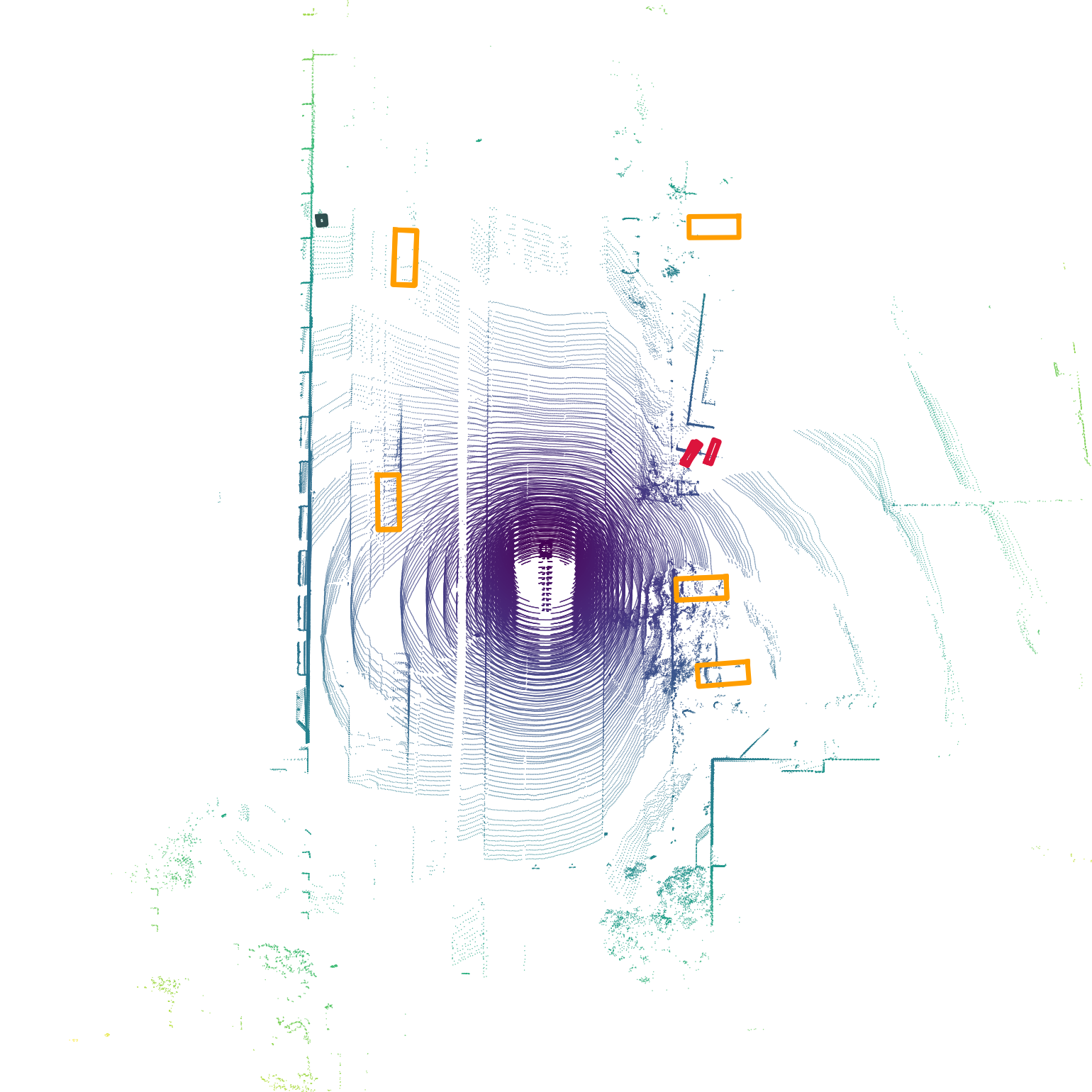}}
&\subfloat{\includegraphics[trim={2cm 35cm 2cm 35cm},clip,width=0.245\textwidth]{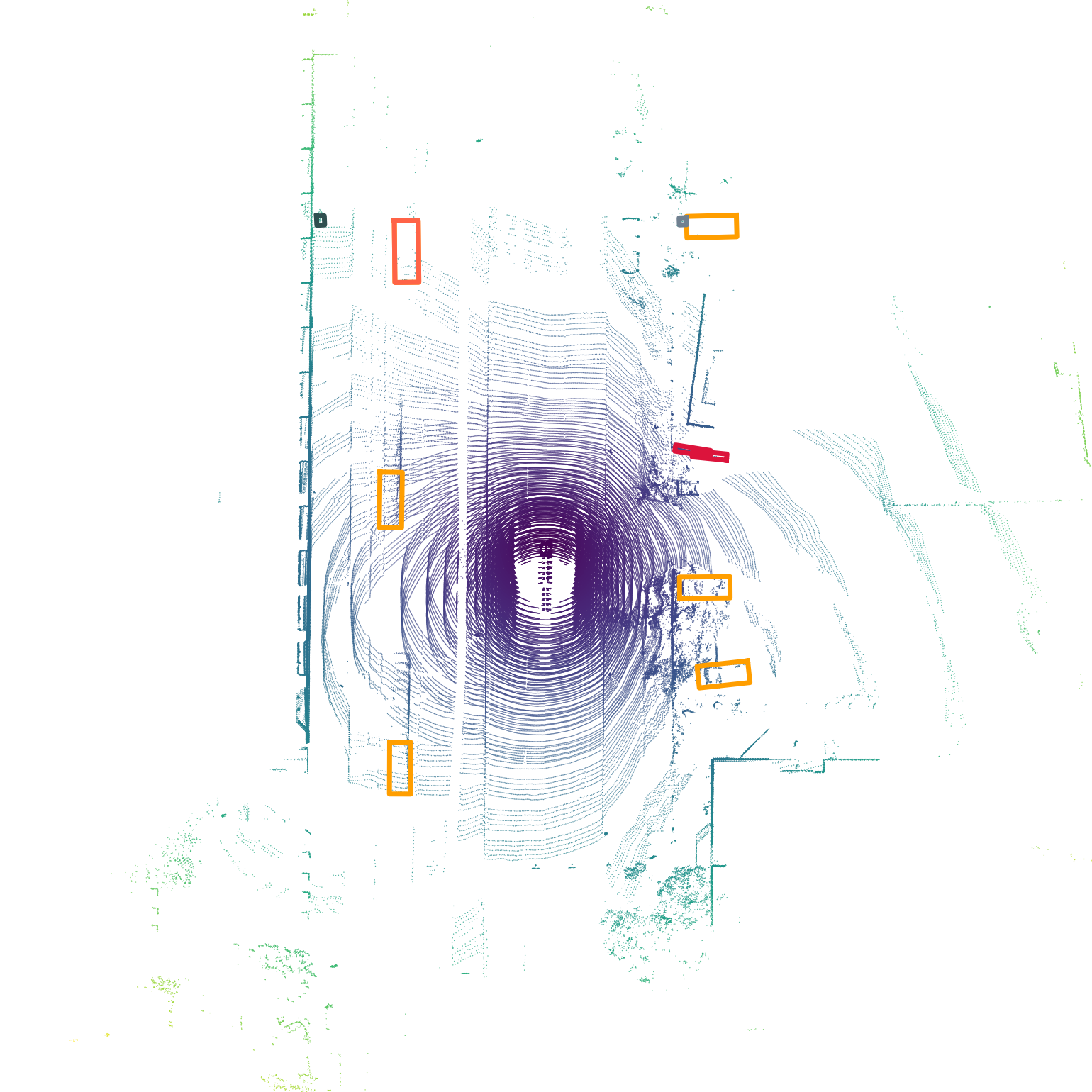}}
&\subfloat{\includegraphics[trim={2cm 35cm 2cm 35cm},clip,width=0.245\textwidth]{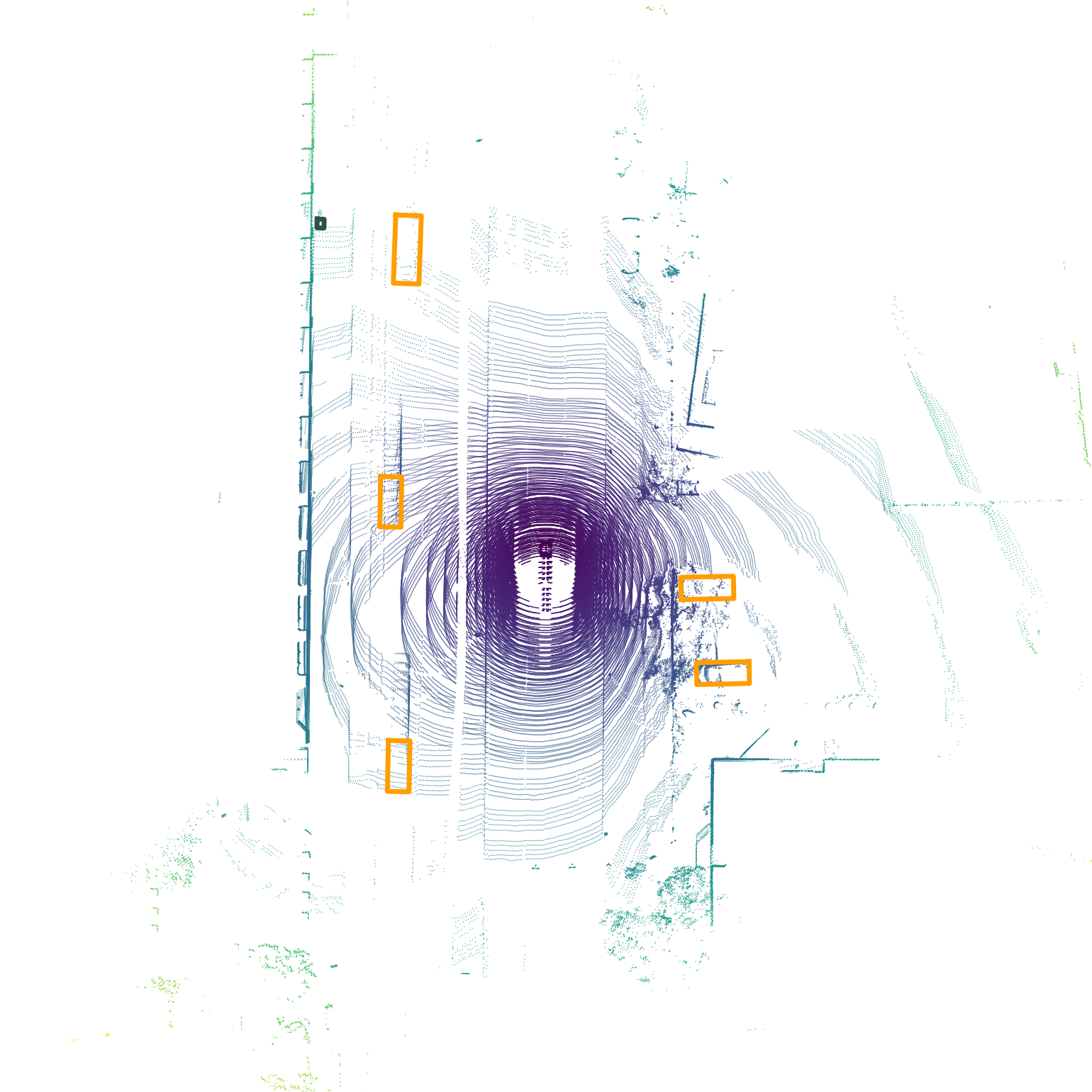}}\\\midrule
(e) &\subfloat{\includegraphics[trim={2cm 35cm 2cm 35cm},clip,width=0.245\textwidth]{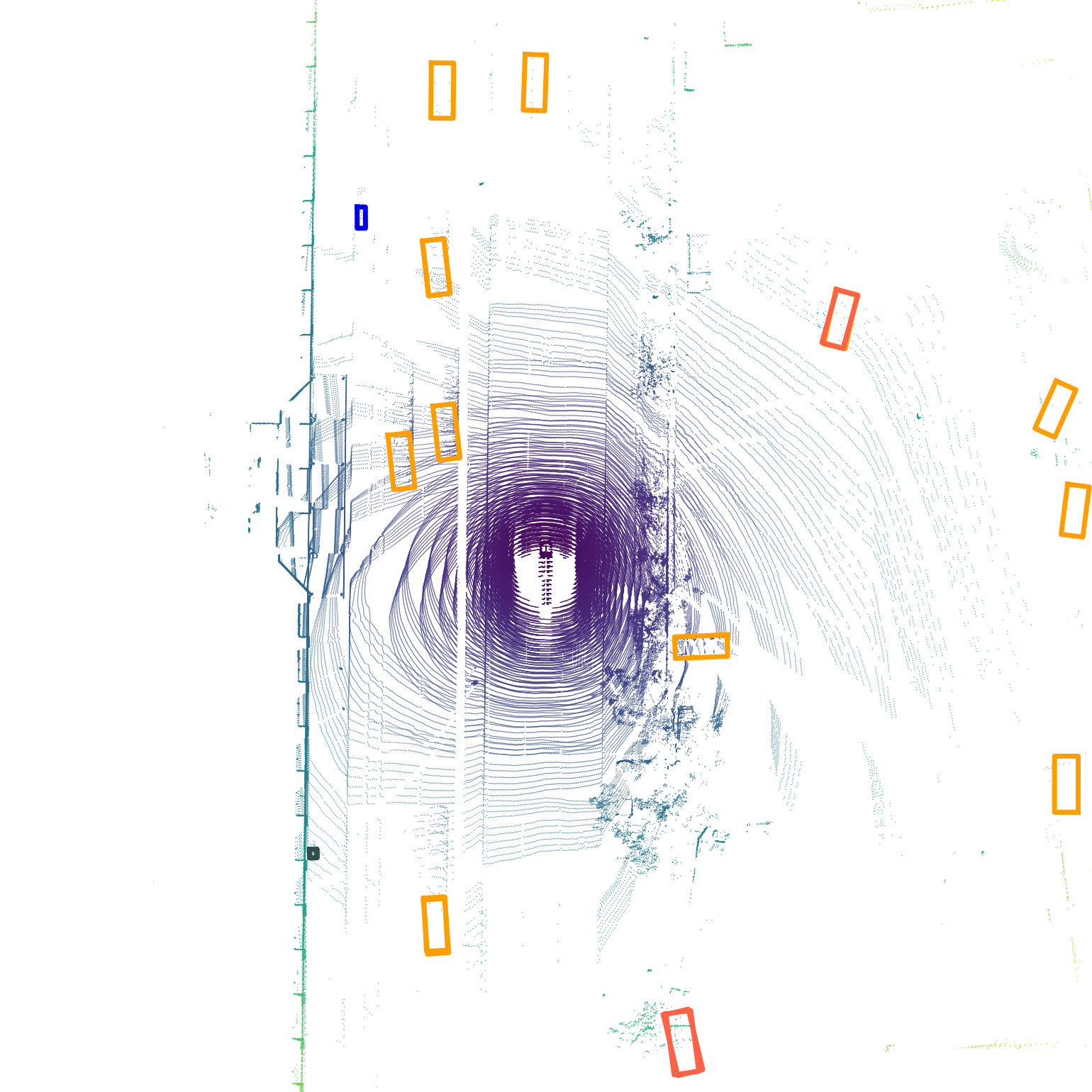}}
&\subfloat{\includegraphics[trim={2cm 35cm 2cm 35cm},clip,width=0.245\textwidth]{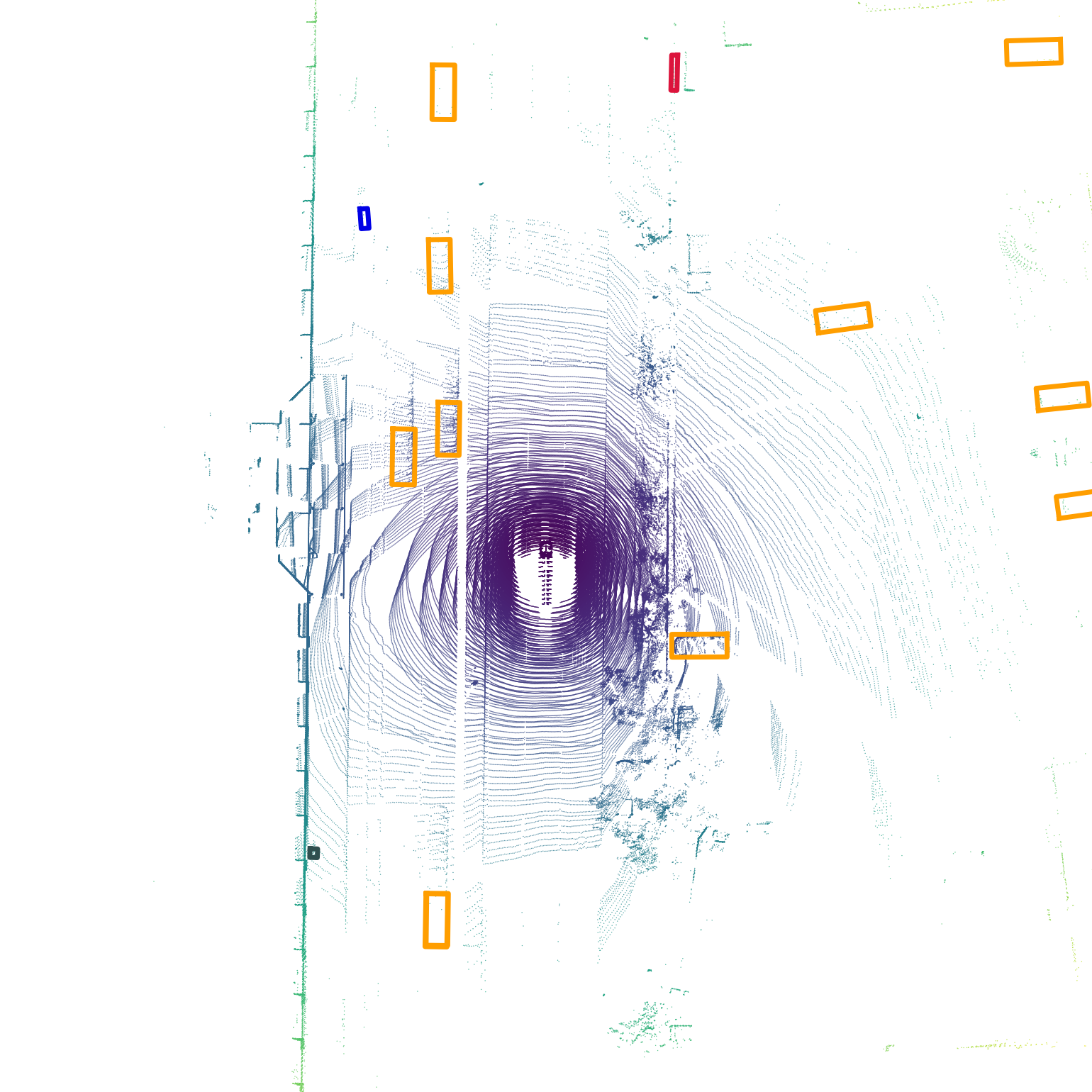}}
&\subfloat{\includegraphics[trim={2cm 35cm 2cm 35cm},clip,width=0.245\textwidth]{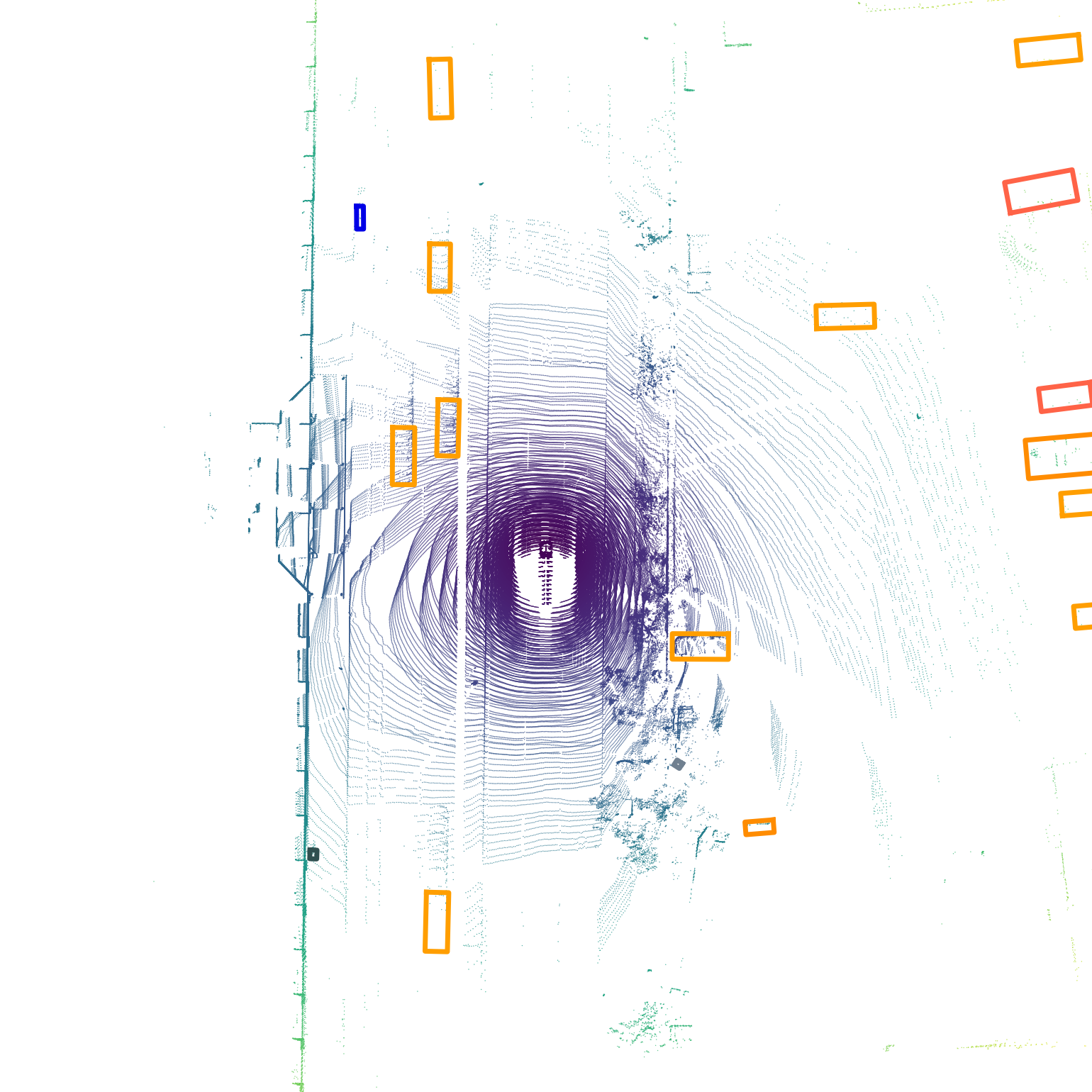}}\\\midrule
(f) &\subfloat{\includegraphics[trim={2cm 35cm 2cm 35cm},clip,width=0.245\textwidth]{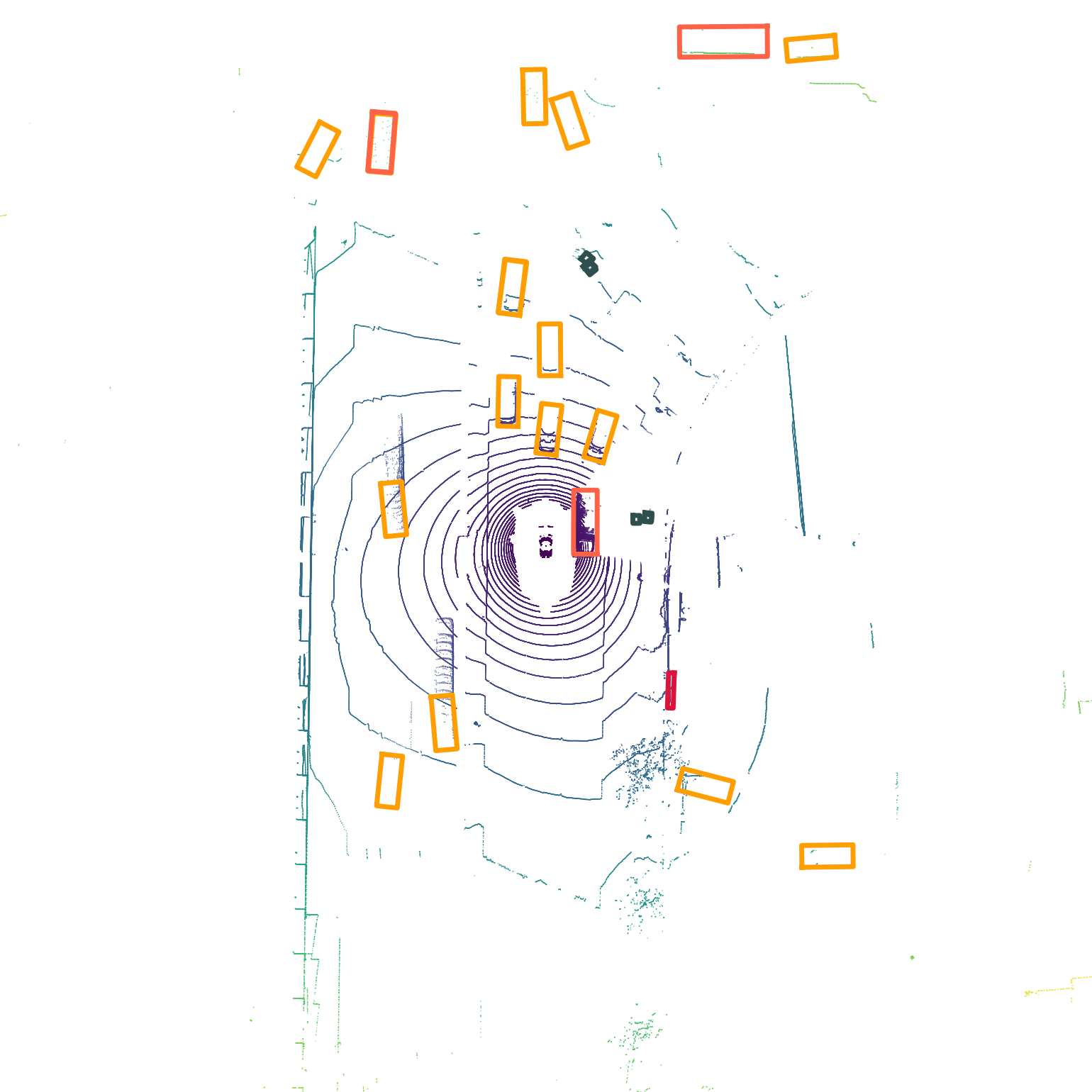}}
&\subfloat{\includegraphics[trim={2cm 35cm 2cm 35cm},clip,width=0.245\textwidth]{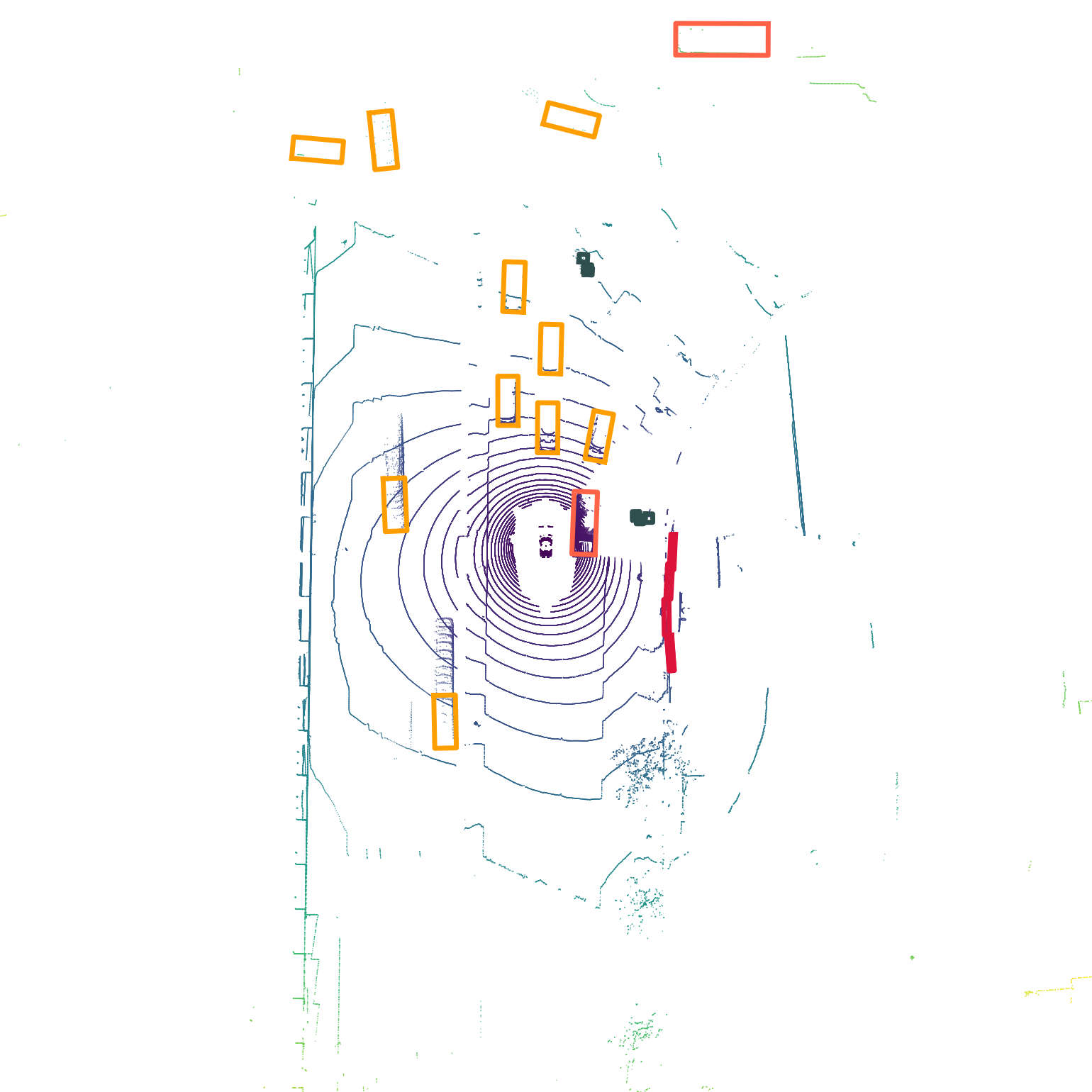}}
&\subfloat{\includegraphics[trim={2cm 35cm 2cm 35cm},clip,width=0.245\textwidth]{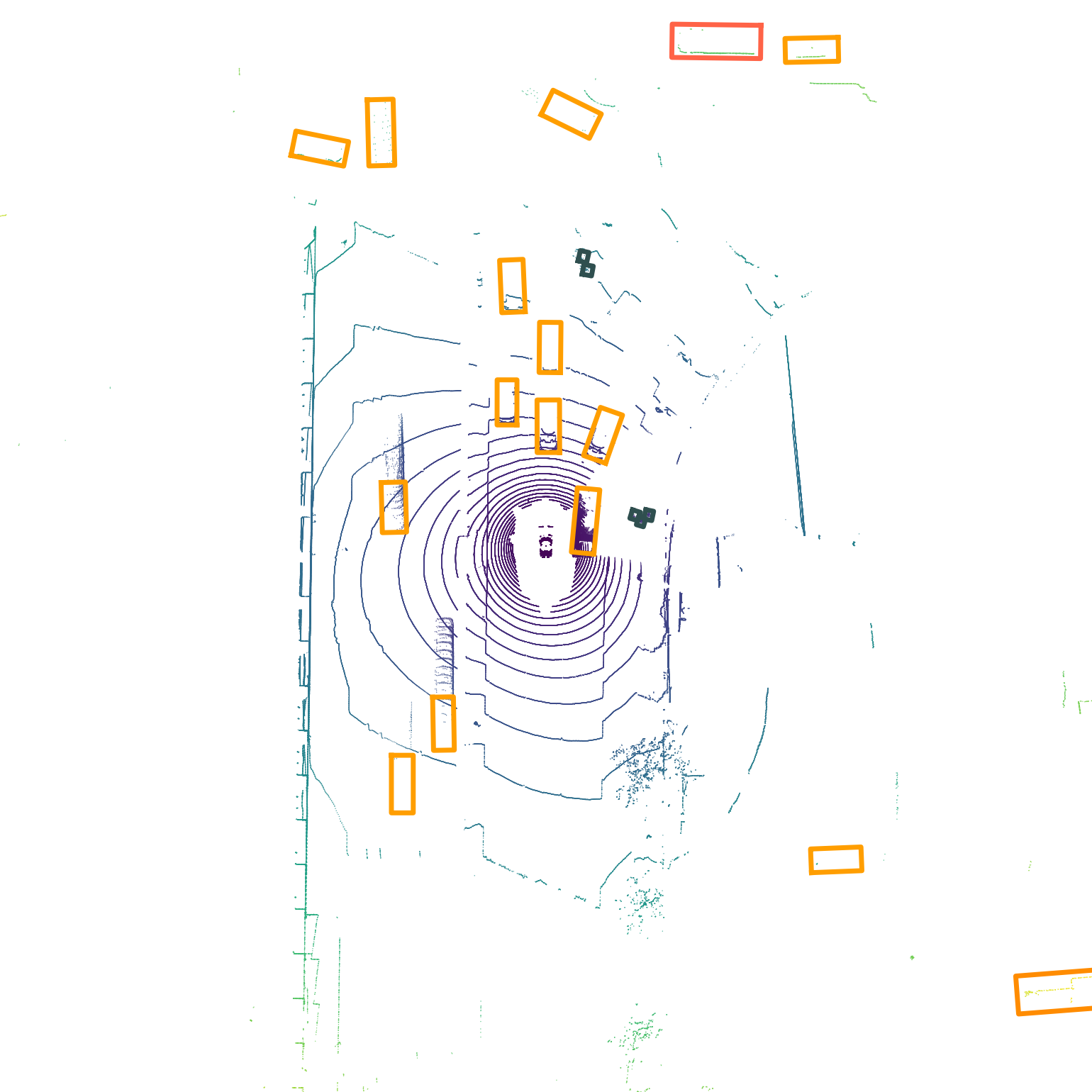}}\\\midrule
(g) &\subfloat{\includegraphics[trim={2cm 35cm 2cm 35cm},clip,width=0.245\textwidth]{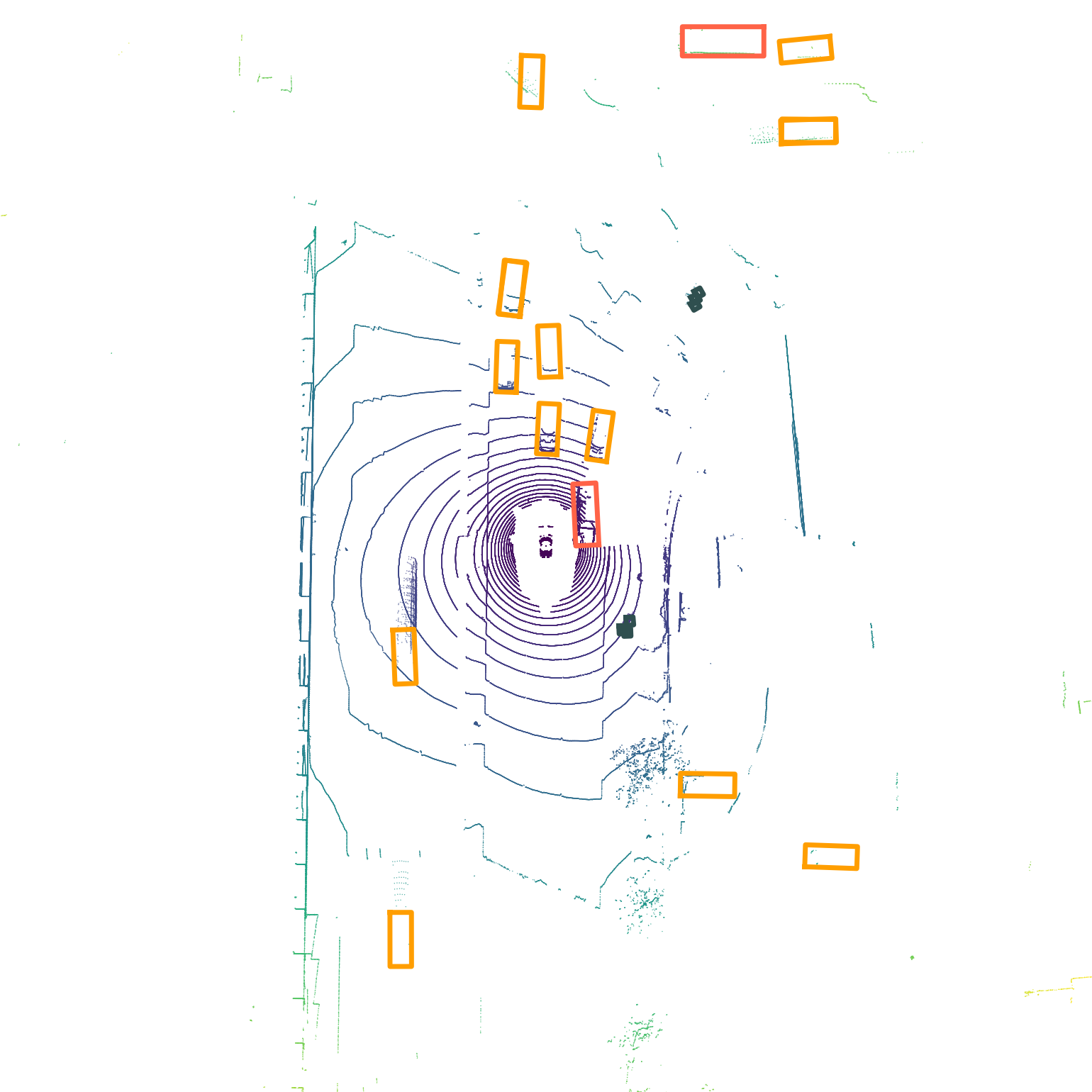}}
&\subfloat{\includegraphics[trim={2cm 35cm 2cm 35cm},clip,width=0.245\textwidth]{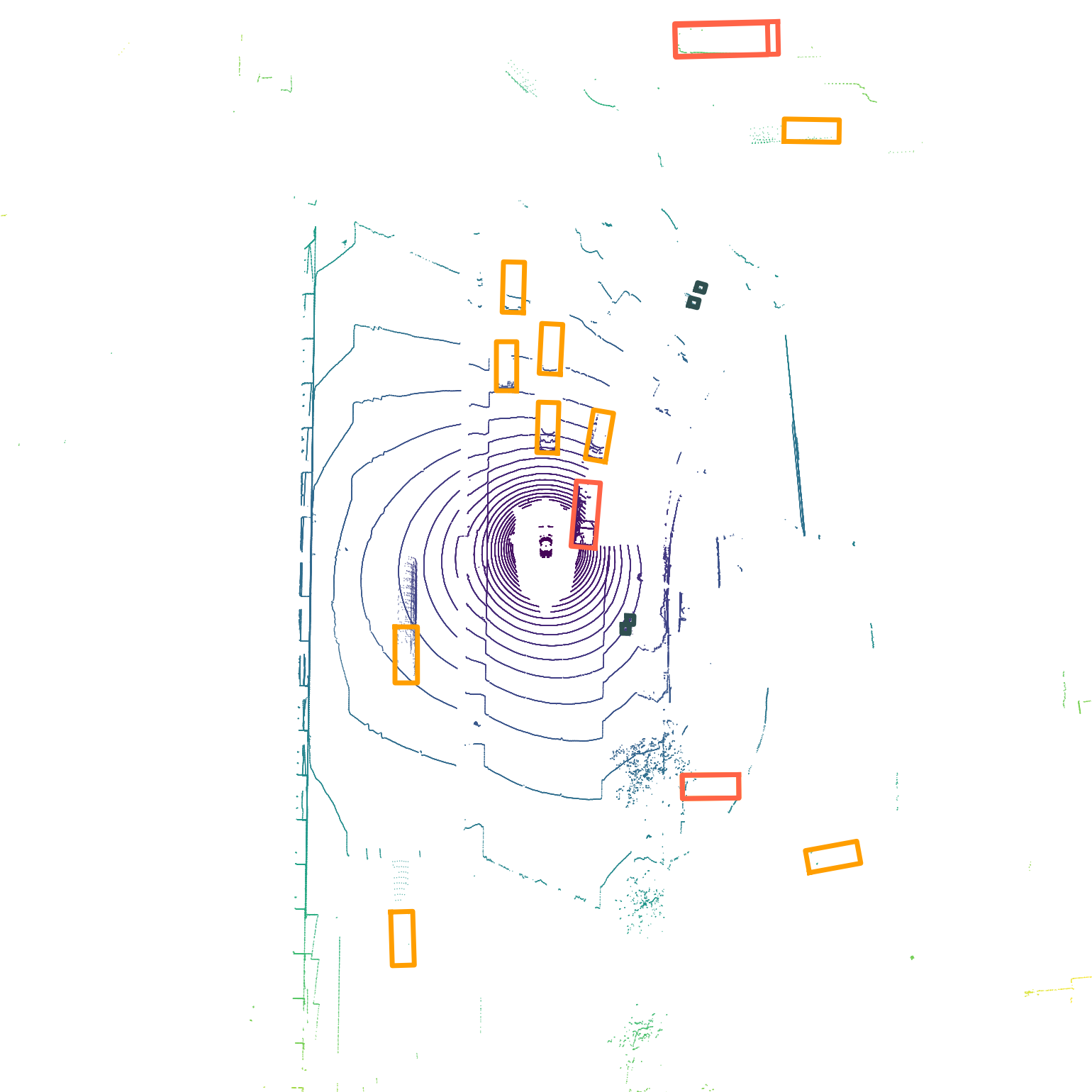}}
&\subfloat{\includegraphics[trim={2cm 35cm 2cm 35cm},clip,width=0.245\textwidth]{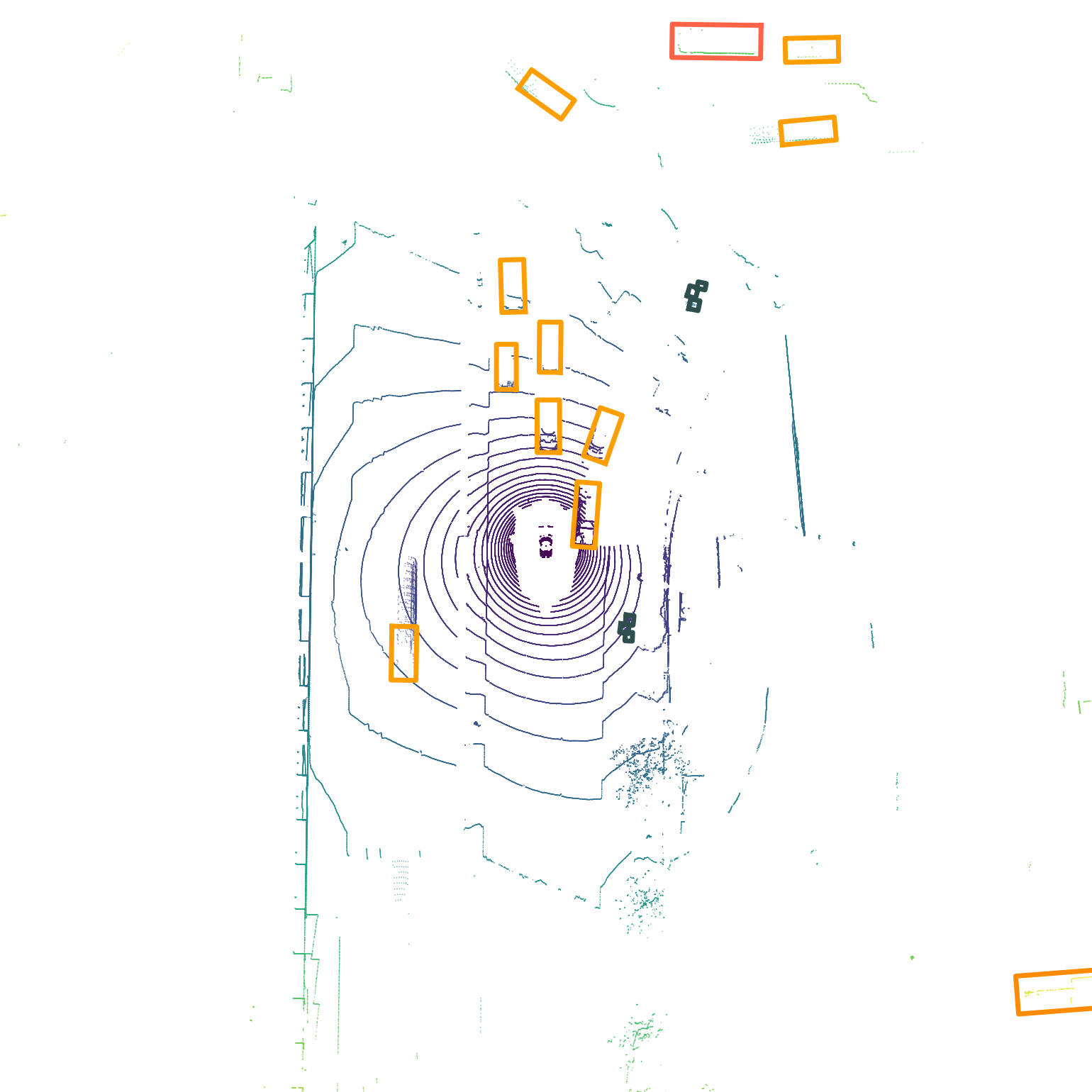}}\\
\end{tabular}
\caption{Qualitative comparison of object detection performance with the baseline UVTR~\citep{li2022uvtr} on nuScenes val set.}
\label{fig:appendix_objdet_qualitative}
\vspace{-0.5em}
\end{figure*}
\section{Additional Object Detection Results}
In this section we report more results on the object detection downstream task. We use the same pretrained model as in Sec.\,\ref{sec:benchmarking_results}, however we finetune the model using 100\% of the labelled data. Additionally we use all modern strategies commonly used for object detection such as CBGS\,\cite{zhu2019cbgs}, and Copy-Paste augmentation\,\cite{yan2018second}. We compare between NOMAE pretraining followed by 5 epochs fineuning and traning from scratch for 20 epochs (any longer schedule does not provide improvement). We also perform the same comparison using the commonly used CNN based SparseEncoder backbone.
The results in Tab.\ref{tab:appendix_objdet} shows that NOMAE brings consistent improvement to both NDS and mAP irrespective of the backbone architecture. Using NOMAE pretraining UVTR+PTv3 backbone achieves new SOTA for object detection on nuScenes val set.

\begin{table}[h]
    \centering
    \footnotesize
    \begin{tabular}{l|c|c}
    \toprule
        NDS/mAP&{Scratch 20ep}&{NOMAE +5ep}\\
        
        \midrule
        UVTR+PTv3&69.3/64.1& \textbf{71.2}/\textbf{67.0} \\
        UVTR+SparseEncoder~[37]&67.7/60.9& 70.4/65.3 \\
        \bottomrule
    \end{tabular}
    \caption{Results on nuScenes object detection val set for training from scratch compared to NOMAE pretraining.}
    \label{tab:appendix_objdet}
\end{table}

\section{Qualitative Analysis}
\label{sec:further_qualitative}
In this section, we present more qualitative results. 
Fig.~\ref{fig:appendix_semseg_qualitative} shows results for semantic segmentation and Fig.~\ref{fig:appendix_objdet_qualitative} object detection. 
For semantic segmentation, we observe the general trend of NOMAE correctly classifying smaller objects that are misclassified by the baseline (examples (c), (e), and (g)). Additionally, the boundaries between different semantic classes are more pronounced in the case of NOMAE (examples (a), (d), and (f)), which shows the importance of self-supervision at finer resolutions. We can also observe that NOMAE helps with the mixing between similar classes, such as trucks and buses in example (b), as well as different ground types in other examples.  
NOMAE also more accurately estimates the orientation of the objects and has higher true positive detections in the case of object detection.

\end{document}